\documentclass[twocolumn,twoside]{IEEEtran}
\usepackage{url}
\usepackage{etex}
\usepackage{amssymb}
\usepackage{cancel}
\newcommand{\cmark}{$\checkmark$}%
\newcommand{\xmark}{$\times$}%
\reserveinserts{28}
\usepackage{amsmath,graphicx}
\usepackage{amssymb}
\usepackage{algorithmic}
\usepackage{algorithm}
\usepackage{multirow,multicol}
\usepackage{pdfpages}
\usepackage{epstopdf}
\usepackage{subfigure}
\usepackage{balance}

\usepackage{cite}

\usepackage{amsthm}

\newtheorem{proposition}{Proposition}

\newtheorem{remark}{Remark}

\newtheorem{definition}{Definition}

{\begin{list}               
    {$\bullet$ \hfill}{
        \setlength{\leftmargin}{\parindent}
        \setlength{\parsep}{0.04\baselineskip}
        \setlength{\itemsep}{0.5\parsep}
        \setlength{\labelwidth}{\leftmargin}
        \setlength{\labelsep}{0em}}
    }
{\end{list}}

\providecommand{\eref}[1]{\eqref{#1}}  
\providecommand{\cref}[1]{Chapter~\ref{#1}}

\providecommand{\fref}[1]{Figure~\ref{#1}}

\providecommand{\E}{\ensuremath{\mathbb{E}}}
\providecommand{\N}{\ensuremath{\mathbb{N}}}

\providecommand{\Pb}{\ensuremath{\mathbb{P}}}

\providecommand{\bydef}{\overset{\text{def}}{=}}

\renewcommand{\vec}[1]{\ensuremath{\boldsymbol{#1}}}
\providecommand{\mat}[1]{\ensuremath{\boldsymbol{#1}}}

\providecommand{\calB}{\mathcal{B}}
\providecommand{\calC}{\mathcal{C}}

\providecommand{\calF}{\mathcal{F}}

\providecommand{\mA}{\mat{A}}
\providecommand{\mB}{\mat{B}}

\providecommand{\mD}{\mat{D}}

\providecommand{\mF}{\mat{F}}
\providecommand{\mG}{\mat{G}}
\providecommand{\mH}{\mat{H}}
\providecommand{\mI}{\mat{I}}

\providecommand{\mL}{\mat{L}}
\providecommand{\mM}{\mat{M}}

\providecommand{\mQ}{\mat{Q}}
\providecommand{\mR}{\mat{R}}

\providecommand{\mT}{\mat{T}}

\providecommand{\mW}{\mat{W}}

\providecommand{\mY}{\mat{Y}}
\providecommand{\mZ}{\mat{Z}}

\providecommand{\vb}{\vec{b}}
\providecommand{\vc}{\vec{c}}

\providecommand{\ve}{\vec{e}}

\providecommand{\vg}{\vec{g}}
\providecommand{\vh}{\vec{h}}

\providecommand{\vx}{\vec{x}}

\providecommand{\valpha}{\vec{\alpha}}
\providecommand{\vbeta}{\vec{\beta}}

\providecommand{\vtheta}{\vec{\theta}}

\providecommand{\vomega}{\vec{\omega}}

\providecommand{\vzero}{\vec{0}}
\providecommand{\vone}{\vec{1}}

\usepackage{color, colortbl}
\definecolor{LightCyan}{rgb}{0.88,1,1}
\definecolor{LightRed}{rgb}{1,0.1,0.4}
\definecolor{Gray}{gray}{0.9}

\begin{document}

\title{Color Filter Arrays for Quanta Image Sensors}
\author{Omar A. Elgendy,~\IEEEmembership{Student Member,~IEEE} and Stanley H. Chan,~\IEEEmembership{Senior Member,~IEEE}
\thanks{The authors are with the School of Electrical and Computer Engineering, Purdue University, West Lafayette, IN 47907, USA. Email: \texttt{\{oelgendy, stanchan\}@purdue.edu}. This work is supported, in part, by the National Science Foundation under grants CCF-1763896 and CCF-1718007.}}


\graphicspath{{pix/}}

\maketitle

\begin{abstract}
Quanta image sensor (QIS) is envisioned to be the next generation image sensor after CCD and CMOS. In this paper, we discuss how to design color filter arrays for QIS and other small pixels. Designing color filter arrays for small pixels is challenging because maximizing the light efficiency while suppressing aliasing and crosstalk are conflicting tasks. We present an optimization-based framework which unifies several mainstream color filter array design methodologies. Our method offers greater generality and flexibility. Compared to existing methods, the new framework can simultaneously handle luminance sensitivity, chrominance sensitivity, cross-talk, anti-aliasing, manufacturability and orthogonality. Extensive experimental comparisons demonstrate the effectiveness of the framework.
\end{abstract}
\begin{IEEEkeywords}
Quanta image sensor, single-photon detector, demosaicing, denoising, image reconstruction, color filter array
\end{IEEEkeywords}

\section{Introduction}
\subsection{Quanta Image Sensor}
\begin{figure*}[t]
\centering
\includegraphics[width=1\linewidth]{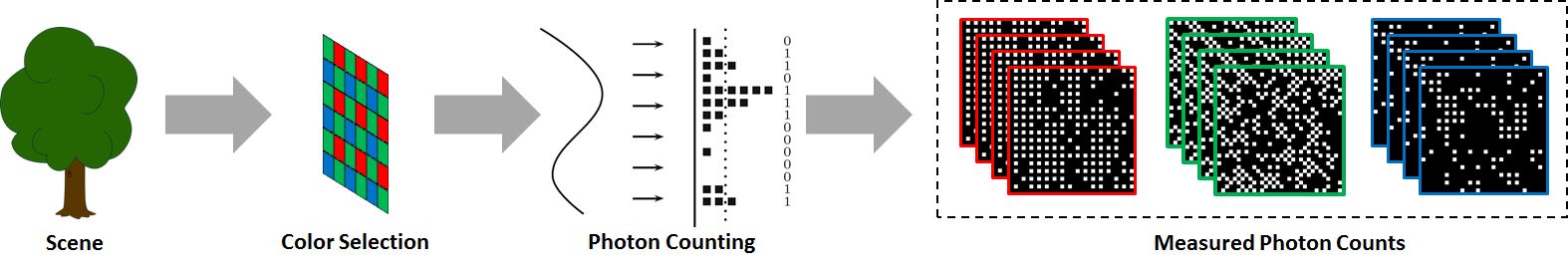}
\caption{\textbf{QIS Imaging Model \cite{Gnanasambandam_Elgendy_Ma_2019}}. When the scene image arrives at the sensor, the CFA first selects the wavelength according to the colors. Each color pixel is then sensed using a photon-detector and reports a binary value based on whether the photon counts exceeds certain threshold or not. The measured data contains three subsampled sequences, each representing a measurement in the color channel.}
\label{fig:qis}
\vspace{-2ex}
\end{figure*}

Quanta Image Sensor (QIS) is a new type of image sensors proposed by E. Fossum in 2005 \cite{Fossum_2011,Fossum_Ma_Masoodian_2016,Fossum_Ma_Masoodian_2016_1} as a candidate for the third generation digital image sensor after CCD and CMOS. The sensor comprises a massive array of sub-diffraction limit single-photon detectors, called ``jots\rq\rq{}, with a pixel pitch of $1.1\mu$m as of today. Having read-out noise of $0.21$e- and dark current of 0.16e-/sec, QIS can count incoming photons to produce a digital output of bit depth in the range of $1-5$ bits, assuming that the exposure does not saturate the sensor. As reported in \cite{Fossum_Ma_Masoodian_2016} and \cite{Fossum_Ma_Masoodian_2016_1}, the latest QIS sensor can achieve a resolution of $1024\times1024$ jots and frame rate up to $1000$ fps by using the commercial 45/65nm 3D-stacked backside illumination CMOS process.
Unlike CMOS image sensors that accumulate photons, QIS oversamples the scene by producing binary measurements. Figure~\ref{fig:qis} depicts the image formation process. See also \cite{Elgendy_Chan_2016,Chan_Elgendy_Wang_2016,Elgendy_Chan_2018} for detailed discussions.	

Despite the rapid advancement in QIS hardware \cite{Fossum_2011,Fossum_Ma_Masoodian_2016,Fossum_Ma_Masoodian_2016_1} and algorithms \cite{Yang_Lu_Sbaiz_2012,Elgendy_Chan_2016,Chan_Elgendy_Wang_2016,Elgendy_Chan_2018}, all reported findings, to-date, are based on monochromatic data. The first color QIS imaging was recently presented by Gnanasambandam et al. \cite{Gnanasambandam_Elgendy_Ma_2019}, where they demonstrated how to reconstruct a color image from the sensor using a Bayer color filter array. In this paper, we discuss how to design color filter array for better image acquisition.
%
\subsection{Color Filter Arrays for QIS}
A color filter array (CFA) is a mask placed on top of the sensor to select (filter) wavelengths. As light passes through the color filter array, the resulting image is a mosaic pattern of the three tri-stimulus RGB colors. Traditionally, CFA is organized as a periodic replica of a 2D kernel called the \textit{color atom}. The de-facto color atom used in the industry is the Bayer pattern \cite{Bayer_1976} because of its simplicity and the readily available demosaicking algorithms, see for instance \cite{Malvar_Cutler_2004,Alleysson_Susstrunk_Herault_2005,Dubois_2005,Hirakawa_Parks_2005,Dubois_2006,Leung_Jeon_Dubois_2011,Jeon_Dubois_2013,Korneliussen_Hirakawa_2014,Gharbi_Chaurasia_Paris_2016,Tan_Zeng_Lai_2017}. More sophisticated CFAs have been proposed \cite{Lukac_Plataniotis_2005,Hirakawa_Wolfe_2008,Lu_Vetterli_2009,Condat_2010,Hao_Li_Lin_2011,
Wang_Zhang_Hao_2011,Chakrabarti_Freeman_Zickler_2014,Amba_Dias_Alleysson_2016,Bai_Li_Lin_2016,Li_Bai_Lin_2017,Li_Bai_Lin_2017_1} to improve the Bayer CFA.

When designing a CFA, there are three factors that should be taken into consideration:

\begin{itemize}
\item \textbf{Aliasing}: Since color filtering is a sampling process, aliasing happens when the sampling rate is less than Nyquist. Aliasing causes false color artifacts at color edges, called the \textit{Moir\'e} artifacts \cite{Hirakawa_Wolfe_2008}. Color filters that are susceptible to aliasing, such as the Bayer CFA, require sophisticated demosaicking algorithms to suppress the Moir\'e artifacts. In contrast, a robust CFA can use less complicated demosaicking algorithms.
\item \textbf{Sensitivity}: Since CFA is a filter, it blocks part of the incoming light. This reduces the sensor sensitivity and makes the image more susceptible to noise. A good CFA design should maximize the sensitivity by allowing transparent or ``panchromatic\rq\rq{} color filters that block as few wavelengths as possible.
\item \textbf{Crosstalk}: Crosstalk can be either optical or electrical \cite{Anzagira_Fossum_2015}. If not treated, crosstalk will make colors look pale or de-saturated. Crosstalk desaturation is corrected by pixel-wise multiplication of the RGB color vector using a color correction matrix. However, color correction enhances residual noise in the image \cite{Chao_Tu_Chou_2011,Anzagira_Fossum_2015}. The situation is worsen in QIS because of its small size.
\end{itemize}

The three factors above are conflicting: Optimizing one generally degrades the others. For conventional CMOS image sensors, crosstalk is not severe, and so most CFA designs in the literature consider aliasing and sensitivity only. The first QIS color filter array design is proposed by Anzagira and Fossum \cite{Anzagira_Fossum_2015}. However, aliasing is not considered because QIS is sufficiently small. We generalize these prior work by considering all three factors so that our framework can be applied to both QIS and CMOS image sensors.

\subsection{Related Work}

The design framework we propose in this paper is a unification of several mainstream CMOS-based color filter arrays. To put our paper in the proper context in the literature, we list a few of the known works.

\vspace{1ex}
\noindent\textbf{Spatial CFA Design}: By suppressing the Moir\'e artifacts and crosstalk while keeping the demosaicing algorithm simple, Lukac and Plataniotis \cite{Lukac_Plataniotis_2005} proposed a CFA and compared it with other CFAs using a universal demosaicking method. However, their work did not provide a mathematical framework to analyze the CFA optimality.

\vspace{1ex}
\noindent\textbf{Spatio-Spectral CFA Design}: Hirakawa and Wolfe \cite{Hirakawa_Wolfe_2008} proposed a method through the spatial and spectral domain analysis. Their CFA reduces aliasing in the frequency domain, and possesses high sensitivity and numerical stability. Condat \cite{Condat_2011} extended the framework by optimizing luminance and chrominance sensitivity. He defined a new form of orthogonality between chrominance channels in the frequency domain. Hao et al. \cite{Hao_Li_Lin_2011} and Wang et al. \cite{Wang_Zhang_Hao_2011} proposed a framework based on symbolic discrete Fourier transform (DFT). Their CFA maximizes the numerical stability of linear demosaicking process under aliasing and physical constraints. Li et al. proposed spatio-spectral CFA design methods that are optimized for sensitivty  \cite{Bai_Li_Lin_2016,Li_Bai_Lin_2017}.

\vspace{1ex}
\noindent\textbf{Learning-based CFA Design}: By minimizing the average error on a color dataset, Lu and Vetterli \cite{Lu_Vetterli_2009} used an iterative algorithm to solve a least squares CFA design problem. Chakrabarti \cite{Chakrabarti_2016} and Henz et al. \cite{Henz_Gastal_Oliveira_2018} proposed to learn the optimal CFA pattern by using a deep neural network.

Besides these mainstream CFA design frameworks, there are a number of other CFA designs such as \cite{Condat_2010,Amba_Dias_Alleysson_2016,Chakrabarti_Freeman_Zickler_2014,Li_Bai_Lin_2017_1}.  On the hardware side, Biay-Cheng et al. \cite{Cheng_Siddiqui_Luo_2015, Siddiqui_Atanassov_Goma_2016} took into account that color filter fabrication technology lags the image sensor technology in terms of miniaturization. They proposed a hardware-friendly CFA assuming the color filter size is $1.5\times$ pixel size.


\subsection{Scope and Contributions}
%
In this paper, we propose an optimization framework that encompasses aliasing, sensitivity and crosstalk in a unified model. This is the first time we can incorporates a quantitative crosstalk metric in an optimization framework for CFA design. Our framework is general as it works with any crosstalk kernel depending on the sensor. Moreover, it works with any color atom size by a mild change in the formulation.

The main contribution of this paper is a general and flexible framework for CFA design. Compared to the existing CFA design framework, the new framework is able to simultaneously (Section~\ref{subsec:criteria})
\begin{itemize}
\item Improve CFA\rq{}s luminance and chrominance sensitivity,
\item Reduce aliasing between luminance and chrominance channels,
\item Suppress crosstalk between spectral sub-bands, and
\item Enforce orthogonality between chrominance channels to permit simple linear demosaicking.
\end{itemize}
The design framework is presented in the form of optimization. We have two designs: A convex optimization and a non-convex optimization. In addition to the formulation, we also present an algorithm to solve the non-convex optimization. (Section~\ref{sec:formulation})


For performance evaluation, we generalize the demosaicking algorithm of Condat \cite{Condat_2011} in Section~\ref{sec:Demo} so that it now works with arbitrary CFAs. This pipeline comprises a demosaicking by frequency selection algorithm for removing the CFA masking effect followed by a color correction step for removing the desaturation effect of crosstalk. Experimental evaluation on the DIV2K evaluation dataset in Section~\ref{subsec:results} shows the robustness of our proposed CFAs compared to other CFAs in literature.


\section{Background and Notations}\label{sec:background}
To facilitate readers to understand the design framework, in this section we introduce a few notations and terminologies. We will start in Section~\ref{subsec:imForm} by describing the image formation using a CFA, then we discuss CFA in different domains in Sections~\ref{subsec:CFAAnalysis} and \ref{subsec:fourier}. Afterwards, in Section~\ref{subsec:def}, we define the optimization variables to simplify the design framework.

\begin{figure*}[t]
\centering
\includegraphics[width=1\linewidth]{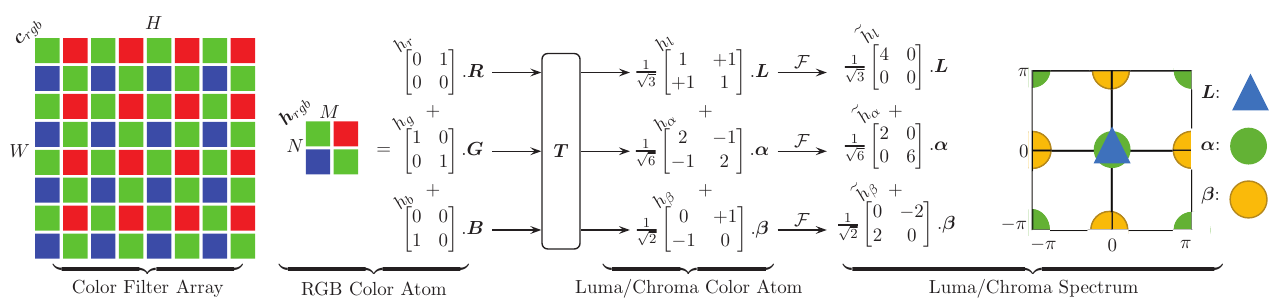}
\caption{Our terminology illustrated on the Bayer CFA example. The building unit of a CFA is a color atom. A transformation $T$ is applied to the color atom to transform it from the canonical RGB color space to a luma/chroma color space to simplify the design process. Foruier transform is applied afterwards to obtain the color atom spectrum in the luma/chroma space.}
\label{fig:terminology}
\vspace{-2ex}
\end{figure*}

\subsection{Color Image Formation}\label{subsec:imForm}
Consider a color image $\mathbf{im}_{rgb}$ of size $H \times W$. We denote the normalized light intensities in the red, green and blue channels for the ($m$-th,$n$-th) pixel of the color image as
\begin{equation}
\mathbf{im}_{rgb}(m,n)= \begin{bmatrix} \mathrm{im}_r(m,n) \\ \mathrm{im}_g(m,n) \\ \mathrm{im}_b(m,n) \end{bmatrix},
\end{equation}
where $m\in\{0,\ldots,H-1\}$, $n\in\{0,\ldots,W-1\}$.

\noindent\textbf{Color Filtering}: To obtain color, we place a color filter on top of each jot to collect light for one of the RGB colors. The CFA is a periodic pattern of the same resolution of $\mathbf{im}_{rgb}$, defined as
\begin{equation}
\vc_{rgb}(m,n) =
\begin{bmatrix}
c_r(m,n) \\ c_g(m,n) \\ c_b(m,n)
\end{bmatrix},
\end{equation}
where $c_r(m,n)$, $c_g(m,n)$, $c_b(m,n) \in [0,1]$ are the opacity rates for the red, green and blue pixels, respectively. For example, a red color filter is defined as $\vc_{rgb}(m,n) = [1,0,0]^T$ as it only passes the red color. The light exposure on the QIS after passing through the CFA is denoted as $\theta(m,n)$, which is a linear combination of the tri-stimulus colors:
\begin{equation}
\label{eq:CFAsampling}
\begin{split}
\theta(m,n) &= \eta\vc_{rgb}(m,n)^T\mathbf{im}_{rgb}(m,n)\\
&=\eta\sum_{i\in\{r,g,b\}} \; c_i(m,n) \,\textrm{im}_i(m,n).
\end{split}
\end{equation}
Here, $\eta$ is a positive scalar representing the sensor gain factor.

\noindent\textbf{Photon Arrival}. The photon arrival is modeled as a Poisson process. Let $\mY \in \N^{HW}$ be a vector of non-negative random integers denoting the number of photons arriving at QIS jots according to the light exposure $\vtheta$. Then, the probability of observing a photon count $Y_j = y_j, j\in\{1,\ldots,HW\}$ is
\begin{equation}\label{eq:prob_Ym}
\Pb(Y_j = y_j)= \frac{\theta_m ^{y_j} e^{-\theta_j}}{y_j!}.
\end{equation}
In this work, we assume \textit{single-bit QIS} \cite{Fossum_2013} that quantizes the photon counts by QIS jots to a binary values $\mB \in \{0,1\}^{HW}$ with $B_j = 1$ if $Y_j \ge q$ and $B_j = 0$ if $Y_j < q$, where $q > 0$ is a threshold. The probability of observing $B_j = b_j$ is
\begin{equation}\label{eq:prob_Bm}
\Pb(B_j = b_j)= \Psi_q(\theta_j)^{1-b_j} \left(1-\Psi_q(\theta_j)\right)^{b_j},
\end{equation}
where $\Psi_q(\cdot)$ is the incomplete Gamma function \cite{Elgendy_Chan_2018}.

\vspace{1ex}
\noindent\textbf{Temporal Oversampling}. With frame rates that reach $1000$ fps, QIS is able to catch the scene movement by taking $T$ temporal samplings for the same scene. This allows us to utilize multiple independent measurements over time to improve the statistics and decrease noise. Hence, for every jot with light exposure $\theta_j$, we have a set of $T$ independent binary measurements $\calB_j = \{b_{j,0},\ldots,b_{j,T-1}\}$.

\subsection{Color Filter Array Analysis in Different Color Spaces}\label{subsec:CFAAnalysis}
Since the CFA $\vc_{rgb}(m,n)$ is a periodic pattern, it is sufficient to use a \emph{color atom} as the optimization variable when designing the CFA. The color atom takes the form
\begin{equation}
\vh_{rgb}(m,n)=
\begin{bmatrix}
h_r(m,n) \\
h_g(m,n) \\
h_b(m,n)
\end{bmatrix},
\end{equation}
where each of $h_r$, $h_g$ and $h_b$ is an $M \times N$ array. For instance, the GRBG Bayer pattern has the following color atom (when $M = N = 2$):
\begin{equation*}
h_r = \begin{bmatrix} 0&1\\0&0 \end{bmatrix}, \; h_g = \begin{bmatrix} 1&0\\0&1 \end{bmatrix}, h_b = \begin{bmatrix} 0&0\\1&0 \end{bmatrix},
\end{equation*}
because the Bayer pattern has one red pixel and one blue pixel located at two opposite diagonals, and two green pixels located in the remaining two positions. Figure~\ref{fig:terminology} illustrates the idea.

While the primal RGB color is common for making the CFA, it would be more convenient if the colors are \emph{decorrelated}. To this end, we change the image representation from the canonical RGB basis to an orthornormal basis using a transformation matrix \cite{Hel_2004,Condat_2011}:
\begin{equation}\label{eq:basis}
\mT = \begin{bmatrix}
1/\sqrt{3} & 1/\sqrt{3} & 1/\sqrt{3}  \\
-1/\sqrt{6} & 2/\sqrt{6} & -1/\sqrt{6} \\
1/\sqrt{2} & 0 & -1/\sqrt{2}
\end{bmatrix}.
\end{equation}
This transformation maps an RGB image $\mathbf{im}_{rgb}$ to an image  $\mathbf{im}_{l\alpha\beta}=[\textrm{im}_l,\textrm{im}_\alpha,\textrm{im}_\beta]^T$ as follows (we drop the spatial indices $(m,n)$ for simplicity)
\begin{align*}
 \mathbf{im}_{l\alpha\beta} =
\begin{bmatrix} \textrm{im}_l \\ \textrm{im}_\alpha\\ \textrm{im}_\beta \end{bmatrix}
=\mT
\begin{bmatrix} \textrm{im}_r \\ \textrm{im}_g \\ \textrm{im}_b  \end{bmatrix}= \begin{bmatrix} \left(\textrm{im}_r+\textrm{im}_g+\textrm{im}_b\right)/\sqrt{3} \\
\left(-\textrm{im}_r+2\textrm{im}_g+\textrm{im}_b\right)/\sqrt{6}\\
\left(\textrm{im}_r-\textrm{im}_b\right)/\sqrt{2} \end{bmatrix},
\end{align*}
where $\textrm{im}_l$ is a luminance (luma) component that contains the high frequency components such as edges and textures, whereas $\textrm{im}_\alpha$ and $\textrm{im}_\beta$ are chrominance (chroma) components that carry the color information.

Since $\mT$ is orthonormal (i.e., $\mT^T\mT=\mI$), we can rewrite the sampling process in Equation \eref{eq:CFAsampling} in the luma/chroma space:
\begin{align} \label{eq:mosaickImage}
\theta(m,n)
&= \eta \vc_{rgb}(m,n)^T\; \boldsymbol{T}^T  \boldsymbol{T}\;
\mathbf{im}_{rgb}(m,n)                                          \notag \\
&= \eta\vc_{l\alpha\beta}(m,n)^T \;\mathbf{im}_{l\alpha\beta}(m,n)  \notag \\
&= \eta\sum_{i\in\{l,\alpha,\beta\}} \; c_i(m,n) \,\textrm{im}_i(m,n),
\end{align}
where $c_l(m,n)$, $c_\alpha(m,n)$ and $c_\beta(m,n)$ are the luma/chroma representation of the CFA, with
\begin{equation}
\vc_{l\alpha\beta}(m,n) =
\mT \; \vc_{rgb}(m,n).
\end{equation}
The luma/chroma representation of the CFA has a corresponding color atom $h_l(m,n)$, $h_\alpha(m,n)$ and $h_\beta(m,n)$. For instance, the luma/chroma color atom of the GRBG Bayer pattern is
\begin{equation}
\vh_{l\alpha\beta}(m,n)=
\begin{bmatrix}
h_l(m,n) \\
h_{\alpha}(m,n) \\
h_{\beta}(m,n)
\end{bmatrix},
\label{eq: h l alpha beta}
\end{equation}
where the individual components are
\begin{align*}
h_l =\frac{1}{\sqrt{3}} \begin{bmatrix} 1&1\\1&1 \end{bmatrix}, h_\alpha = \frac{1}{\sqrt{6}}\begin{bmatrix} 2&-1\\-1&2 \end{bmatrix}, h_\beta = \frac{1}{\sqrt{2}}\begin{bmatrix} 0&1\\-1&0\end{bmatrix}.
\end{align*}

\begin{remark}
In principle, there are infinite choices for the luma/chroma basis $\mT$. We choose the one in Equation \eref{eq:basis} because it makes the components of natural images statistically independent in the first order approximation.
\end{remark}

\subsection{Color Filter Array in Fourier Space}\label{subsec:fourier}
When analyzing the aliasing effects of the CFAs, we need to transform the color atom into the Fourier domain. For simplicity, we represent the Fourier transform of a signal by putting a tilde on top of the symbol, e.g., $h\overset{\calF}{\rightarrow}\widetilde{h}$. The 2D discrete Fourier transform (DFT) of the $i$-th color atom is
\begin{equation}\label{eq:DFTatom}
\widetilde{h}_i(u,v) = \sum_{m=0}^{M-1} \sum_{n=0}^{N-1}\; h_i(m,n) e^{-j2\pi\left(\frac{mu}{M}+\frac{nv}{N}\right)}
\end{equation}
where $u\in\{0,\ldots,M-1\}$, $v\in\{0,\ldots,N-1\}$.

\begin{figure}[h]
\centering
\includegraphics[width=1\linewidth]{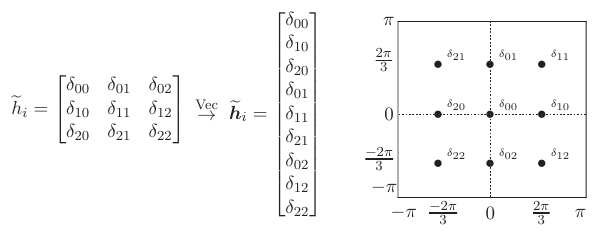}
\caption{The Fourier representation of an arbitrary $3\times3$ color atom $i$. From left to right: The atom representation, the vector representation and the 2D frequency plane representation. Notice that the frequency plane is divided into $9$ regions of size $2\pi/3$, and the spectrum comprises pure sinusoids placed at $(\frac{2\pi u}{3},\frac{2\pi v}{3})\; \forall u,v\in\{0,1,2\}$.}
\label{fig:freqDomain1}
\vspace{-2ex}
\end{figure}

For example, the discrete Fourier transform of the luma/chroma color atoms in Equation \eref{eq: h l alpha beta} are
\begin{equation*}
\widetilde{h}_l =\frac{1}{\sqrt{3}} \begin{bmatrix} 4&0\\0&0 \end{bmatrix}, \widetilde{h}_\alpha=\frac{1}{\sqrt{6}}\begin{bmatrix} 2&0\\0&6 \end{bmatrix}, \widetilde{h}_\beta = \frac{1}{\sqrt{2}}\begin{bmatrix} 0&-2\\2&0\end{bmatrix}.
\end{equation*}
Here, we observe that the Fourier transform of the color atom has the same size as the original color atom. The luminance channel has only one baseband components at $(0,0)$, whereas the $\alpha$ chrominance channel has one baseband component and a component at $(\pi,\pi)$. The $\beta$ chrominance channel has two components at $(0,\pi)$ and $(\pi,0)$. \fref{fig:freqDomain1} illustrates how these frequency locations are identified from a $3\times3$ color atom.

While the Fourier transform of the color atom is useful, for demosaicing we also need to analyze the spectrum of the entire CFA. As shown by Hao et al. \cite{Hao_Li_Lin_2011}, the Fourier transform of the entire CFA can be written in terms of the Fourier transform of the color atoms:
\begin{equation}
	\widetilde{c}_i(\vomega) = \sum_{u=0}^{M-1} \sum_{v=0}^{N-1} \widetilde{h}_{i}(u,v)\delta\left(\vomega-\vomega(u,v)\right),
\end{equation}
where $i\in\{l,\alpha,\beta\}$, $\vomega$ is the 2D angular frequency, and
\begin{equation}\label{eq:omega}
\vomega(u,v)=\left(\frac{2\pi u}{M},\frac{2\pi v}{N}\right) \; \forall \begin{array}{c}u\in\{0,\ldots,M-1\}\\ v\in\{0,\ldots,N-1\}\end{array}.
\end{equation}
is the ($u$-th,$v$-th) 2D angular frequency. It is worth noting that the Fourier transform of the CFA comprises pure sinusoids of amplitudes $\widetilde{h}_{i}(u,v)$. These sinusoids are placed at $MN$ discrete 2D frequencies $\vomega(u,v)$ that divide the 2D frequency plane $[-\pi,\pi]\times[-\pi,\pi]$ into $MN$ equal regions. Therefore, the spectrum of the mosaicked image $\widetilde{\theta}(\vomega)$ can be written as
\begin{align} \label{eq:modulation}
\widetilde{\theta}(\vomega) &= \calF \left(\sum_{i\in\{l,\alpha,\beta\}} \; c_i \;\textrm{im}_i\right) = \sum_{i\in\{l,\alpha,\beta\}} \; \widetilde{c}_i(\vomega) \circledast \widetilde{\textrm{im}}_i(\vomega) \nonumber \\
	&\hspace{-2ex}=\sum_{i\in\{l,\alpha,\beta\}} \sum_{u=0}^{M-1} \sum_{v=0}^{N-1} \; \widetilde{h}_{i}(u,v)\;\widetilde{\textrm{im}}_i(\vomega-\vomega(u,v)),
\end{align}
where $\circledast$ is the standard 2D convolution operator. Having the spectrum of the mosaicked image $\widetilde{\theta}(\vomega)$, we can now discuss the optimization variables in our problem.

\subsection{Design Variables}\label{subsec:def}
To formulate the CFA design problem as an optimization problem, we define the following variables. We denote $\vh_r$, $\vh_g$ and $\vh_b$ the vectorized representations of the red, green and blue color atoms, respectively. To ensure physical realizability, we require $\vh_r$, $\vh_g$, $\vh_b \in \left[0,1\right]^{K \times 1}$, where $K\bydef MN$, and we stack all design variables into one long vector
\begin{equation*}
\vx=\begin{bmatrix} \vh_r \\ \vh_g \\ \vh_b \end{bmatrix}\in \mathbb{R}^{3K\times 1}.
\end{equation*}
The design variable $\vx$ is related to the vectorized RGB and luma/chroma color atoms as
\begin{equation*}
\begin{bmatrix}
\vh_r\\
\vh_g\\
\vh_b
\end{bmatrix}
=
\begin{bmatrix}
\mZ_r\\
\mZ_g\\
\mZ_b
\end{bmatrix}
\vx\quad\textrm{and}\quad
\begin{bmatrix}
\vh_l\\
\vh_\alpha\\
\vh_\beta
\end{bmatrix}
=
\begin{bmatrix}
\mZ_l\\
\mZ_\alpha\\
\mZ_\beta
\end{bmatrix}
\vx,
\end{equation*}
where the $\mZ$ matrices are defined by Equation \eref{eq:basis} as
\begin{align*}
\begin{array}{lcl}
\mZ_r =[\mI, \boldsymbol{0}, \boldsymbol{0}] &             &\mZ_l = [\mI, \mI, \mI]/\sqrt{3},\\
\mZ_g=[ \boldsymbol{0}, \mI, \boldsymbol{0}] & \mbox{ and }& \mZ_\alpha=[-\mI, 2\mI, -\mI]/\sqrt{6},\\
\mZ_b=[\boldsymbol{0}, \boldsymbol{0},\mI]   &             &\mZ_\beta= [ \mI, \boldsymbol{0}, -\mI]/\sqrt{2}.
\end{array}
\end{align*}
Given the design variable $\vx$, we also need to analyze its spectrum.
We write the 2D Fourier transform equation \eref{eq:DFTatom} as a matrix-vector product by using the Fourier transform matrix $\mF\in \mathcal{C}^{K \times K}$. Hence, the vectorized spectra of the luma/chroma color atoms can be written in terms of $\vx$ as
\begin{align}
\widetilde{\vh}_i = \mF \vh_i =  \mF\mZ_i \vx, \; i\in\{l,\alpha,\beta\}.
\label{eq: h tilde}
\end{align}
where $\widetilde{\vh}_i\in\calC^{K\times 1}$, for $i \in \{l,\alpha,\beta\}$. The relation between the matrix and the vector forms of the Fourier transform is:
\begin{equation}
\widetilde{h}_i(u,v)= \mbox{vec}^{-1}(\widetilde{\vh}_i)
\end{equation}
where $\widetilde{h}_i(u,v)$ is the Fourier coefficient.

\section{Design Criteria}\label{subsec:criteria}

\begin{table*}[t]
\centering
\caption{CFA Design Criteria}
\begin{tabular}{c|c|c|c|c|c|c|c}
\multirow{ 2}{*}{Criterion} & \multirow{ 2}{*}{Purpose}  & \multicolumn{4}{c|}{Regular Pixels} & \multicolumn{2}{c}{QIS}   \\
\cline{3-8}
& & \cite{Hirakawa_Wolfe_2008} & \cite{Condat_2011} & \cite{Hao_Li_Lin_2011} & \cite{Cheng_Siddiqui_Luo_2015} & \cite{Anzagira_Fossum_2015} & Ours\\
\hline
\hline
Proposition~\ref{prop:gammas} & To minimize noise power after linear demosaicking  & \xmark & \cmark & \xmark &  \xmark & \xmark & \cmark \\
\hline
Proposition~\ref{prop:uniformluma} & To simplify denoising of luminance channel        & \cmark & \cmark & \cmark &  \cmark & \cmark & \cmark \\
\hline
Proposition~\ref{prop:aliasing} & To maximize spatial resolution                               & \cmark & \cmark & \cmark &   \xmark & \xmark & \cmark \\
\hline
Proposition~\ref{prop:ctk} & To mitigate cross-talk                               & \xmark & \xmark & \xmark &  \cmark & \cmark & \cmark \\
\hline
Definition~\ref{def: orthogonality} & Enforce total orthogonality      & \cmark & \xmark & \cmark & \xmark & \xmark & \cmark \\
\hline
Definition~\ref{def: orthogonality} & Enforce quadrature orthogonality & \xmark & \cmark & \xmark & \xmark & \xmark & \cmark \\
\hline
\end{tabular}
\label{tab:criteria}
\vspace{-2ex}
\end{table*}

We now present the design criteria. Our criteria unify the three major approaches in the literature: (i) Sensitivity of luma/chroma channels to noise by Condat \cite{Condat_2011}; (ii) Aliasing between different color components in the frequency domain by Hirakawa and Wolfe \cite{Hirakawa_Wolfe_2008}; (iii) Crosstalk between neighboring pixels in the spatial domain by Anzagira and Fossum \cite{Anzagira_Fossum_2015}. Note that the first two criteria were developed for CMOS, whereas the third criterion was developed for QIS. The proposed framework integrates all these criteria into a unified formulation. Table~\ref{tab:criteria} summarizes the difference between this paper and the previous works. \fref{fig:ex} shows an example CFA, and its frequency representation.

In the following subsections, we present the design criteria and express them in terms of matrix-vectors for the optimization framework in Section IV.


\begin{figure}[t]
\centering
\includegraphics[width=1\linewidth]{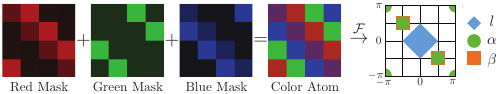}
\caption{A $4\times4$ CFA generated by our design framework. Luminance sensitivity $\gamma_l$ and chrominance sensitivity $\gamma_c$ are maximized to improve robustness to noise (Section~III-A). No chrominance components ($\alpha$ or $\beta$) are modulated on the vertical and horizontal frequency axes (Section~III-B) to mitigate aliasing with the luminance component $l$. The total variation of the red, green and blue masks is upper-bounded by $\textrm{TV}_{\max}$ to mitigate crosstalk (Section~III-C).}
\label{fig:ex}
\vspace{-2ex}
\end{figure}

\subsection{Luminance and Chrominance Sensitivity}
\begin{definition}
The luminance sensitivity $\gamma_l$ and the chrominance sensitivity $\gamma_c$ of a CFA with color atom $\{\vh_l, \vh_{\alpha}, \vh_{\beta}\}$ of size $M\times N$ are defined as
\begin{equation}\label{eq:def}
\gamma_l \bydef \frac{1}{K}||\widetilde{\vh}_l||_2, \textrm{ and } \gamma_c\bydef\frac{1}{K}\textrm{min}\left(||\widetilde{\vh}_{\alpha}||_2,||\widetilde{\vh}_{\beta}||_2\right).
\end{equation}
where $K=MN$ is a normalization factor.
\end{definition}

Intuitively, the luminance and chrominance sensitivity are measures of the signal power that can be transmitted through the color filter. A more transparent color filter allows more light to pass through, and hence the signal power is higher. This is reflected by the magnitudes $\|\widetilde{\vh}_i\|_2$ for $i \in \{l, \alpha,\beta\}$, which according to Parseval's Theorem they are equivalent to $\|\vh_i\|_2$.

The following proposition shows how can we compute $\gamma_l$ and $\gamma_c$ in terms of the optimization vector $\vx$.
\begin{proposition}\label{prop:gammas}
For a CFA with color atoms represented by the vector $\vx$, the luminance and chrominance sensitivity can be calculated as
\begin{equation}
\begin{split}
\gamma_l(\vx) &= \frac{1}{K}\vone^T \mZ_l \vx = \vb^T \vx \\
\gamma_c(\vx) &= \min \left( \sqrt{\vx^T \mQ_\alpha \vx}, \;\; \sqrt{\vx^T \mQ_\beta \vx}\right),
\end{split}
\end{equation}
where $\vb = \frac{1}{K}\mZ_l^T \vone $, $\mQ_{\alpha} = \mZ_\alpha^T\mZ_\alpha$ and $\mQ_{\beta} = \mZ_\beta^T\mZ_\beta$.
\end{proposition}
\begin{IEEEproof}
See Appendix A.
\end{IEEEproof}

The luminance sensitivity and the chrominance sensitivity cannot be arbitrarily chosen. One practical consideration is to ensure uniform noise power across the luma channel so that the denoising procedure can be simplified (because the noise will be i.i.d.). Thus, the luminance color atom $h_l$ should be constant, i.e.,  $h_l(m,n)=c,\forall m,n$, where $c$ is a positive constant. Taking Fourier transform, this means that $\widetilde{h}_l$ comprises only one impulse at baseband $\widetilde{h}_l(0,0)$, and no impulses at all other frequencies. In vector form, we need
\begin{equation}\label{eq:basebad}
\widetilde{\vh}_l - \mbox{diag}(\ve_1 )\widetilde{\vh}_l = \boldsymbol{0},
\end{equation}
where $\ve_1 = [1, 0, \ldots, 0]^T$ is the standard basis. Putting in terms of the optimization variable $\vx$, we have a constraint.

\begin{proposition}[Uniform Luminance Constraint]\label{prop:uniformluma}
If a CFA has a uniform luminance sensitivity, then $\vx$ needs to satisfy
\begin{equation}
(\mI-\mbox{diag}(\ve_1)) \mF \mZ_l \, \vx = \vzero.
\end{equation}
\end{proposition}
\begin{proof}
Using Equation \eref{eq: h tilde}, substitute $\widetilde{\vh}_l =  \mF \mZ_l \vx$ into Equation \eref{eq:basebad}.
\end{proof}

\subsection{Anti-Aliasing}
\label{subsec:aliasing}
In the frequency representation of a mosaicked image, the luminance controls the baseband whereas the chrominance controls the sideband of the spectrum. To minimize spectral interference, i.e., aliasing, CFA design methods modulate the chrominance as far as possible from the baseband. Typically, these methods do not put chrominance on the vertical and horizontal axes to prevent aliasing. However, since QIS is very small and it oversamples the scene, it is possible to relax the aliasing constraint. To this end, we allow chrominance channels to be placed on the vertical and horizontal axis as long as they are placed at frequencies higher than or equal $\pi/2$. This enlarges the feasible set of the solutions.

Mathematically, the anti-aliasing requirement is formulated by forcing the Fourier coefficients of the chrominance color atoms at $(\pm \pi,v)$ and $(u, \pm \pi)$ to zero for all $u$ and $v$ greater than $\pi/2$.  In terms of our design variable $\vx$, we require the following constraint.
\begin{proposition}[Anti-aliasing Constraint]\label{prop:aliasing} The chrominance placed on frequencies less than $\pi/2$ in the vertical and horizontal directions must be set to 0. Hence, $\vx$ must satisfy
\begin{equation}\label{eq:aliasing}
\begin{bmatrix} \mW_\alpha \\ \mW_\beta\end{bmatrix} \vx = \mW \vx =  \mathbf{0}
\end{equation}
where $\mW_\alpha$ and $\mW_\beta$ are the matrices formed by choosing the rows in $\mF\mZ_\alpha$ and $\mF\mZ_\beta$, respectively, that correspond to vertical and horizontal frequency components that are less than $\pi/2$.
\end{proposition}

To quantify the amount of aliasing for every CFA, we define the following aliasing criterion.
\begin{definition}
For a CFA, aliasing between luminance and chrominance channels is measured by
\begin{equation}\label{eq:aliasing}
J_{l} \bydef \frac{1}{HW} \int_{[-\pi,\pi)^2} \frac{\left(S_l(\vomega)S_{\alpha}(\vomega)+S_l(\vomega)S_{\beta}(\vomega)\right)}{S_\theta(\vomega)} d\vomega,
\end{equation}
where $S_l$, $S_\alpha$, $S_\beta$ and $S_\theta$ denote the power spectral density of the luminance channel $\mathbf{im}_l$, the two chrominance channels $\mathbf{im}_\alpha$ and $\mathbf{im}_\beta$, and the mosaicked image $\vtheta$, respectively.
\end{definition}

\subsection{Crosstalk}\label{subsec:ctk}
Crosstalk is caused by the leakage of electrical and optical charge from a pixel to its adjacent pixels \cite{Hirakawa_2008,Anzagira_Fossum_2015}. Crosstalk leads to color de-saturation. To model crosstalk, we follow \cite{Anzagira_Fossum_2015} by defining three scalars $\delta_r$, $\delta_g$, and $\delta_b$ representing the proportion of leaked charges to neighboring pixels. These three scalars then form a \emph{crosstalk kernel},
\begin{equation}\label{eq:crosstalkKernels}
g_i=\begin{bmatrix}
0 & \delta_i/4 & 0\\
\delta_i/4 & 1-\delta_i & \delta_i/4\\
0 & \delta_i/4 & 0
\end{bmatrix}, \; i\in\{r,g,b\},
\end{equation}
which can be considered as a spatial lowpass filter of the mosaicked image. Applying the crosstalk kernel to the CFA is equivalent to a spatially invariant convolution
\begin{equation*}
h_i^{\mathrm{ctk}} = g_i \circledast h_i, \; i\in\{r,g,b\},
\end{equation*}
where $h_i^{\mathrm{ctk}}$ denotes the effective CFA in the presence of crosstalk.

\begin{figure}[t]
\centering
\begin{tabular}{ccc}
\includegraphics[width=0.28\linewidth]{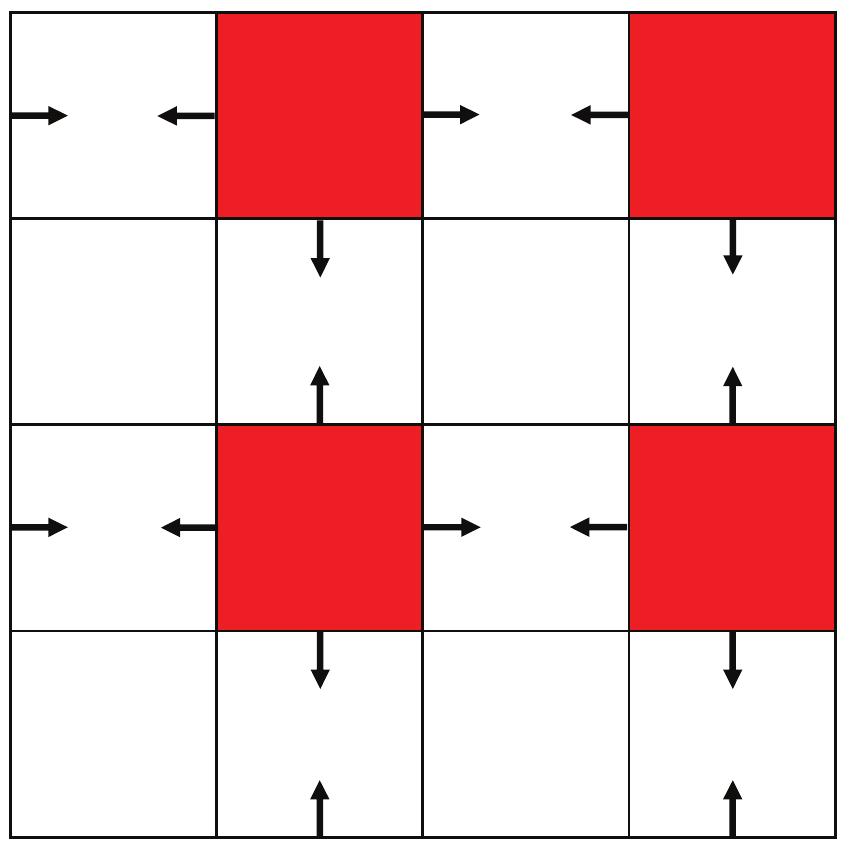}&
\includegraphics[width=0.28\linewidth]{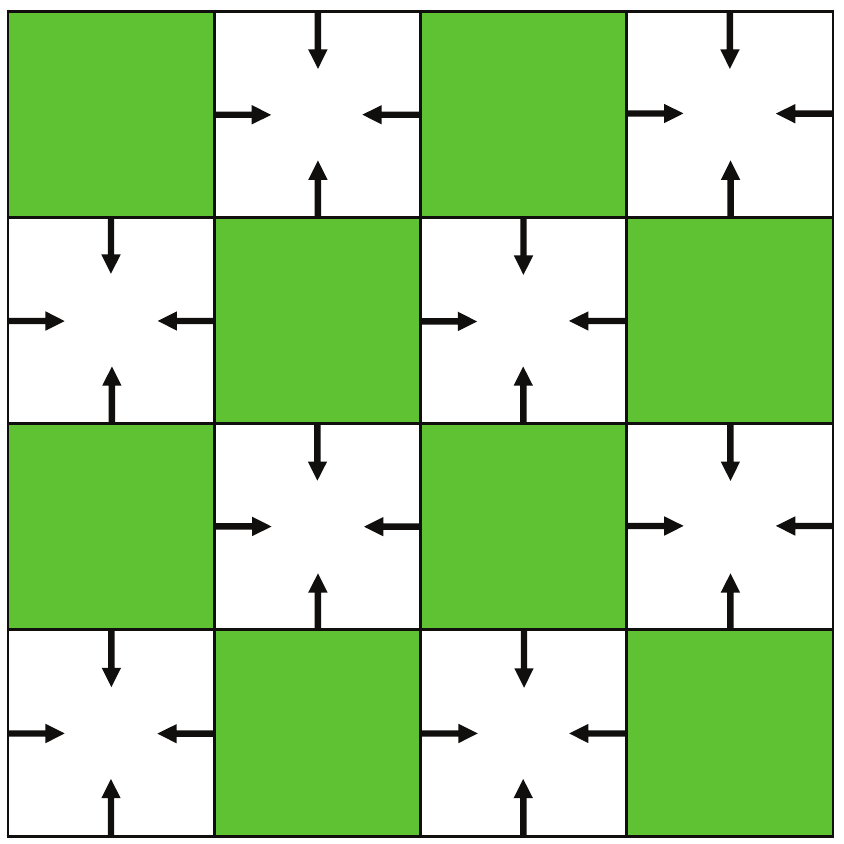}&
\includegraphics[width=0.28\linewidth]{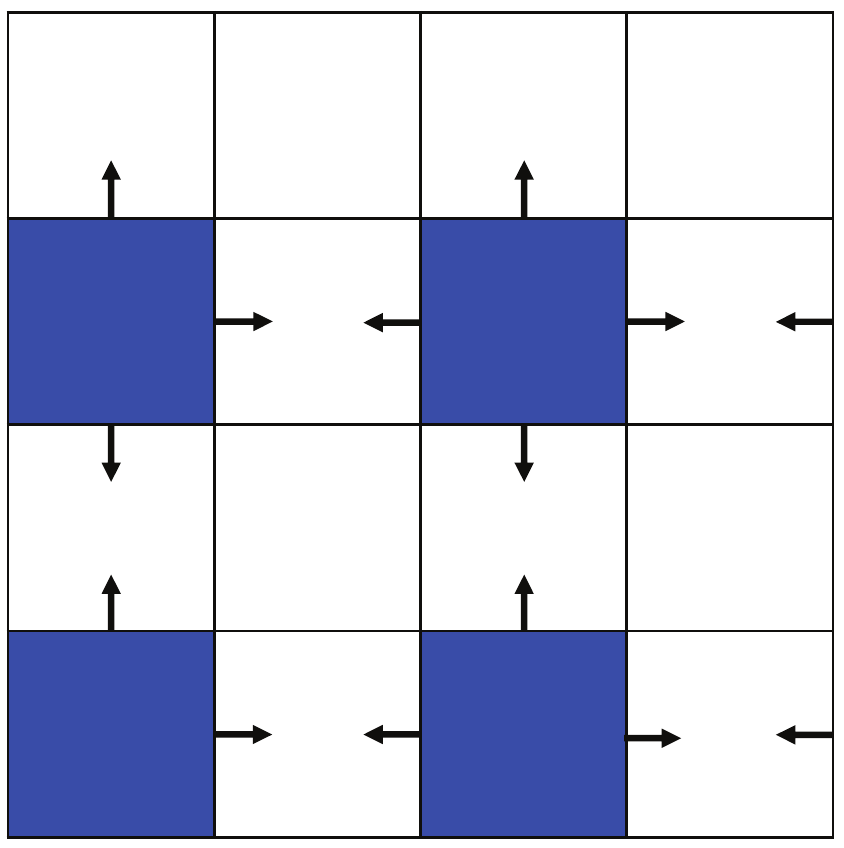}\\
\footnotesize{(a) Red Channel} &
\footnotesize{(b) Green Channel} &
\footnotesize{(c) Blue Channel} \\
\end{tabular}
\caption{Crosstalk in Bayer Color Atom. Each color pixel leak some of its charge to its horizontal and vertical neighbors. Amount of leakage is parametrized by the positive scalars $\delta_r$, $\delta_g$ and $\delta_b$.}
\label{fig:ctkaliasing}
\vspace{-2ex}
\end{figure}

The effect of crosstalk is more severe when the adjacent colors are different. For example, in Figure~\ref{fig:ctkaliasing}, the red and blue pixels are surround by 8 neighbors of different colors and the green pixels are surrounded by 4 neighbors of different colors. This is equivalent to saying that there is a red pixel having a value $1$ and is surrounded by pixels having the value $0$. Using similar argument, we can see that if the color atoms have more rapid changes of colors, then the resulting CFA is more susceptible to crosstalk.

We propose to quantify the variation of the color atoms (and hence crosstalk) is by means of measuring the total variation of the color atom. The total variation is a proxy of the complexity of the color filter array. A color filter array with high total variation means a more complicated pattern and so it is more susceptible to crosstalk. Our total variation is defined as follows.
\begin{definition}[Total Variation]
For a CFA defined by the color atoms $\vh_r$, $\vh_g$ and $\vh_b$, the weighted total variation is defined as
\begin{align}\label{eq:tv}
\textrm{TV}(\vx) \bydef \sum_{i\in\{r,g,b\}} \delta_i \|\mD \vh_i\|_1 = \sum_{i\in\{r,g,b\}} \delta_i \|\mD \mZ_r \vx\|_1
\end{align}
where $\mD\bydef[\mD_x,\mD_y]^T$ is an operator that computes the vertical and horizontal derivatives with circular boundary conditions, and $\delta_i$ is the leakage factor defined in the crosstalk kernel in Equation \eref{eq:crosstalkKernels}.
\end{definition}

To control the amount of variations in the CFA (so that we can limit the amount of crosstalk), we upper bound the total variation by a scalar $\textrm{TV}_{\mathrm{max}}$. This leads to the following constraint.
\begin{proposition}[Crosstalk Constraint]\label{prop:ctk}
The crosstalk is limited by upper-bounding the total variation metric $\mathrm{TV}(\vx)$:
\begin{equation}
\textrm{TV}(\vx) = \sum_{i\in\{r,g,b\}} \delta_i ||\mD \mZ_r \vx||_1 \leq \textrm{TV}_{\mathrm{max}}.
\end{equation}
\end{proposition}

Figure~\ref{fig:ex} shows two CFAs proposed in literature. The first one, proposed in \cite{Hao_Li_Lin_2011} is more robust to aliasing than the second one proposed in \cite{Anzagira_Fossum_2015}. This is because the chrominance channels are modulated at high frequencies which are far from baseband luminance. However, \cite{Anzagira_Fossum_2015} is more robust to crosstalk than \cite{Hao_Li_Lin_2011} because the color atom have less variation in colors. We can also see this in the total variation values: $0.206$ compared to $0.131$ at $(\delta_r,\delta_g,\delta_b)=(0.23,0.15,0.10)$. This trade-off constitutes a gap in literature: Color filter designs can improve robustness of either aliasing or crosstalk, but not for both. Our proposed design framework allows us to optimize them simultaneously.

Since the best value for $TV_{max}$ changes for different atom sizes, we choose it according to the best value in the state-of-the-art CFA atoms. For example, we choose $TV_{max}=0.131$ for $4\times4$ atoms since this is the lowest value achieved by Anzagira and Fossum $4\times4$ RGBCWY CFA \cite{Anzagira_Fossum_2015}. Other $4\times4$ CFAs have higher $TV$ values.

\begin{figure*}[!ht]
\centering
\begin{tabular}{cccccc}
\rotatebox{90}{\footnotesize{\hspace{8ex}\cite{Hao_Li_Lin_2011}}}&
\includegraphics[width=0.15\linewidth]{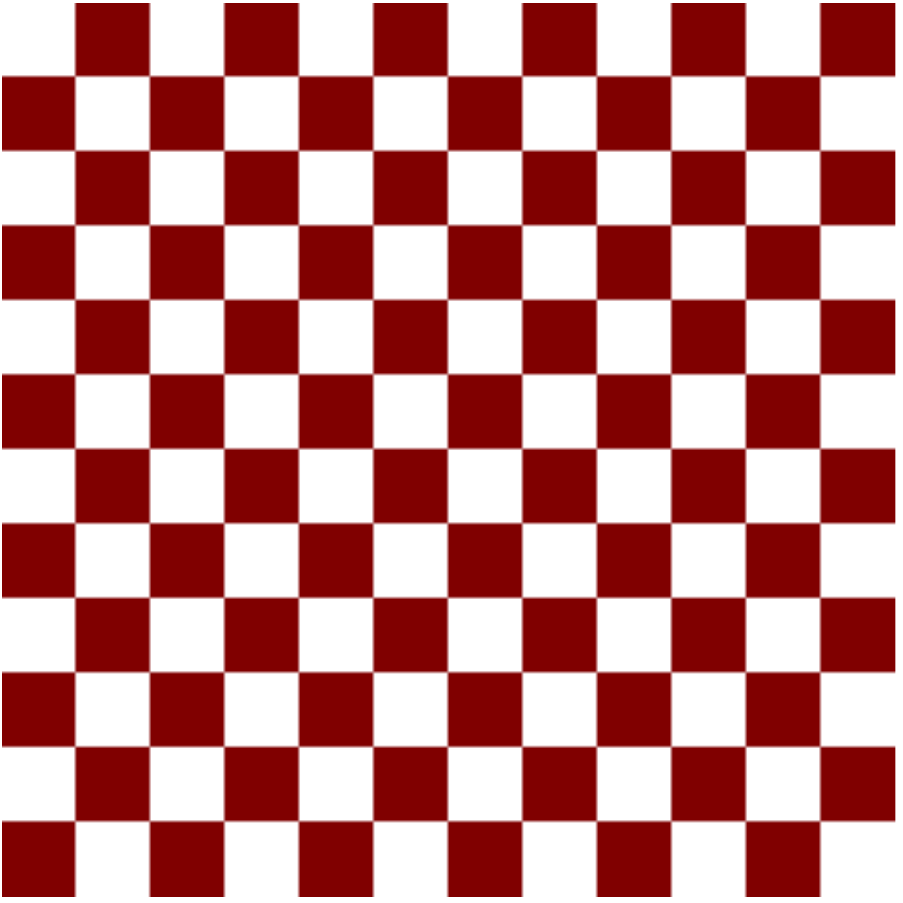}&
\includegraphics[width=0.15\linewidth]{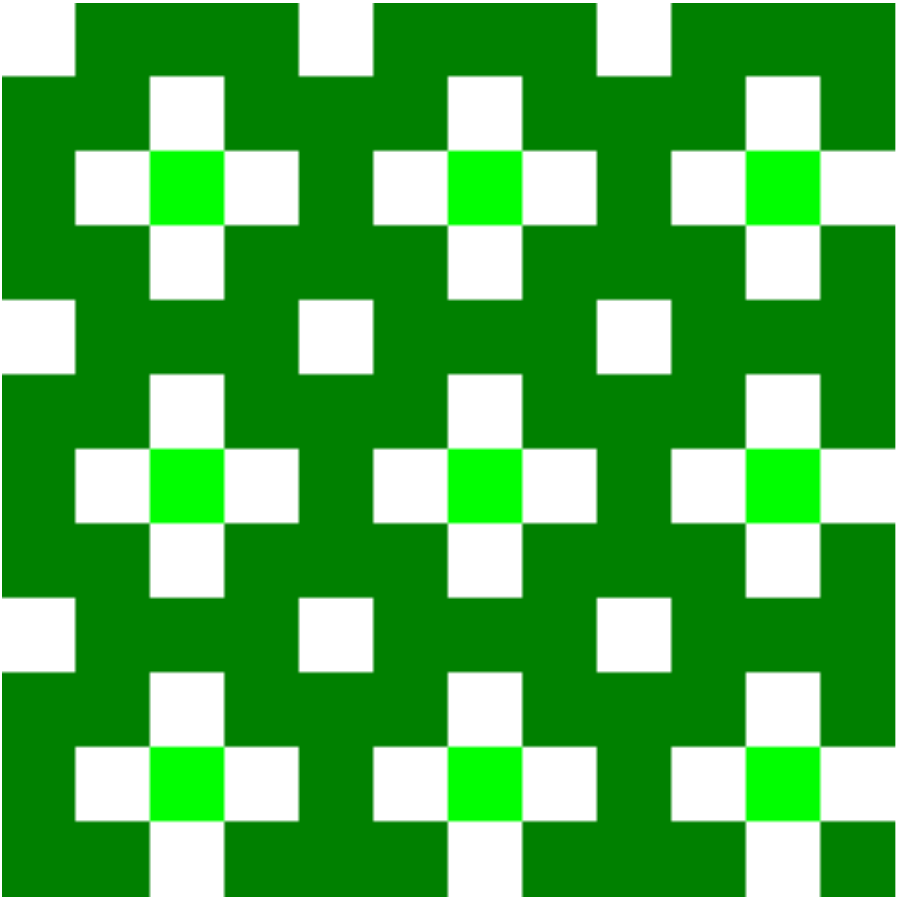}&
\includegraphics[width=0.15\linewidth]{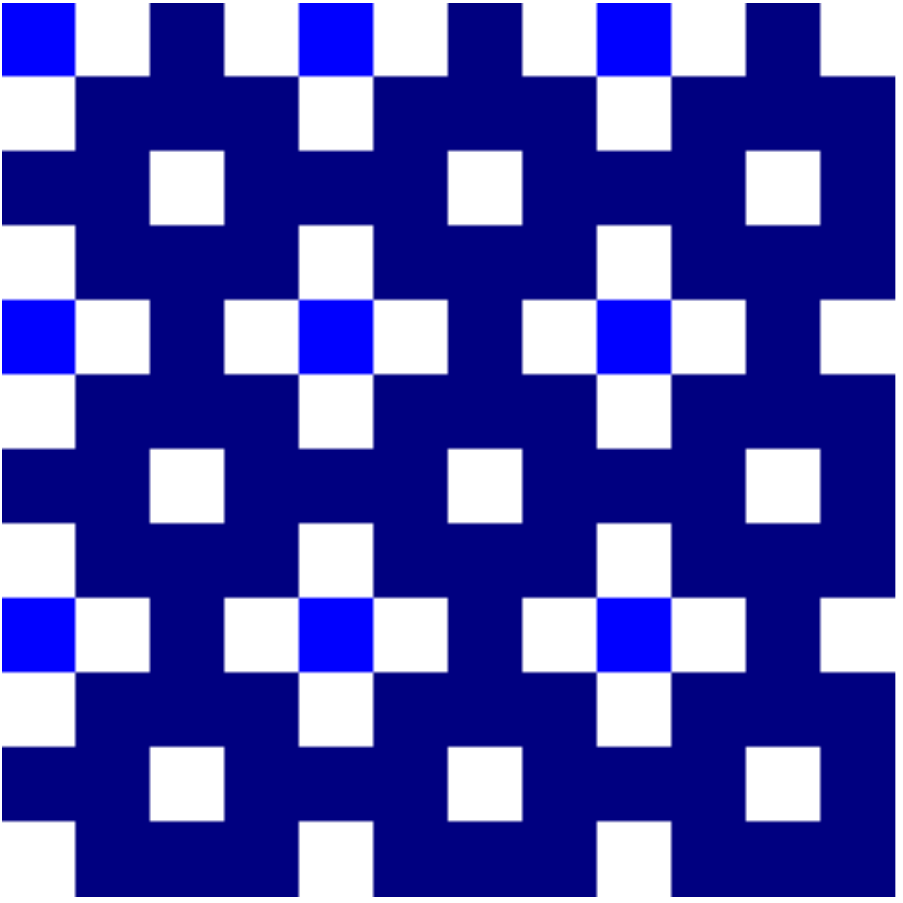}&
\includegraphics[width=0.15\linewidth]{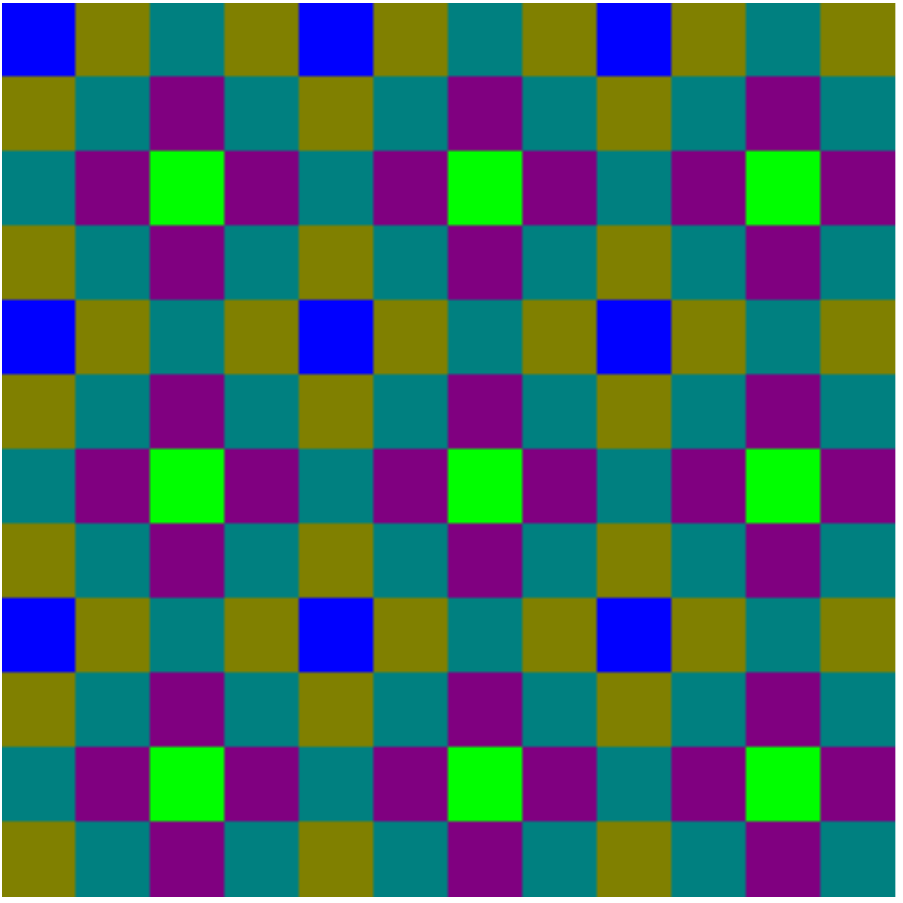}&
\includegraphics[width=0.225\linewidth]{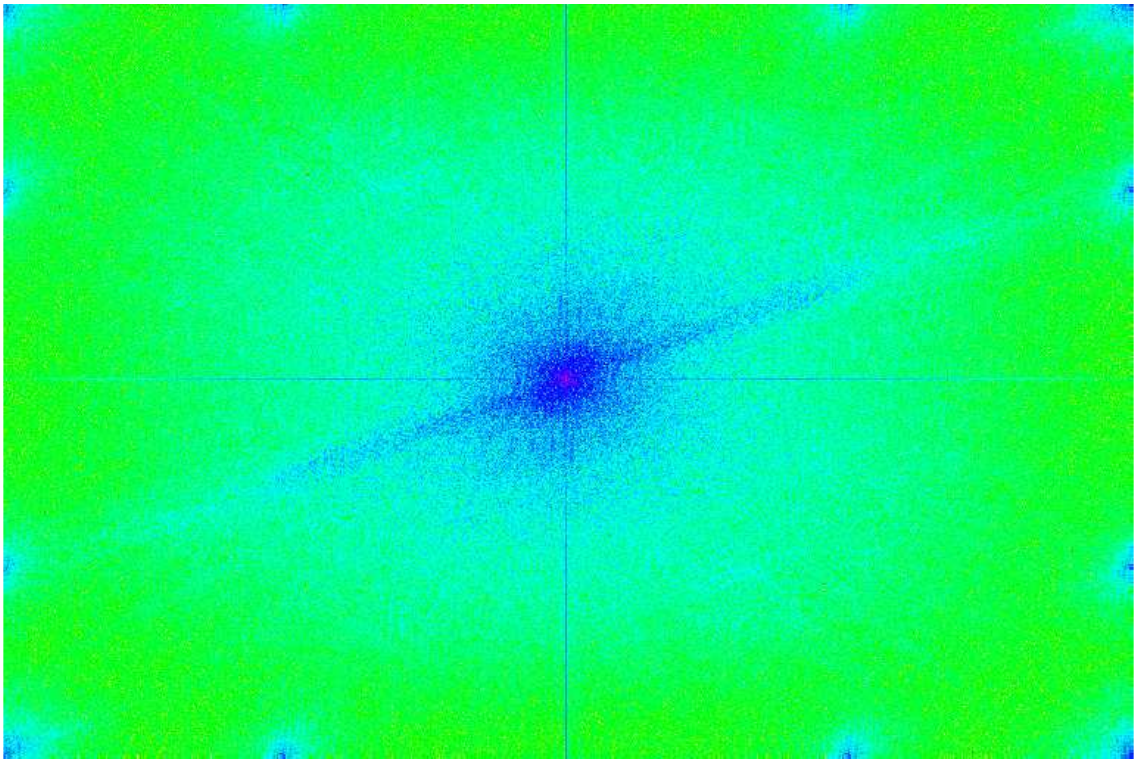}\\
\rotatebox{90}{\footnotesize{\hspace{3ex} RGBCWY \cite{Anzagira_Fossum_2015}}}&
\includegraphics[width=0.15\linewidth]{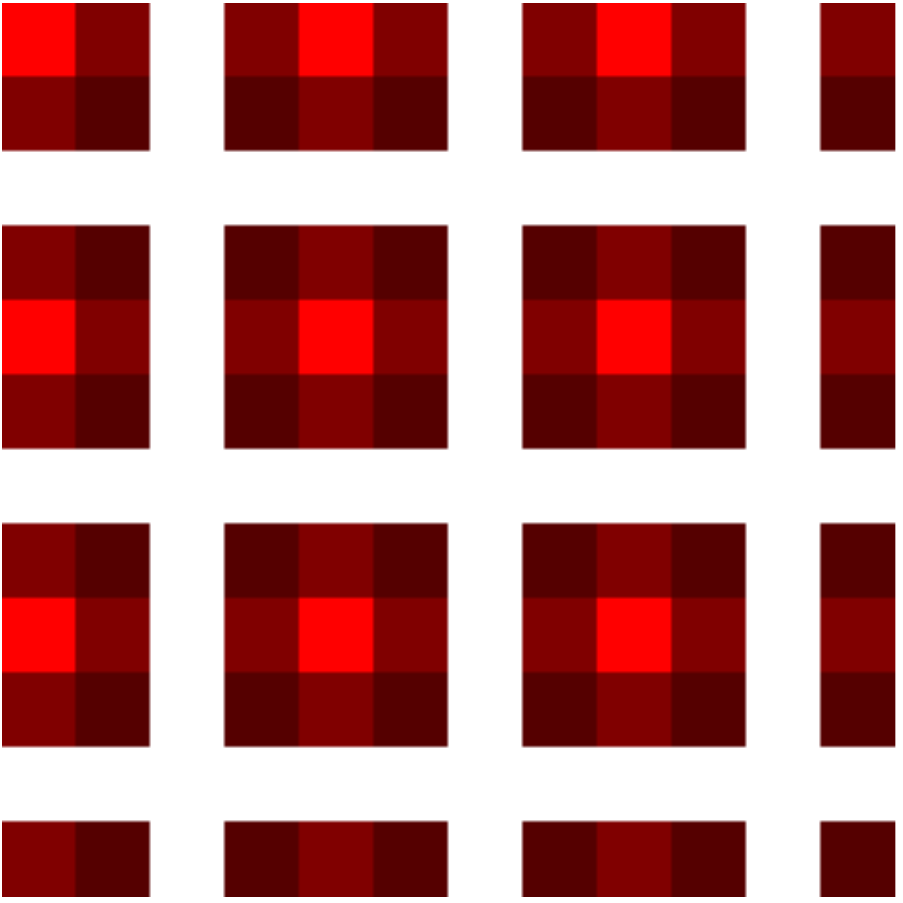}&
\includegraphics[width=0.15\linewidth]{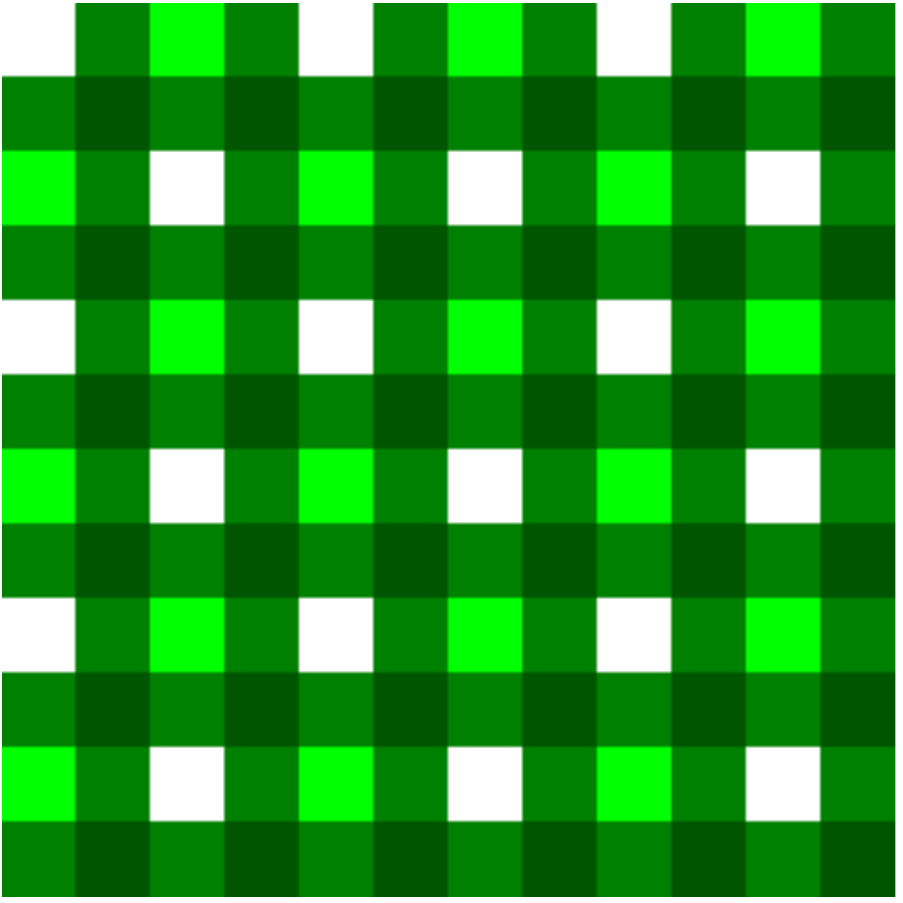}&
\includegraphics[width=0.15\linewidth]{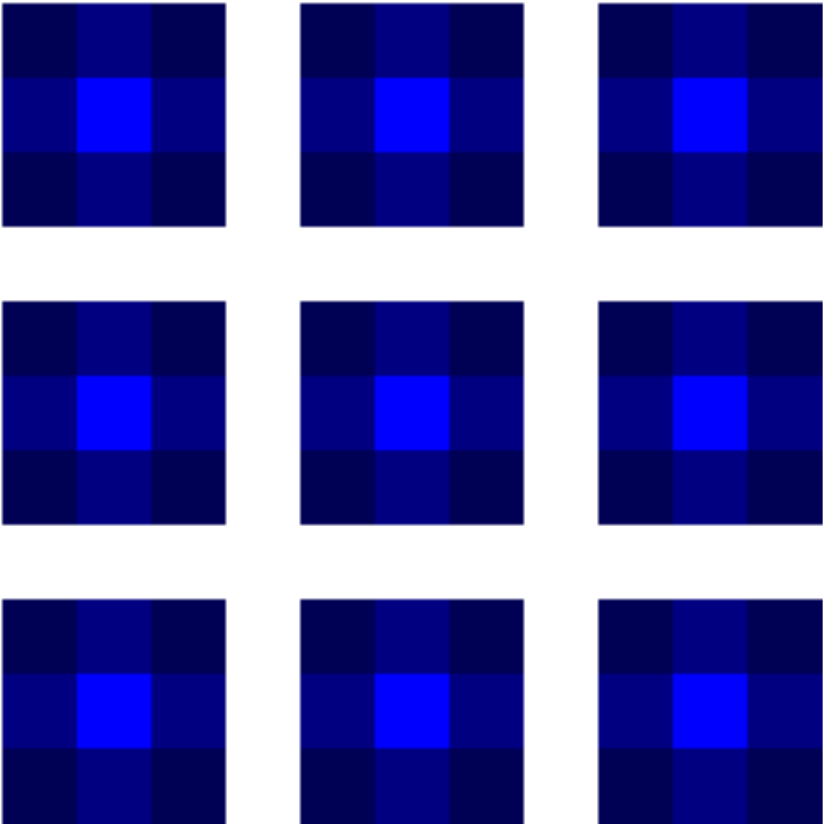}&
\includegraphics[width=0.15\linewidth]{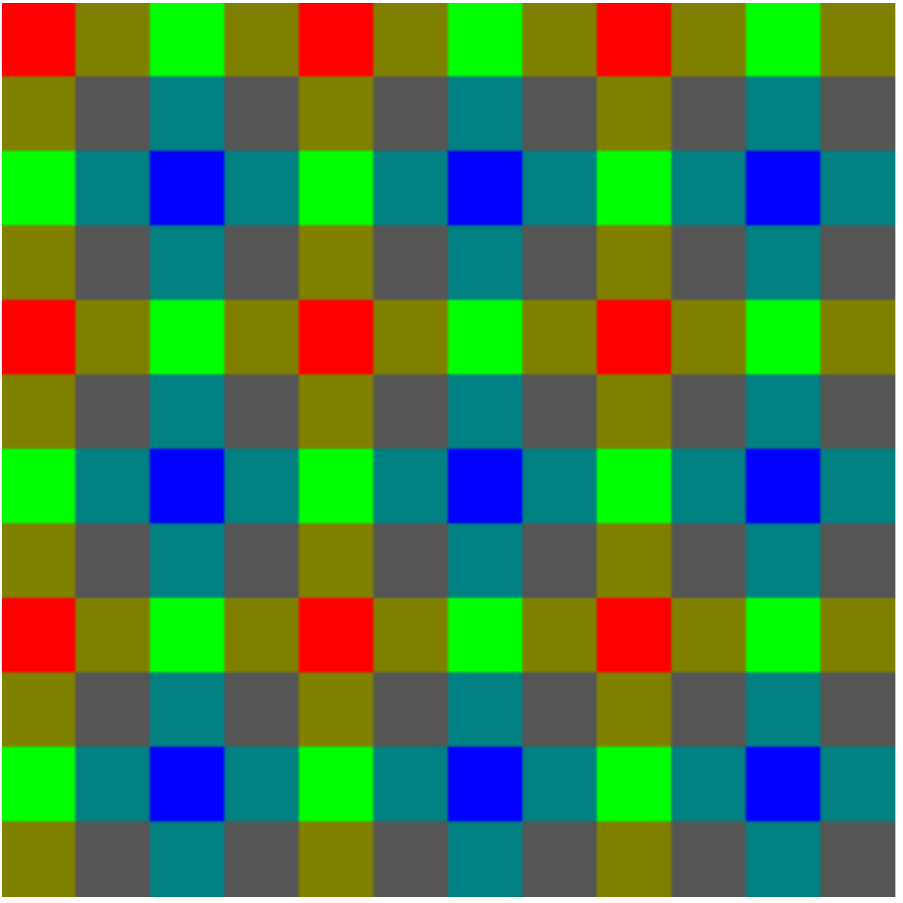}&
\includegraphics[width=0.225\linewidth]{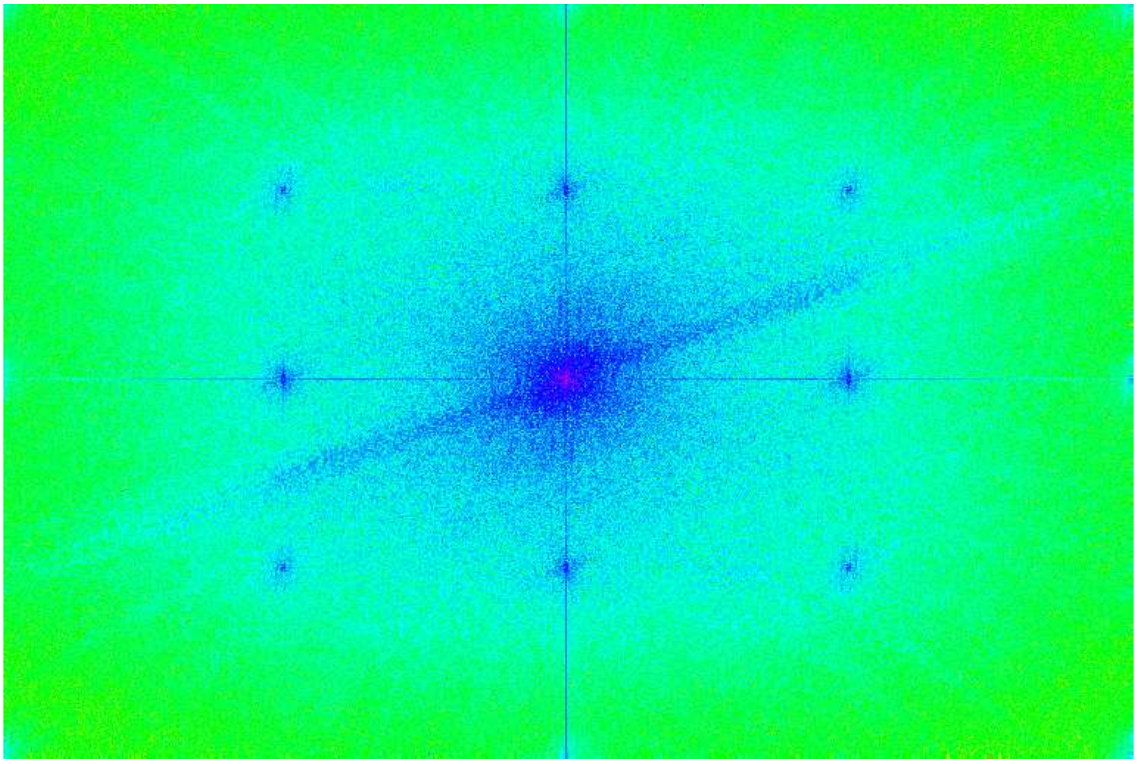}\\
&
\footnotesize{(a) Red Mask}&
\footnotesize{(b) Green Mask}&
\footnotesize{(c) Blue Mask}&
\footnotesize{(d) CFA Atom}&
\footnotesize{(e) Spectrum}\\
\end{tabular}
\caption{Examples of two CFAs. (Top row) Proposed in \cite{Hao_Li_Lin_2011}, this array has good aliasing properties, where chrominance channels are placed far away from luminance channel, but it has bad crosstalk properties: $\mathrm{TV}(\vx)=0.413$. (Bottom row) Proposed in \cite{Anzagira_Fossum_2015}, this array has good crosstalk properties $\mathrm{TV}(\vx)=0.263$, but it has bad aliasing properties.}
\label{fig:ex}
\vspace{-2ex}
\end{figure*}

\subsection{Orthogonality of Chrominance Channels}\label{subsec:ortho}
When designing a CFA, one should take into consideration of the complexity of the demosaicking process. Recall Equation \eref{eq:modulation} where we show that
\begin{align*}
\widetilde{\theta}(\vomega) =\sum_{i\in\{l,\alpha,\beta\}} \sum_{u=0}^{M-1} \sum_{v=0}^{N-1} \; \widetilde{h}_{i}(u,v)\;\widetilde{\textrm{im}}_i(\vomega-\vomega(u,v)).
\end{align*}
This is a modulation of the signal by a modulating frequency $\vomega(u,v)$. Therefore, to reconstruct the signal, one approach is to demodulate by shifting the channels to the baseband by multiplying pure sinusoids and then applying a lowpass filter \cite{Condat_2011}. Demodulation can be done efficiently if there is orthogonality between the channels. Following the literature, our optimization takes into account of two forms of orthogonality.
\begin{itemize}
\item Total Orthogonality \cite{Hao_Li_Lin_2011} and \cite{Hirakawa_Wolfe_2008}: The idea is to make one chroma channel zero and the other non-zero at any $(u,v)$. For example, $\widetilde{h}_\alpha(u,v)=0.9$ and $\widetilde{h}_\beta(u,v)=0$.
\item Quadrature Orthogonality \cite{Condat_2011}: The idea is to make one chroma channel real and the other imaginary at any $(u,v)$, i.e., the two channels are modulated by a frequency $\vomega(u,v)$ but in quadrature phase. Translating the spatial domain, this means that
    \begin{subequations}\label{eq:quad}
    \begin{align}
    h_{\alpha}(m,n) &= \gamma_c \sqrt{2} \cos \left(\vomega(u,v)^T \begin{bmatrix}
    m \\n \end{bmatrix}-\phi\right) \label{eq:cos}\\
    h_{\beta}(m,n) &= \gamma_c \sqrt{2} \sin \left(\vomega(u,v)^T \begin{bmatrix}
    m \\n \end{bmatrix}-\phi\right) \label{eq:sin}
    \end{align}
    \end{subequations}
where $m\in\{0,\ldots,M-1\}$, $n\in\{0,\ldots,N-1\}$, and $\phi$ is the phase angle. In this way, the two channels can be easily separated during the demosaicking process using the orthogonality of cosine and sine functions.
\end{itemize}

We formulate the orthogonality criteria as a penalty function that is a surrogate of both approaches.
\begin{definition} [Orthogonality Penalty]
\label{def: orthogonality}
For a CFA having chrominance channels with spectra $\widetilde{\vh}_\alpha(\vomega)$ and $\widetilde{\vh}_\beta(\vomega)$, the orthogonality penalty is defined as
\begin{align}\label{eq:pen}
\rho(\vh_\alpha,\vh_\beta) &\bydef \sum_{u=0}^{M} \sum_{v=0}^{N} \Big(|\Re{\widetilde{h}_\alpha(u,v)}|+|\Re{\widetilde{h}_\beta(u,v)}|\Big)\nonumber\\
&+\sum_{u=0}^{M} \sum_{v=0}^{N}\Big(|\Im{\widetilde{h}_\alpha(u,v)}|+|\Im{\widetilde{h}_\beta(u,v)}|\Big)
\end{align}
which can be written as a function in $\vx$ as follows
\begin{align}
\rho(\vx)&= \left(\|\Re{\mF \mZ_\alpha \vx}\|_1+\|\Re{\mF \mZ_\beta \vx}\|_1\right)\nonumber\\
& + \left(\|\Im{\mF \mZ_\alpha \vx}\|_1+\|\Im{\mF \mZ_\beta \vx}\|_1\right)
\end{align}
\end{definition}
Looking at the first summation in Equation \eref{eq:pen}, we notice that for every 2D frequency $(u,v)$, the term $|\Re{\widetilde{h}_\alpha(u,v)}|+|\Re{\widetilde{h}_\beta(u,v)}|$ is the $\ell_1$-norm of a 2-dimensional vector $[\Re{\widetilde{h}_\alpha(u,v)},\Re{\widetilde{h}_\beta(u,v)}]^T$. Therefore, minimizing this $\ell_1$-norm promotes either one of the components to zero (or both). Similar argument applies for the imaginary components in the second summation. As a result, the total variation can be regarded as a proxy to the orthogonality condition.

\subsection{Condition Number}\label{subsec:cond}
When designing a color filter array, one should also be aware of the simplicity of the demosaicking algorithm. Since the luminance/chrominance transformation, color filtering and crosstalk are all linear processes, we can represent them by an overall color acquisition matrix $\mA$. To demosaic the image, in principle we need to invert the $\mA$ matrix. To avoid the amplification of the estimation error of luminance and chrominance channels, the condition number of $\mA$ should be minimized for numerical stability. This metric was discussed in \cite{Hao_Li_Lin_2011}, but the authors considered the condition number of the luminance/chrominance transformation matrix $\mT$ only. In our case, we generalize this metric by taking the color filtering and crosstalk into account as well.

To represent the image acquisition in frequency domain as a linear process, we assume the crosstalk kernels for red, green and blue pixels are the same $g_r=g_g=g_b$. Define the following frequency domain variables:
\begin{equation}
\widetilde{\textbf{im}}_{rgb}=\begin{bmatrix}
\widetilde{\textbf{im}}_{r}^T \\ \widetilde{\textbf{im}}_{g}^T \\\widetilde{\textbf{im}}_{b}^T
\end{bmatrix}, \widetilde{H}=[\widetilde{\vh}_l,\widetilde{\vh}_\alpha,\widetilde{\vh}_\beta], \mbox{ and } \widetilde{\mG} = \mathrm{diag(\widetilde{\vg})}
\end{equation}
where  $\widetilde{\vg}\overset{\calF}{\leftarrow}\vg$ is the vectorized version of the $M\times N$ discrete Fourier transform of the crosstalk kernel $g$. Hence, the mosaicked image $\widetilde{\theta}$ can be written as
\begin{equation}
\widetilde{\vtheta} =  \widetilde{\mG} \widetilde{\mH} \mT \widetilde{\textbf{im}}_{rgb} = \mA \widetilde{\textbf{im}}_{rgb}
\end{equation}
where we define the color acquisition matrix as $\mA\bydef \widetilde{\mG} \widetilde{\mH} \mT$. Denote by $\kappa(\mA)$ the condition number of $\mA$, i.e.,
\begin{equation}
\kappa(\mA) = \mathrm{cond}(\mA) \in [1,\infty]
\end{equation}
Low values of $\kappa(\mA)$ imply stable demosaicking process that involves mild amplification of estimation errors in the luminance and chrominance components.


\section{Formulation of Optimal CFA Design Problem}
\label{sec:formulation}
Using the variables and constraints defined in the previous section, we present two different optimization formulations of the CFA design problem in this section: (i) A non-convex formulation that integrates all the above information into a single optimization, and (ii) convex relaxation that makes the problem more tractable.

\subsection{Non-Convex CFA Design}\label{subsec:nonconv}
The non-convex CFA optimization puts the objectives and constraints defined in the previous section into a single optimization problem. This gives us
\begin{align}\label{eq:nonconv1}
&\underset{\vx}{\mbox{maximize}} \quad  \gamma_c(\vx) + \lambda_l \gamma_l(\vx) - \lambda_\rho \rho(\vx)  \\
&\begin{array}{llll}
\hspace{-1.0ex}\mbox{subject to} & & & \\
& \vx \in[0,1]^{3L} & \mbox{(Realizability)} & \mbox{(a)}\\
& (\mI-\mbox{diag}(\ve_1)) \mF \mZ_l \, \vx = \vzero & \mbox{(Proposition~\ref{prop:uniformluma})} & \mbox{(b)}\\
& \mW \vx = \vzero &\mbox{(Proposition~\ref{prop:aliasing})} & \mbox{(d)} \\
& \textrm{TV}(\vx) \leq \textrm{TV}_{\textrm{max}} & \mbox{(Proposition~\ref{prop:ctk})} & \mbox{(e)}
\end{array}\nonumber
\end{align}
where $\lambda_l$ and $\lambda_\rho$ are the regularization parameters controlling the relative weights of the luminance sensitivity and the orthogonality penalty. The penalty function $\rho(\vx)$ is added to the objective with a negative sign so that it is minimized. By lower bounding $\gamma_c(\vx)$ with a constant $\tau$, we can rewrite Equation \eref{eq:nonconv1} as
\begin{align} \label{eq:nonconv2}
&\underset{\vx,\tau}{\mbox{maximize}} \quad  \tau  + \lambda_l \gamma_l(\vx) - \lambda_\rho \rho(\vx)  \\
&\begin{array}{llll}
\hspace{-1.0ex}\mbox{subject to} & & & \\
& \vx \in[0,1]^{3L} & \mbox{(Realizability)} & \mbox{(a)}\\
& (\mI-\mbox{diag}(\ve_1)) \mF \mZ_l \, \vx = \vzero & \mbox{(Proposition~\ref{prop:uniformluma})} & \mbox{(b)}\\
& \mW \vx = \vzero &\mbox{(Proposition~\ref{prop:aliasing})} & \mbox{(d)} \\
& \textrm{TV}(\vx) \leq \textrm{TV}_{\textrm{max}} & \mbox{(Proposition~\ref{prop:ctk})} & \mbox{(e)} \\
& \vx^T\mQ_\alpha \vx \geq \tau^2 & \mbox{({Proposition~\ref{prop:gammas}})}& \mbox{(f)}\\
& \vx^T\mQ_\beta \vx \geq \tau^2 & \mbox{({Proposition~\ref{prop:gammas}})} &  \mbox{(g)}
\end{array}\nonumber
\end{align}

In this optimization problem, the objective and constraints are convex except for \eref{eq:nonconv2}(f) and \eref{eq:nonconv2}(g). This is because these inequalities include convex quadratic form in the ``$\ge$'' side, where convexity comes from the fact that $\mQ_\alpha$ and $\mQ_\beta$ are positive semidefinite matrices. Hence, the optimization problem is non-convex.

\subsection{Solving the Optimization}
While problem \eref{eq:nonconv2} is non-convex, we can find a local minimum by successive convex approximations \cite{Opial_1967}. The idea of successive convex approximation is to replace the quadratic terms in the non-convex constrains \eref{eq:nonconv2}(f) and \eref{eq:nonconv2}(g) by first order approximations around the initial guess $\vx^{(0)}$. Since the quadratic form is convex, its first order approximation constitutes a lower bound. Hence, we are replacing the non-convex constraints \eref{eq:nonconv2}(f) and \eref{eq:nonconv2}(g) with convex but tighter constraints that limit the feasible set of $\vx$. The algorithm repeats until $\tau$ converges to a fixed-point, which is the final chrominance sensitivity.


\begin{algorithm}[!t]
	\caption{Successive Convex Approximations}
	\label{alg:nonconvex}
	\begin{algorithmic}
    \REQUIRE Initial guess $\vx^{(0)}$, $k=0$.
    \WHILE{$\gamma_c$ not converge}
    \STATE Replace the quadratic terms $\vx^T \mQ_\alpha \vx$ and $ \vx^T \mQ_\beta \vx$ in inequalities \eref{eq:nonconv2}(f-g) by their first order Taylor approximations around $\vx^{(k)}$:
		\begin{align*}
		\vx^T \mQ_\alpha \vx &\approx  \vx^{(k)T} \mQ_\alpha 		\vx^{(k)} + 2(\vx-\vx^{(k)})\mQ_\alpha \vx^{(k)} \geq {\tau^2}  \\
		 \vx^T \mQ_\beta \vx &\approx  \vx^{(k)T} \mQ_\beta 	\vx^{(k)} + 2(\vx-\vx^{(k)})\mQ_\beta \vx^{(k)}\geq {\tau^2}
		\end{align*}
	\STATE Solve the convex approximation of \eref{eq:nonconv2} to get $\gamma_c^{(k)}$
    \STATE $k = k+1$
	\ENDWHILE
    \RETURN $\vx$
	\end{algorithmic}
	\vspace{0ex}
\end{algorithm}

\begin{figure}[!t]
\centering
\includegraphics[width=0.8\linewidth]{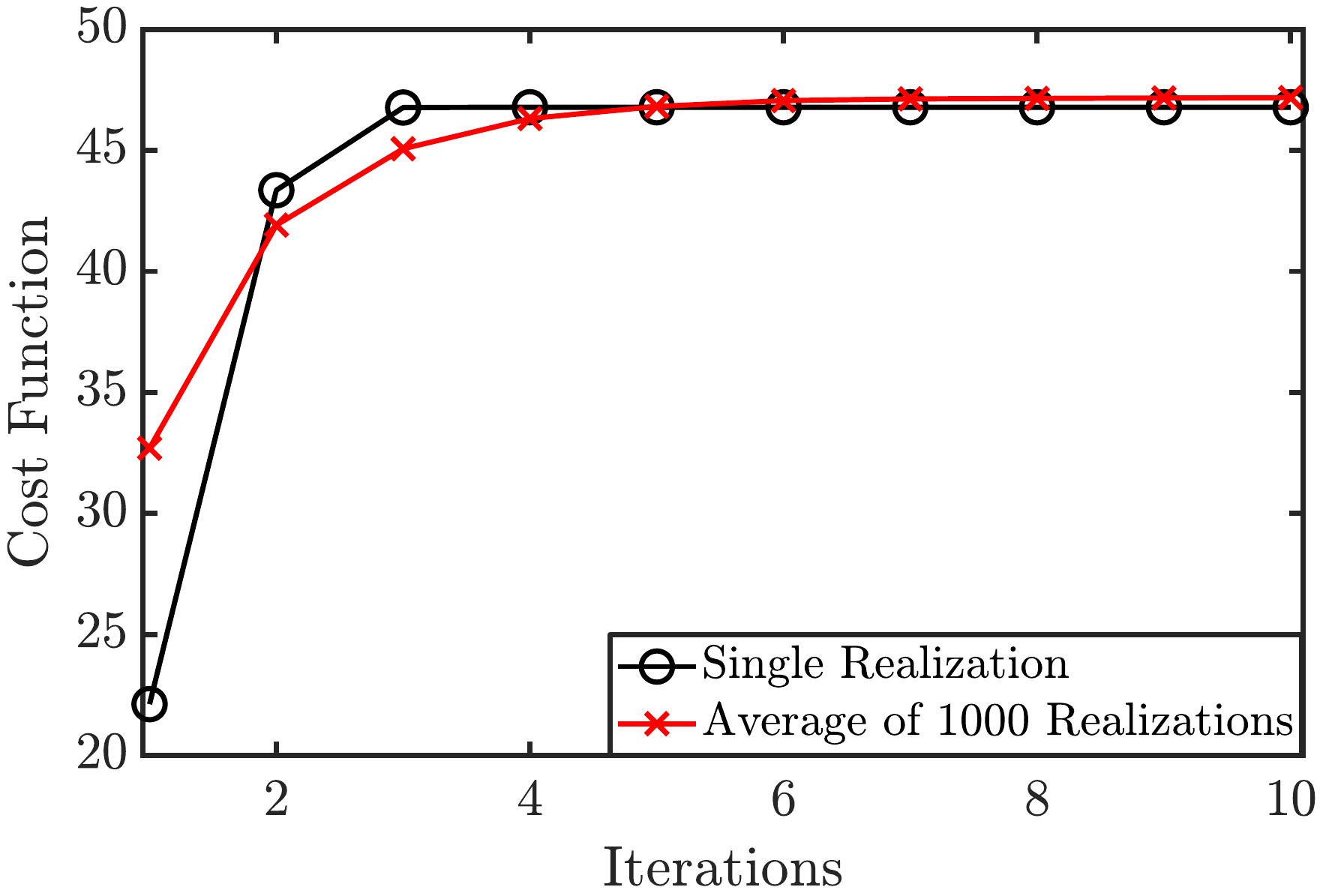}
\caption{Convergence of Algorithm 1 for $4\times4$ color filter design for single realization and average of 1000 realizations.}
\label{fig:conv_nonConv}
\vspace{-3ex}
\end{figure}

The overall algorithm is summarized in Algorithm~\ref{alg:nonconvex}. Figure~\ref{fig:conv_nonConv} shows the convergence of Algorithm~\ref{alg:nonconvex} for designing a $4\times 4$ color atom. We notice the monotonic increase of $\tau$ until it converges to a fixed point. Since the original problem is non-convex, solution to the problem could be a local minimum depending on how the initialization is done. In practice, we solve the problem for multiple instances with different randomly generated initial guesses which approximately cover the design space (e.g., using the Latin hypercube sampling \cite{Stein_1987}), and pick the best solution among them.

\subsection{Convex CFA Design}\label{subsec:conv}
The relaxation from non-convex to convex can be done by explicitly forcing part of the chrominance components to zero. Specifically, we modulate the chrominance channels on the same frequency $\vomega(u,v)=(\frac{2\pi u}{M},\frac{2\pi v}{N})$ using the quadrature orthogonality mentioned in Equation \eref{eq:quad}. In terms of $\vx$, these two equations can be written as:
\begin{equation}
\mZ_\alpha \vx = \gamma_c \vx_c, \quad  \mZ_\beta \vx = \gamma_c \vx_s
\end{equation}
where $\vx_c$ and $\vx_s$ are constant vectors that represent the vectorized version of the cosine and sine signals on the right hand side of Equations \eref{eq:cos} and \eref{eq:sin}, respectively, i.e.,
\begin{subequations}
\begin{align}
\vx_c &= \textrm{vec}\left\{\sqrt{2} \cos \left(\vomega(u,v)^T \begin{bmatrix} m \\n \end{bmatrix}-\phi\right)_{m=0,n=0}^{M-1,N-1}\right\}\\
\vx_s &= \textrm{vec}\left\{\sqrt{2} \sin \left(\vomega(u,v)^T \begin{bmatrix} m \\n \end{bmatrix}-\phi\right)_{m=0,n=0}^{M-1,N-1}\right\}
\end{align}
\end{subequations}
Since we explicitly choose the modulation frequencies of chrominance channels manually, we can drop the aliasing constraint in Proposition~\ref{prop:aliasing}. However, we still need the uniform luminance constraint in Propositions~\ref{prop:uniformluma}. Moreover, since the luminance and chrominance gains are adversarial, the objective of this formulation is to maximize their weighted sum. To this end, the problem is written as:
\begin{align} \label{eq:conv1}
&\underset{\vx,\gamma_c}{\mbox{maximize}} \quad  \gamma_c + \lambda_l \gamma_l(\vx)  \\
&\begin{array}{llll}
\hspace{-1.0ex}\mbox{subject to} & & & \\
& \vx \in[0,1]^{3L} & \mbox{(Realizability)} & \mbox{(a)} \\
& (\mI-\mbox{diag}(\ve_1)) \mF \mZ_l \, \vx = \vzero & \mbox{(Proposition~\ref{prop:uniformluma})} & \mbox{(b)}\\
& \textrm{TV}(\vx) \leq \textrm{TV}_{\textrm{max}} & \mbox{(Proposition~\ref{prop:ctk})} & \mbox{(d)} \\
& \mZ_\alpha \vx - \gamma_c \vx_c = 0 & & \mbox{(e)}\\
&\mZ_\beta \vx - \gamma_c \vx_s = 0 & &  \mbox{(f)}
\end{array}\nonumber
\end{align}
In our terminology, the optimization problem of \cite{Condat_2011} is obtained from Equation \eref{eq:conv1} by removing the crosstalk constraint \eref{eq:conv1}(d). Hence, our optimization limits the search space of the optimization in \cite{Condat_2011} to get CFAs that have acceptable crosstalk performance.

\begin{figure}[t]
\centering
\begin{tabular}{cc}
\includegraphics[width=0.45\linewidth]{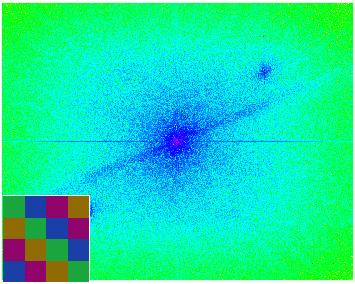}&
\includegraphics[width=0.45\linewidth]{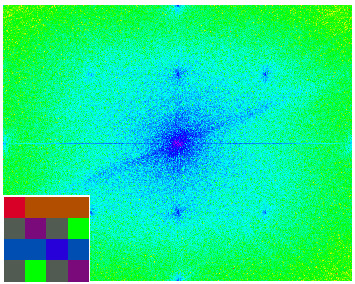}\\
\footnotesize{(a) Convex Formulation} &
\footnotesize{(b) Non-Convex Formulation} \\
\footnotesize{$\gamma_c(\vx)=0.08$}&
\footnotesize{$\gamma_c(\vx)=0.12$}
\end{tabular}
\caption{$4\times 4$ color atoms and corresponding spectra obtained using convex and non-convex formulations. Spectra are obtained from mosaicking the ``Bikes\rq\rq{} image in Kodak color dataset by the color atoms. Both have the same luminance sensitivity $\gamma_l(\vx)=0.577$ and same Total variation $TV(\vx)=0.131$.}
\label{fig:convVsNonConv}
\vspace{-3ex}
\end{figure}

Figure~\ref{fig:convVsNonConv} shows two color atoms obtained using the convex and non-convex formalizations. In the convex formulation, we select the modulation frequency to be $\omega_0=[\pi,\pi]$ and the phase that maximizes $\gamma_c$ at this frequency is found to be $\phi=\pi/12$. Then, we solve the problem to get $(\gamma_l,\gamma_c,TV)=(0.573,0.08,0.263)$. As for the non-convex formulation, we let the optimization to choose modulation frequencies subject to crosstalk and aliasing constraints. Solving the non-convex formulation yields $(\gamma_l,\gamma_c,TV)=(0.573,0.09,0.263)$. We notice that the non-convex formulation achieves higher chrominance sensitivity because of its flexibility in choosing the modulation frequencies.

\section{Universal Demosaicking} \label{sec:Demo}
In this section, we present a universal demosaicking algorithm which can be used to all CFAs presented in this paper. Our algorithm comprises two main parts: (i) a demosaicking step to remove the color filtering effect (Section~\ref{subsec:freqSelection}) and (ii) a color correction step to mitigate the crosstalk effect (Section~\ref{subsec:colorcorr}).


\subsection{Special Consideration for QIS.}
Before we talk about the demosaicking algorithm, we should briefly discuss the photon statistics of QIS. In CMOS, the measured voltage can be modeled as a nominal value corrupted by i.i.d. Gaussian noise. For single-bit QIS, previous work showed that the measured photon counts follow a truncated Poisson process \cite{Chan_Elgendy_Wang_2016}. When averaging over a number of temporal frames, the truncated Poisson becomes a Binomial \cite{Elgendy_Chan_2018}. If the photon count is sufficiently high, this binomial will approximately approach a Gaussian. Applying the law of large numbers on the distribution of $\mB$ in Equation \eref{eq:prob_Bm}, the average is
\begin{equation*}
\frac{1}{T}\sum_{t=0}^{T-1}\; b_{j,t} \overset{a.s.}{\longrightarrow} \E[B_j]=1-\Psi_q(\theta_j),
\end{equation*}
and the maximum-likelihood estimate of the signal is
\begin{equation*}
\theta_j = \Psi_q^{-1} \left(1-\frac{1}{T}\sum_{t=1}^T\; b_{j,t}\right)
\end{equation*}
As discussed in \cite{Elgendy_Chan_2018}, we can regard this equation as a tone-mapping of the photon counts. We regard $\theta_j$ as the $j$-th pixel of the mosaicked image generated by the CFA. The goal of demosaicing is to reconstruct a color image from $\theta_j$.


\subsection{Demosaicking by Frequency Selection}
\label{subsec:freqSelection}
Our demosaicking algorithm is based on frequency selection \cite{Alleysson_Susstrunk_Herault_2005}. It generalizes \cite{Dubois_2005} as it works for any CFA as long as it satisfies the orthogonality constraints in Section~\ref{subsec:ortho} 

\begin{figure}[t]
\centering
\includegraphics[width=1\linewidth]{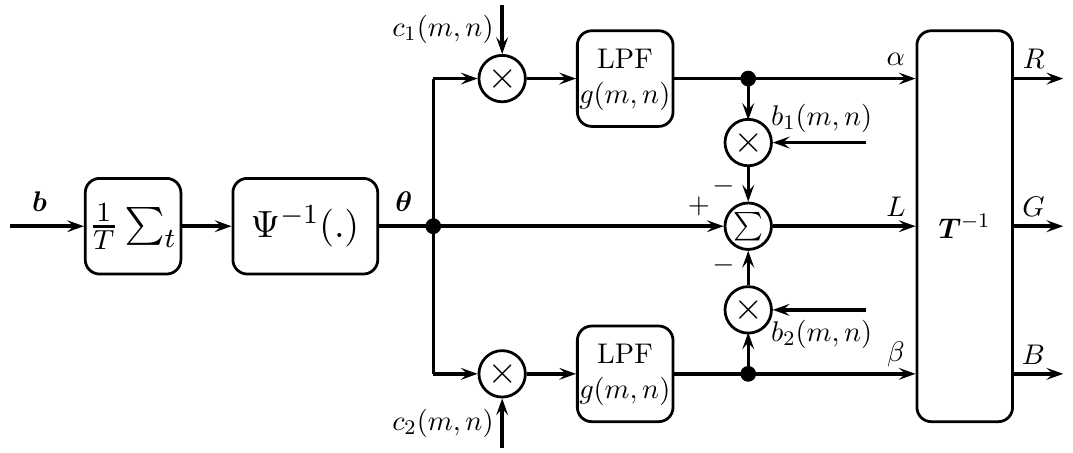}
\caption{Illustration of Algorithm~\ref{alg:freqSelection} of demosaicking by frequency selection for a special case of a CFA that has strictly one replica of the $\alpha$ and $\beta$ chrominance channels. Variable on the figure are defined in Algorithm~\ref{alg:freqSelection}.}
\label{fig:demosFreqSel}
\vspace{-0ex}
\end{figure}

The key idea of the algorithm is to shift every chrominance channel to the baseband by multiplying with its carrier, then use a low-pass filter to reconstruct it. For chrominance components that are replicated over distinct carriers, we combine them by simple averaging. After obtaining the $\alpha$ and $\beta$ chrominance channels, they are re-modulated to their original positions and subtracted from the mosaicked image to obtain the luminance channel. This process is summarized in Algorithm~\ref{alg:freqSelection} for a special case of a CFA that has strictly one replica of the $\alpha$ and $\beta$ chrominance channels. It is also illustrated by Figure~\ref{fig:demosFreqSel}. Extension of the algorithm to the general case is straightforward.

To apply Algorithm~\ref{alg:freqSelection} on CFAs proposed in \cite{Hao_Li_Lin_2011}, \cite{Anzagira_Fossum_2015} and \cite{Cheng_Siddiqui_Luo_2015}, they must satisfy the orthogonality constraints in Section~\ref{subsec:aliasing}. However, this is not satisfied with our choice of the luminance/chrominance basis defined by $\mT$ in Equation \eref{eq:basis}.  Hence, we use for every CFA the transformation matrix $\mT$ that makes its luminance and chrominance channel orthogonal. To ensure fairness, we normalize the matrix rows to unity so that all luminance and chrominance have the same noise power. The transformation matrices are provided in the supplementary.
\begin{algorithm}[!t]
	\caption{Demosaicking by Frequency Selection}
	\label{alg:freqSelection}
	\begin{algorithmic}
    \REQUIRE The image $\vtheta$ which is mosaicked by a CFA of size $M\times N$ as defined in Equation \eref{eq:mosaickImage}, a luminance/chrominance transformation matrix $\mT$, a low-pass filter $g$, a scalar $K\bydef MN$ and a scalar $r=\begin{cases}2 & \mbox{if } \vomega=(\pi,\pi)\\ 1 & \mbox{otherwise} \end{cases}$.
    \ENSURE $\alpha$ and $\beta$ chrominance channels are modulated on carriers $\vomega(u_1,v_1)$ and $\vomega(u_2,v_2)$, respectively.
    \STATE 1) Reconstruct the $\alpha$ chrominance channel
    \begin{equation*}
	\alpha(m,n) = \left(\theta(m,n)c_1(m,n)\right) \circledast g(m,n)
	\end{equation*}
where $$c_1(m,n)=\frac{K}{|a_1|}\cos\left(\vomega(u_1,v_1)^T\begin{bmatrix}m\\n\end{bmatrix}+\angle a_1\right)$$ and $a_1=\widetilde{h}_\alpha(u_1,v_1)$
\STATE 2) Reconstruct the $\beta$ chrominance channel
	\begin{equation*}
	\beta(m,n) = \left(\theta(m,n)c_2(m,n)\right)\circledast g(m,n)
	\end{equation*}
where $$c_2(m,n)=\frac{K}{|a_2|}\cos\left(\vomega(u_2,v_2)^T\begin{bmatrix}m\\n\end{bmatrix}+\angle a_2\right)$$ and $a_2=\widetilde{h}_\beta(u_2,v_2)$
\STATE 3) Reconstruct the luminance channel 
\begin{equation*}
L(m,n) = \theta(m,n) - \alpha(m,n) b_1(m,n) - \beta(m,n) b_2(m,n)
\end{equation*}
where $$b_1(m,n)=\frac{2|a_1|^2}{rK^2}c_1(m,n) \mbox{ and } b_2(m,n)=\frac{2|a_2|^2}{rK^2}c_2(m,n)$$
\RETURN $[\mR,\mG,\mB]^T =  \mT^{-1}[\mL,\valpha,\vbeta]^T$
\end{algorithmic}
\end{algorithm}

\subsection{Color Correction}\label{subsec:colorcorr}
The demosaicking algorithm in Algorithm~\ref{alg:freqSelection} does not take into account of the crosstalk effect. Like most of the mainstream image and signal processing (ISP) pipelines, we reduce the cross-talk via a color correction step.
\begin{figure}[t]
\centering
\begin{tabular}{cc}
\hspace{-2.5ex}\includegraphics[width=0.48\linewidth]{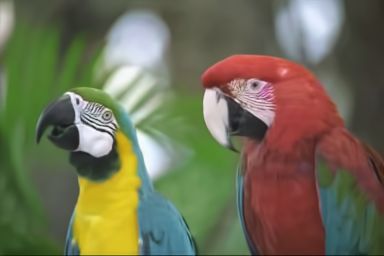}&
\hspace{-2ex}\includegraphics[width=0.48\linewidth]{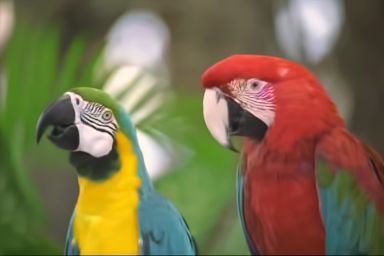}\\
\footnotesize{Before Color Correction}&
\footnotesize{After Color Correction}
\end{tabular}
\caption{Effect of color correction on retaining vivid image colors.}
\label{fig:colorcorrection}
\vspace{-2ex}
\end{figure}

Given the demosaicked color pixel $\widehat{\textbf{im}}(m,n)$, the color correction multiplies $\widehat{\textbf{im}}(m,n)$ by a $3 \times 3$ matrix $\mM$ such that $\mM \widehat{\textbf{im}}(m,n)$ is the color-corrected pixel. The matrix $\mM$ is learned by comparing a measured color pixel to a known color chart value. Mathematically, suppose we have $K$ measured color pixels forming a $3 \times K$ matrix $\mQ_{\textrm{False}}$, and a corresponding true color values forming another $3 \times K$ matrix $\mQ_{\textrm{GT}}$, we can estimate $\mM$ by solving
\begin{align}\label{eq:lincolorcorr}
&\mM = \arg \min_{\mM} \epsilon_c(\mM)
\end{align}
where $\epsilon_c(\mM)=\mathrm{Tr}\left\{\left(\mM\mQ_{\textrm{False}}-\mQ_{\textrm{GT}}\right)^T\left(\mM\mQ_{\textrm{False}}-\mQ_{\textrm{GT}}\right)\right\}$ is the squared color error.  To minimize the noise amplification, it is advised to add regularization when estimating $\mM$ \cite{Peng_Tinku_1999}. Since a standard color chart comprises $24$ color patches, we can estimate the noise by computing the norm of covariance matrix of every color patch, and get the average value over the $24$ color patches. Hence, the optimization problem is rewritten as
\begin{align}\label{eq:lincolorcorrNoise}
&\mM = \arg \min_{\mM} \epsilon_c(\mM) + \mu \sum_{i=1}^{24} \, ||\textrm{Cov}(\mM\mQ^{(i)}_\textrm{False})||_2^2
\end{align}
where $\mQ^{(i)}_\textrm{False}$ represents the pixels of the $i$th color patch, and $\mu$ is a positive scalar that controls the noise amplification effect. By varying $\mu$, we can draw a tradeoff curve between color reproduction accuracy and noise amplification.

Figure~\ref{fig:colorcorrection} shows reconstructed images before and after color correction. In this figure, the crosstalk parameters are $(\delta_r,\delta_r,\delta_r)=(0.23, 0.15, 0.1)$. We can see the effect of color correction in the more saturated red and yellow feathers and in the green leaves in the background.

\begin{table*}[t]
\centering
\caption{CFA parameters and Reconstruction quality measured by YSNR and SMI metrics on Macbeth ColorChecker and average CPSNR on DIV2K evaluation dataset. An arrow is placed after each metric to show whether higher or lower is better}
\begin{tabular}{c|c|ccccc|cc|cc|cc}
& &  \multicolumn{5}{c}{Experiment 1} & \multicolumn{4}{|c}{Experiment 2} &  \multicolumn{2}{|c}{Experiment 3} \\
\hline
\multirow{ 2}{*}{Size} & \multirow{ 2}{*}{CFA Pattern}   & \multicolumn{5}{c}{CFA Parameters} & \multicolumn{2}{|c}{YPSNR (dB) $\uparrow$} & \multicolumn{2}{|c}{SMI (\%) $\uparrow$} & \multicolumn{2}{|c}{CPSNR (dB) $\uparrow$} \\
\cline{3-13}
 &  & $\gamma_l \uparrow$ &  $\gamma_c \uparrow$ & TV $\downarrow$ & $J_{l} \downarrow$ & $\kappa(\mA)$ $\downarrow$ & w/o Ctk & w/ Ctk & w/o Ctk & w/ Ctk & w/o Ctk & w/ Ctk \\
\hline
\hline
\multirow{ 4}{*}{\rotatebox[origin=c]{0}{$4\times 4$}}
& RGBCY \cite{Anzagira_Fossum_2015}                 & 0.679 &   \textbf{0.125} & 0.156 &   6.53&2.376&23.24 &  22.15 & 93.65 &   92.90&30.65&30.63\\
& RGBCWY \cite{Anzagira_Fossum_2015}              & 0.597 &   0.102 & \textbf{0.131} & 2.67& 2.143& 23.57 & 22.45 & 93.71 &  93.10&30.75&30.74\\
& Hao et al. \cite{Hao_Li_Lin_2011}                      & 0.586 &   0.094 & 0.206   & \textbf{1.00}& 2.305& 24.13&   18.50 & \textbf{94.34} & 91.31&\textbf{31.34}&30.83\\
& Ours & 0.577                                               &   0.121 & \textbf{0.131} &  2.86& \textbf{1.795} &  \textbf{24.16} & \textbf{23.17} & 94.07 & \textbf{93.47}&31.10&\textbf{31.12}\\
\hline
\multirow{ 2}{*}{\rotatebox[origin=c]{0}{$3\times 3$}}
& Biay-Cheng et al. \cite{Cheng_Siddiqui_Luo_2015} & 0.612 &   \textbf{0.167} & 0.175    &\textbf{1.65} & 1.892 & 23.92 &  22.04 & 94.44 & 93.88&31.07&31.14\\
& Ours &	 0.577 &   0.160& 0.175 & 1.74& \textbf{1.561}& \textbf{24.38} &  \textbf{23.64} & \textbf{94.49}  &\textbf{94.36}&\textbf{31.24}&\textbf{31.37}\\
\hline
\multirow{ 2}{*}{\rotatebox[origin=c]{0}{$3\times 2$}}
& Condat \cite{Condat_2011}  & 0.866 &   0.250 & 0.317 &   \textbf{1.16} &1.915 & \textbf{25.33} &  21.68 & \textbf{94.71} & \textbf{93.87}&\textbf{31.30}&31.17\\
& Ours   &  0.866 &   0.250 & \textbf{0.212} &   1.40 & \textbf{1.837}  &25.28 &  \textbf{24.30} & 94.25 & 93.77&31.27&\textbf{31.32}\\
\hline
\multirow{ 2}{*}{\rotatebox[origin=c]{0}{$4\times 2$}}
& Hirakawa-Wolfe \cite{Hirakawa_Wolfe_2008} & 0.866 &   0.125 & 0.275   & \textbf{0.98} &2.767 & \textbf{25.38} &  21.22 & \textbf{94.57} & 93.58&\textbf{31.42}&31.08\\
& Ours &	 0.866 &   \textbf{0.187} & \textbf{0.256} &   1.27& \textbf{2.166} & 24.89 &  \textbf{22.34} & 94.36 &  \textbf{94.03}&31.40&\textbf{31.31}\\
\hline
\multirow{ 2}{*}{\rotatebox[origin=c]{0}{$7\times 7$}}
& Bai et al. \cite{Bai_Li_Lin_2016} & 0.577 &  0.091 & 0.204   & \textbf{1.87} & 2.242& 24.26 & 19.68 & \textbf{94.56} & 92.94&\textbf{31.41}&31.09\\
& Ours &	 0.577 &   \textbf{0.238} & 0.204 &   1.45& \textbf{1.294} & \textbf{24.86} & \textbf{23.85} & 94.45 & \textbf{94.35}&31.01&\textbf{31.20}\\
\hline
\end{tabular}
\label{tab:PSNR}
\vspace{-2ex}
\end{table*}

\section{Experimental Evaluation} \label{subsec:results}
In this section, we present CFAs obtained using our optimization framework in Section~\ref{subsec:sol}. Afterwards, we evaluate the performance of different CFAs using Algorithm~\ref{alg:freqSelection}. First, we simulate in Section~\ref{subsec:macbeth} the reconstruction performance of a Macbeth ColorChecker under D65 illuminant. Second, we show in Section~\ref{subsec:imRec} a quantitative and qualitative comparison of the reconstruction performance of all CFAs on  DIV2K evaluation dataset \cite{Agustsson_Timofte_2017}. This dataset comprises 100 color images of resolution $2040\times1356$. We downsample them to $1020\times678$ for our experiments. In all experiments, QIS parameters are chosen as $(q,\eta,T)=(1,2,1000)$, where the value of $T$ is set as a high value to minimize the noise power.

\subsection{Experiment 1: Proposed Solutions of CFA Design Problem}\label{subsec:sol}

We focus on the non-convex formulation \eref{eq:nonconv2} since it is more flexible than the convex formulation \eref{eq:conv1}. We set the parameters of formulation \eref{eq:nonconv2} as
$\lambda_l=0.1$ and $\lambda_\rho=0.02$. We run multiple instances of Algorithm~\ref{alg:nonconvex} ($2000$ instances) using different random initializations for the color atoms $\vx^{(0)}$. Then, we pick the solution with the highest chrominance sensitivity. To ensure that the initial guess spans the feasible set of $\vx$, we use uniform Latin hypercube sampling of the domain $[0,1]^{3L}$.

\begin{figure*}[t]
\centering
\begin{tabular}{cccccc}
\multicolumn{2}{c}{\hspace{-3.0ex}$\overset{4\times4}{\overbrace{\hspace{12em}}}$}&
\hspace{-1.8ex}$\overset{3\times3}{\overbrace{\hspace{6em}}}$ &
\hspace{-2.2ex}$\overset{3\times2}{\overbrace{\hspace{6em}}}$&
\hspace{-2.2ex}$\overset{4\times2}{\overbrace{\hspace{6em}}}$&
\hspace{-2.2ex}$\overset{7\times7}{\overbrace{\hspace{6em}}}$\\
\hspace{-1.9ex}\includegraphics[width=0.16\linewidth]{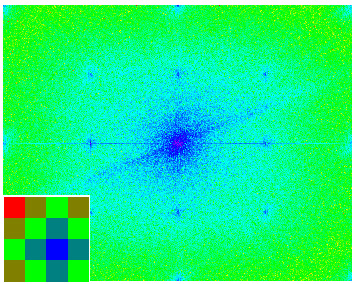}&
\hspace{-2.3ex}\includegraphics[width=0.16\linewidth]{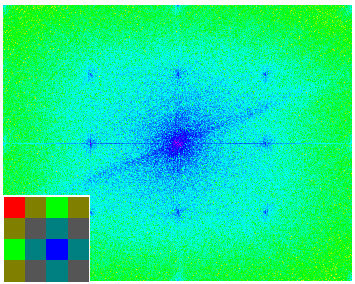}&
\hspace{-2.3ex}\includegraphics[width=0.16\linewidth]{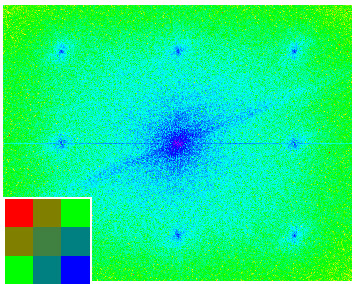}&
\hspace{-2.3ex}\includegraphics[width=0.16\linewidth]{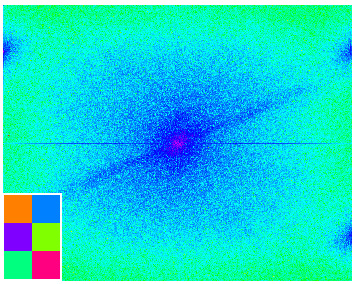}&
\hspace{-2.3ex}\includegraphics[width=0.16\linewidth]{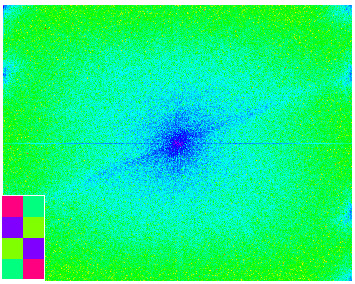}&
\hspace{-2.3ex}\includegraphics[width=0.16\linewidth]{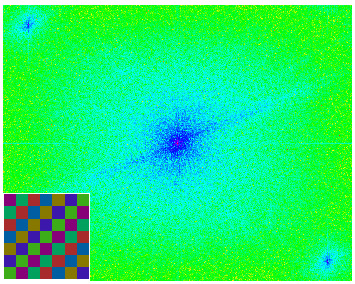}\\
\footnotesize{RGBCY \cite{Anzagira_Fossum_2015}} &
\footnotesize{RGBCWY \cite{Anzagira_Fossum_2015}}&
\footnotesize{Biay-Cheng et al. \cite{Cheng_Siddiqui_Luo_2015}} &
\footnotesize{Condat \cite{Condat_2011}} &
\footnotesize{Hirakawa-Wolfe \cite{Hirakawa_Wolfe_2008}} &
\footnotesize{ Bai et al. \cite{Bai_Li_Lin_2016}} \\
\hspace{-1.8ex}\includegraphics[width=0.16\linewidth]{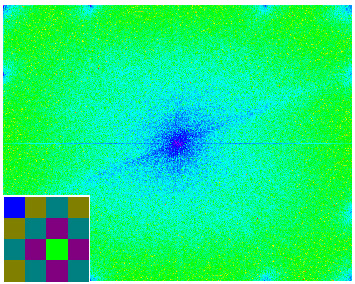}&
\hspace{-2.2ex}\includegraphics[width=0.16\linewidth]{elgen20}&
\hspace{-2.2ex}\includegraphics[width=0.16\linewidth]{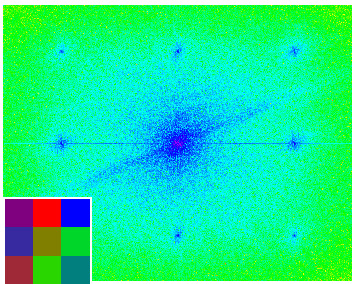}&
\hspace{-2.2ex}\includegraphics[width=0.16\linewidth]{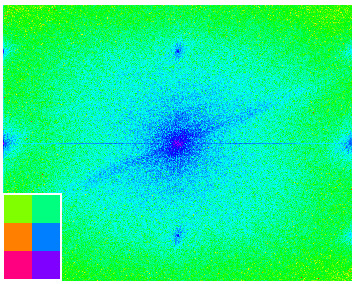}&
\hspace{-2.2ex}\includegraphics[width=0.16\linewidth]{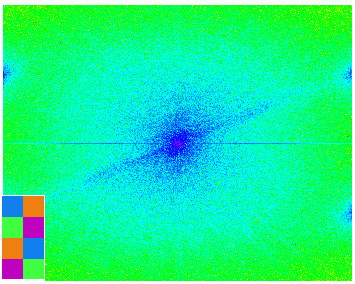}&
\hspace{-2.2ex}\includegraphics[width=0.16\linewidth]{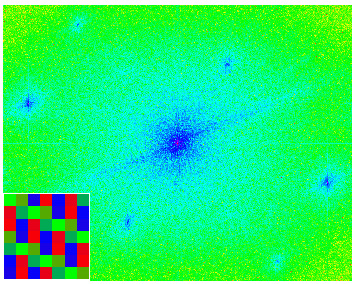}\\
\footnotesize{Hao et al. \cite{Hao_Li_Lin_2011}} &
\footnotesize{Ours} &
\footnotesize{Ours} &
\footnotesize{Ours} &
\footnotesize{Ours} &
\footnotesize{Ours}
\end{tabular}
\caption{Our proposed CFAs compared to other CFAs in literature. For every CFA, we show the spectrum of the ``bike" Kodak image mosaicked by this CFA. We also show the organization of luminance ($L$) and chrominance channels ($\alpha$ and $\beta$) for CFAs that satisfy orthogonality constraint.}
\label{fig:CFAs}
\vspace{-1ex}
\end{figure*}

Figure~\ref{fig:CFAs} shows our proposed CFAs and their accompanied spectra compared to other CFAs in the literature.  For every CFA, we compute 1) the luminance and chrominance gains ($\gamma_l$ and $\gamma_c$) in Equation \eref{eq:def} to measure robustness to noise, 2) the total variation metric $TV(x)$ (Proposition~\ref{prop:ctk}) to measure robustness to crosstalk, and 3) the aliasing metric $J_l$ in Equation \eref{eq:aliasing} to measure robustness to aliasing, and the condition number $\kappa(\mA)$ defined in Section~\ref{subsec:cond}. To calculate the aliasing metric for \cite{Hirakawa_Wolfe_2008}, \cite{Hao_Li_Lin_2011}, \cite{Anzagira_Fossum_2015} and \cite{Cheng_Siddiqui_Luo_2015}, we use the transformation matrices that make the luminance and chrominance channels orthogonal as mentioned in Section~\ref{subsec:freqSelection}. Leakage factors $(\delta_r,\delta_g,\delta_b)$ are chosen as $(0.23,0.15,0.10)$. Results are summarized in Table~\ref{tab:PSNR}.

\begin{itemize}
\item \textbf{$4\times4$}: Among all $4\times4$ CFAs in Table~\ref{tab:PSNR}, Hao et al. \cite{Hao_Li_Lin_2011} is the most robust CFA to aliasing, but the least robust to crosstalk; whereas RGBCWY \cite{Anzagira_Fossum_2015} is the most robust to crosstalk, but it is not as robust to aliasing. Our CFA achieves the best of both worlds by having the same total-variation like RGBCWY, and good aliasing metric. Moreover, it has the lowest condition number and it is more sensitive since it comprises 4 panchromatic or ``white" pixels with $(r,g,b)=(1/3,1/3,1/3)$.
\item \textbf{$3\times3$}: Compared to Biay-Cheng et al. \cite{Cheng_Siddiqui_Luo_2015}, our CFA has less aliasing. The high aliasing metric of \cite{Cheng_Siddiqui_Luo_2015} is attributed to its design which overlooks frequency domain aliasing.
\item \textbf{$3\times2$}: Compared to Condat \cite{Condat_2011}, our CFA is more robust to crosstalk and aliasing. It is worth noting that we get exactly the same CFA as Condat's automatically when we remove the crosstalk constraint.
\item \textbf{$4\times2$}: Compared to Hirakawa-Wolfe \cite{Hirakawa_Wolfe_2008}, our CFA has higher chrominance sensitivity and it is more robust to crosstalk.
\item \textbf{$7\times7$}: Compared to Bai et al. \cite{Bai_Li_Lin_2016}, our CFA is more robust to crosstalk and aliasing.
\end{itemize}

\begin{figure*}[t]
\centering
\begin{tabular}{ccccc}
\includegraphics[width=0.17\linewidth]{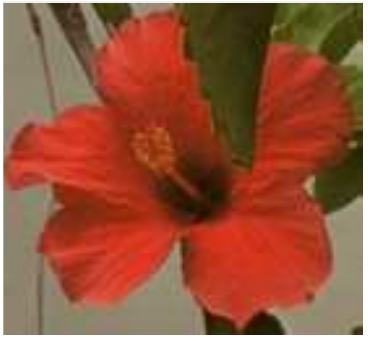}&
\hspace{-2.0ex}\includegraphics[width=0.17\linewidth]{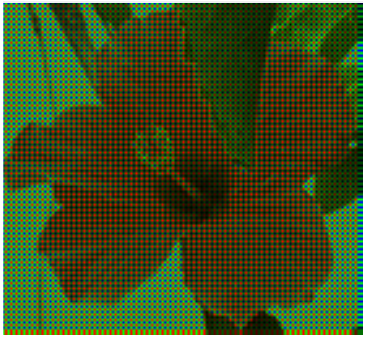}&
\hspace{-2.0ex}\includegraphics[width=0.17\linewidth]{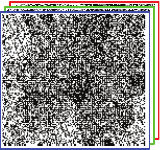}&
\hspace{-2.0ex}\includegraphics[width=0.17\linewidth]{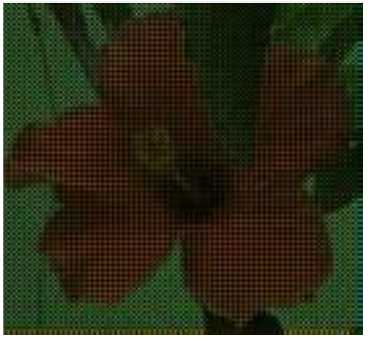}&
\hspace{-2.0ex}\includegraphics[width=0.17\linewidth]{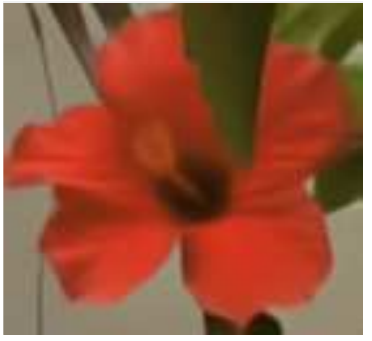}\\
\footnotesize{(a) Ground Truth}&
\hspace{-2.0ex}\footnotesize{(b) Mosaicked Image}&
\hspace{-2.0ex}\footnotesize{(c) Binary Measurements}&
\hspace{-2.0ex}\footnotesize{(d) Summed Frame}&
\hspace{-2.0ex}\footnotesize{(e) Reconstruction,  $31.64$dB}
\end{tabular}
\caption{\footnotesize{Simulation Setup: The ground truth image (a) is color filtered by a CFA to produce a mosaicked image (b). QIS generates $T=1000$ binary frames (c) using the mosaicked image as light exposure. The $T$ binary framed are summed to give an approximately clean image (d). Then, ADMM is applied to obtain the demosaicked image (e). Crosstalk is not added to this example, so there is no need for color correction.}}
\label{fig:pipeline}
\vspace{-2ex}
\end{figure*}

\subsection{Experiment 2: Macbeth ColorChecker Reconstruction}
\label{subsec:macbeth}
In this experiment, we simulate the performance of different CFAs in reconstructing the Macbeth ColorChecker image. Pixel response is determined using the incident photon flux of D65 light and the spectral reflectance of Macbeth ColorChecker integrated over the visible light spectrum. QIS parameters and primary color filters are taken from \cite{Anzagira_Fossum_2015}. For every CFA, we generate mosaicked images under two scenarios: 1) crosstalk kernels with leakage factors $(\delta_r,\delta_g,\delta_b)=(0,0,0)$, i.e., no crosstalk, and 2) crosstalk kernels with leakage factors $(\delta_r,\delta_g,\delta_b)=(0.45,0.30,0.20)$ as suggested by \cite{Anzagira_Fossum_2015}. For fairness of comparison, we use Algorithm~\ref{alg:freqSelection} for demosaicking all CFAs including RGBCY and RGBCWY CFAs proposed in \cite{Anzagira_Fossum_2015}. Color correction with white balance is performed after color demosaicking for the crosstalk case for removing the crosstalk effect.

We use the following metrics \cite{Anzagira_Fossum_2015} to evaluate the CFAs:
\begin{itemize}
\item Sensitivity metamerism index (SMI) which measures the drop in color reproduction accuracy due to crosstalk. It is obtained as a function of the CIEDE2000 color error metric which is obtained by calculating the mean square color difference in the CIELAB color space.
\item Luminance SNR (YSNR) which measures the visual noise of luminance channel as defined in ISO 12232 \cite{ISO}.
\end{itemize}
Table~\ref{tab:PSNR} shows these metrics for different CFAs with and without crosstalk. Our CFAs achieve higher color reproduction accuracy compared to others since they are optimized for crosstalk. This gain in color accuracy happens by trading the noise performance as observed by the drop of YPSNR metric.



\subsection{Experiment 3: Natural Color Image Reconstruction}
\label{subsec:imRec}
In this experiment, we evaluate the performance of different CFAs for natural color image reconstruction. To this end, we use the DIV2K evaluation dataset to generate 100 mosaicked images according to QIS model. Two scenarios are assumed: 1) No crosstalk, and 2) Moderate crosstalk with leakage factors $(\delta_r,\delta_g,\delta_b)=(0.23,0.15,0.10)$. The low pass filter in Algorithm~\ref{alg:freqSelection} is chosen as $21\times21$ Gaussian with standard deviation $\sigma=21/3$ and multiplied by a hamming window to mitigate the windowing effect. We simulate the diffraction of light occurring in QIS by blurring the ground truth image with a $5\times5$ box kernel

Table~\ref{tab:PSNR} shows the reconstruction quality averaged on 100 images in the DIV2K evaluation dataset. We choose the color PSNR as a quality metric, which is calculated as $\mathrm{CPSNR}=10\log_{10}\left(\frac{1}{\mathrm{MSE}}\right)$ where $\mathrm{MSE}$ is the total mean squared error summed over the three color channels. Our CFAs achieve better quality for the crosstalk case. Visually, our CFAs obtain color images with less aliasing artifacts and better details as shown in Figure~\ref{fig:recImages}.

\begin{figure*}[!t]
\centering
\begin{tabular}{cccccccc}
&
\multicolumn{2}{c}{\hspace{-3.0ex}$\overset{4\times4}{\overbrace{\hspace{14em}}}$}&
\hspace{-3.0ex}$\overset{3\times3}{\overbrace{\hspace{6em}}}$ &
\hspace{-3.0ex}$\overset{3\times2}{\overbrace{\hspace{6em}}}$&
\hspace{-3.0ex}$\overset{4\times2}{\overbrace{\hspace{6em}}}$&
\hspace{-3.0ex}$\overset{7\times7}{\overbrace{\hspace{6em}}}$\\
\hspace{-2.5ex}\includegraphics[width=0.14\linewidth]{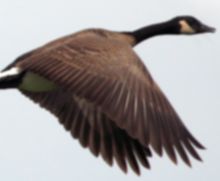}&
\hspace{-2.5ex}\includegraphics[width=0.14\linewidth]{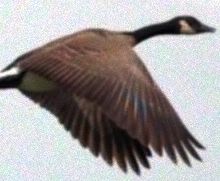}&
\hspace{-2.5ex}\includegraphics[width=0.14\linewidth]{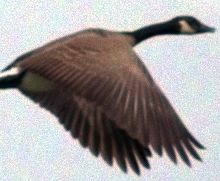}&
\hspace{-2.5ex}\includegraphics[width=0.14\linewidth]{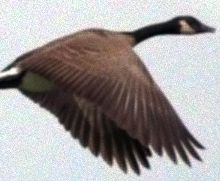}&
\hspace{-2.5ex}\includegraphics[width=0.14\linewidth]{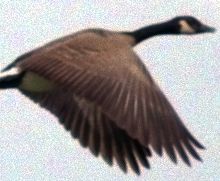}&
\hspace{-2.5ex}\includegraphics[width=0.14\linewidth]{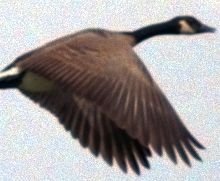}&
\hspace{-2.5ex}\includegraphics[width=0.14\linewidth]{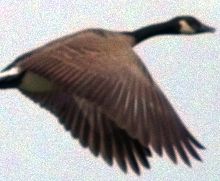}\\
\hspace{-2.5ex}\footnotesize{Ground Truth}&
\hspace{-2.5ex}\footnotesize{RGBCY\cite{Anzagira_Fossum_2015}: $29.28$dB}&
\hspace{-2.5ex}\footnotesize{\cite{Hao_Li_Lin_2011}: $28.71$dB}&
\hspace{-2.5ex}\footnotesize{\cite{Cheng_Siddiqui_Luo_2015}: $29.44$dB}&
\hspace{-2.5ex}\footnotesize{\cite{Condat_2011}: $28.23$dB}&
\hspace{-2.5ex}\footnotesize{\cite{Hirakawa_Wolfe_2008}: $28.07$dB}&
\hspace{-2.5ex}\footnotesize{\cite{Bai_Li_Lin_2016}: $28.93$dB}\\
\hspace{-2.5ex}\includegraphics[width=0.14\linewidth]{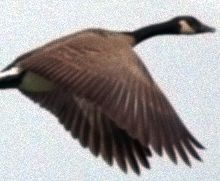}&
\hspace{-2.5ex}\includegraphics[width=0.14\linewidth]{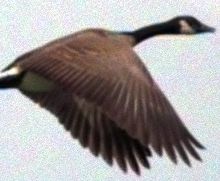}&
\hspace{-2.5ex}\includegraphics[width=0.14\linewidth]{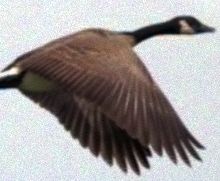}&
\hspace{-2.5ex}\includegraphics[width=0.14\linewidth]{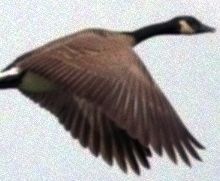}&
\hspace{-2.5ex}\includegraphics[width=0.14\linewidth]{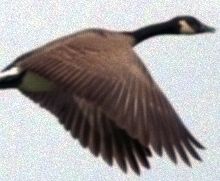}&
\hspace{-2.5ex}\includegraphics[width=0.14\linewidth]{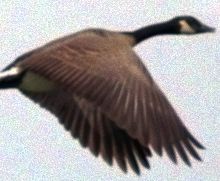}&
\hspace{-2.5ex}\includegraphics[width=0.14\linewidth]{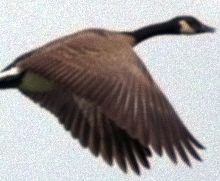}\\
\hspace{-2.5ex}\footnotesize{Bayer: $29.29$dB}&
\hspace{-2.5ex}\footnotesize{RGBCWY\cite{Anzagira_Fossum_2015}: $29.36$dB}&
\hspace{-2.5ex}\footnotesize{Ours: $29.60$dB}&
\hspace{-2.5ex}\footnotesize{Ours: $29.77$dB}&
\hspace{-2.5ex}\footnotesize{Ours: $28.49$dB}&
\hspace{-2.5ex}\footnotesize{Ours: $28.32$dB}&
\hspace{-2.5ex}\footnotesize{Ours: $29.74$dB}\\
\end{tabular}
\caption{Reconstructed color images from the QIS measurements on ``birds" image. Each subfigure shows the result using a particular CFA design.}
\label{fig:recImages}
\vspace{-2ex}
\end{figure*}

\begin{figure*}[!t]
\centering
\begin{tabular}{cccccccc}
&
\multicolumn{2}{c}{\hspace{-3.0ex}$\overset{4\times4}{\overbrace{\hspace{14em}}}$}&
\hspace{-3.0ex}$\overset{3\times3}{\overbrace{\hspace{6em}}}$ &
\hspace{-3.0ex}$\overset{3\times2}{\overbrace{\hspace{6em}}}$&
\hspace{-3.0ex}$\overset{4\times2}{\overbrace{\hspace{6em}}}$&
\hspace{-3.0ex}$\overset{7\times7}{\overbrace{\hspace{6em}}}$\\
\hspace{-2.5ex}\includegraphics[width=0.14\linewidth]{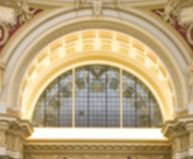}&
\hspace{-2.5ex}\includegraphics[width=0.14\linewidth]{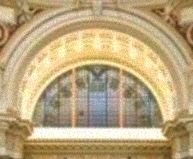}&
\hspace{-2.5ex}\includegraphics[width=0.14\linewidth]{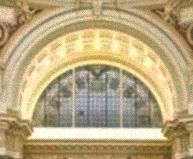}&
\hspace{-2.5ex}\includegraphics[width=0.14\linewidth]{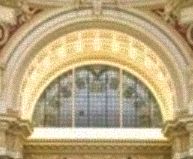}&
\hspace{-2.5ex}\includegraphics[width=0.14\linewidth]{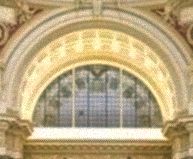}&
\hspace{-2.5ex}\includegraphics[width=0.14\linewidth]{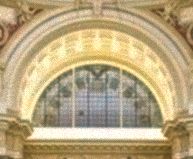}&
\hspace{-2.5ex}\includegraphics[width=0.14\linewidth]{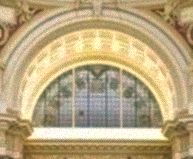}\\
\hspace{-2.5ex}\footnotesize{Ground Truth}&
\hspace{-2.5ex}\footnotesize{RGBCY\cite{Anzagira_Fossum_2015}: $30.71$dB}&
\hspace{-2.5ex}\footnotesize{\cite{Hao_Li_Lin_2011}: $30.98$dB}&
\hspace{-2.5ex}\footnotesize{\cite{Cheng_Siddiqui_Luo_2015}: $31.42$dB}&
\hspace{-2.5ex}\footnotesize{\cite{Condat_2011}: $31.44$dB}&
\hspace{-2.5ex}\footnotesize{\cite{Hirakawa_Wolfe_2008}: $31.33$dB}&
\hspace{-2.5ex}\footnotesize{\cite{Bai_Li_Lin_2016}: $31.22$dB}\\
\hspace{-2.5ex}\includegraphics[width=0.14\linewidth]{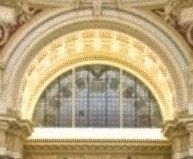}&
\hspace{-2.5ex}\includegraphics[width=0.14\linewidth]{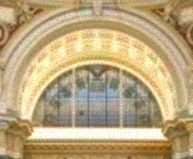}&
\hspace{-2.5ex}\includegraphics[width=0.14\linewidth]{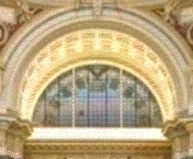}&
\hspace{-2.5ex}\includegraphics[width=0.14\linewidth]{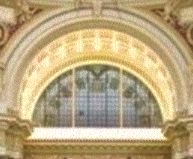}&
\hspace{-2.5ex}\includegraphics[width=0.14\linewidth]{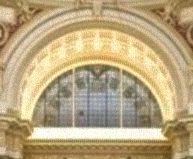}&
\hspace{-2.5ex}\includegraphics[width=0.14\linewidth]{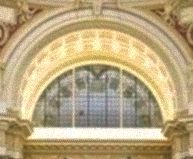}&
\hspace{-2.5ex}\includegraphics[width=0.14\linewidth]{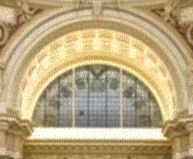}\\
\hspace{-2.5ex}\footnotesize{Bayer: $31.08$dB}&
\hspace{-2.5ex}\footnotesize{RGBCWY\cite{Anzagira_Fossum_2015}: $30.77$dB}&
\hspace{-2.5ex}\footnotesize{Ours: $31.30$dB}&
\hspace{-2.5ex}\footnotesize{Ours: $31.59$dB}&
\hspace{-2.5ex}\footnotesize{Ours: $31.74$dB}&
\hspace{-2.5ex}\footnotesize{Ours: $31.69$dB}&
\hspace{-2.5ex}\footnotesize{Ours: $31.37$dB}\\
\end{tabular}
\caption{Reconstructed color images from the QIS measurements on ``building" image. Each subfigure shows the result using a particular CFA design.}
\label{fig:recImages}
\vspace{-2ex}
\end{figure*}

\section{Conclusion}\label{sec:conc}
We presented a general and flexible optimization framework to design color filter arrays for QIS. Our framework unifies the CMOS-based color filter array designs and extends to QIS. The color filter arrays designed by our framework are robust to crosstalk between the primary color channels, robust to aliasing between the luminance and chrominance channels, and are robust to noise. We verified the designs both theoretically and numerically through extensive experiments. Our evaluation indicated the effectiveness of the framework in offering trade-offs between different design criteria.

\section*{Acknowledgment}
The authors thank Professor Keigo Hirakawa (Univ. Dayton), Professor Eric Fossum (Dartmouth College), Jiaju Ma (GigaJot Tech Inc.) and Abhiram Gnanasambandam (Purdue Univ.) for many insightful discussions about CFA designs. This work is supported, in part, by the National Science Foundation under grants CCF-1763896 and CCF-1718007.

\appendices
\section{} \label{apx:proofs}
\subsection{Proof of Proposition \ref{prop:gammas}}
Since the luminance channel comprises only one baseband component in the frequency domain, the luminance sensitivity in the amplitude of this component, i.e.,
\begin{align*}
\gamma_l &= \frac{1}{K}||\widetilde{\vh}_l||_2 = \frac{1}{K} \sqrt{\widetilde{h}_l^2(0,0) + 0 +\ldots + 0}\nonumber\\
&=\frac{1}{K}\widetilde{h}_l(0,0).
\end{align*}
Substituting in the DFT equation with $u=v=0$, we get
\begin{align*}
\gamma_l(\vx) &= \frac{1}{K} \, \sum_{m=0}^{M-1} \sum_{n=0}^{N-1} h_l(m,n) \nonumber \\
&= \frac{1}{K} \vone^T \vh_l = \frac{1}{K} \vone^T \mZ_l \vx = \vb^T \vx,
\end{align*}
where $\vb\bydef\frac{1}{K}\vone^T \mZ_l$. As for the chrominance sensitivity $\gamma_c$, by squaring the definition in Equation \eref{eq:def}, we get
\begin{align}\label{eq:gamma_c}
\gamma_c(\vx)^2 &= \frac{1}{K^2}\textrm{min}\left(||\widetilde{\vh}_{\alpha}||_2^2,||\widetilde{\vh}_{\beta}||_2^2\right)\\
&\overset{(a)}{=}\min \left(||\vh_\alpha||_2^2, ||\vh_\beta||_2^2\right) \nonumber \\
&= \min \left(||\mZ_\alpha x||_2^2,||\mZ_\beta x||_2^2\right) = \min \left(\vx^T \mQ_\alpha \vx, \vx^T \mQ_\beta \vx\right), \notag
\end{align}
where (a) follows from Parseval theorem, and $\mQ_\alpha\bydef \mZ_\alpha^T\mZ_\alpha$ and $\mQ_\beta\bydef \mZ_\beta^T\mZ_\beta$ are two positive semidefinite matrices.

\bibliographystyle{IEEEbib}
\bibliography{refs}

\end{document}


\author{Omar A. Elgendy, Stanley H. Chan\\
\small{Purdue University}}

\maketitle
\vspace{10mm}
\begin{abstract}
\noindent This supplementary report provides the following additional information of the main article.
\begin{itemize}
\item Luminance/Chrominance Transformation Matrices of Other CFAs
\item Experiment 4: Color-Noise Trade-off 
\item Experiment 5: Natural Image Reconstructionn on Kodak and McMaster datasets
\end{itemize}
\noindent Please view this document in presentation mode, e.g.,
\begin{itemize}[leftmargin=7cm, noitemsep]
\item[Acrobat Adobe Reader:]  \texttt{Ctrl+L} on Windows and \texttt{Command+L} on Mac
\item[Evince (Ubuntu):]  \texttt{F5}
\item[Preview.app (OSX):]  \texttt{Command+Shift+F}.
\end{itemize}
To compare the images resulting from different CFA designs, please click the corresponding hyperlinks below each figure. For each image we provide a few details that are worth observing.
\begin{center}
\color{red}{\textbf{It is important to look at the results in presentation mode so that the images are aligned.}}
\end{center}
\end{abstract}
\blindfootnote{Document created with code from: \href{https://github.com/gorazione/make_supplementary}{\color{black}https://github.com/gorazione/make\_supplementary}}
\blindfootnote{The authors are with the School of Electrical and Computer Engineering, Purdue University, West Lafayette, IN 47907, USA. Email:\texttt{\{ oelgendy, stanchan\}@purdue.edu}}.


\section{Luminance/Chrominance Transformation Matrices of Other CFAs}
\label{sec:transformation}
Algorithm II in the main manuscript performs demosaicking by frequency selection with the assumption of orthogonality. However, the CFAs proposed in \cite{Hao_Li_Lin_2011}, \cite{Anzagira_Fossum_2015} and \cite{Cheng_Siddiqui_Luo_2015} do not satisfy the orthogonality constraint with our choice of $\mT$ \cite{Condat_2011}. In this section, we derive for every CFA the transformation matrix $\mT$ that makes its luminance and chrominance channel orthogonal so that we can apply Algorithm II.

\begin{figure}[!t]
\centering
\includegraphics[width=0.5\linewidth]{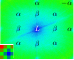}
\caption{Frequency structure of RGBCY CFA \cite{Anzagira_Fossum_2015} using the luminance/chrominance transformation \eref{eq:freqStruc}}
\label{fig:freq}
\vspace{0ex}
\end{figure}

Following the symbolic DFT method in \cite{Hao_Li_Lin_2011}, the frequency structure of RGBCY CFA proposed in \cite{Anzagira_Fossum_2015} has the following form:
\begin{equation}
\frac{1}{16}\begin{bmatrix}3B + 10G + 3R &   2R - 2B & B - 2G + R&   2R - 2B\\
2R - 2B& B - 2G + R&           0& B - 2G + R\\
B - 2G + R&           0& 2G - B - R&           0\\
2R - 2B& B - 2G + R&           0& B - 2G + R\end{bmatrix}
\end{equation}
Hence, we can choose the luminance/chrominance transformation as
\begin{equation}
\begin{bmatrix}L\\\alpha\\\beta\end{bmatrix} = \frac{1}{16}\begin{bmatrix}3 & 10 & 3\\1 & -2 & 1\\2 & 0 & -2\end{bmatrix}\begin{bmatrix}R\\G\\B\end{bmatrix} \leftrightarrow \mT_{\mathrm{RGBCY}} =\frac{1}{16}\begin{bmatrix}3 & 10 & 3\\1 & -2 & 1\\2 & 0 & -2\end{bmatrix} 
\end{equation}
As a result, the frequency structure is orthogonal where every chrominance component is modulated on distinct carrier as shown in Figure~\ref{fig:freq} and shown in the following matrix representation
\begin{equation}\label{eq:freqStruc}
\frac{1}{16}\begin{bmatrix}L &   \beta & \alpha&   \beta\\
\beta& \alpha&           0& \alpha\\
\alpha&           0& -\alpha&           0\\
\beta& \alpha&           0& \alpha\end{bmatrix}
\end{equation}
To ensure fairness between different CFAs, we normalize the matrix rows to unity so that all luminance and chrominance have the same noise power. To this end, the transformation matrix of RGBCY CFA can be written as
\begin{equation}
\mT_{\mathrm{RGBCY}}=\begin{bmatrix}\frac{3}{\sqrt{118}} & \frac{10}{\sqrt{118}} & \frac{3}{\sqrt{118}}\\\frac{1}{\sqrt{6}} & \frac{-2}{\sqrt{6}} & \frac{1}{\sqrt{6}}\\\frac{1}{\sqrt{2}} & 0 & \frac{-1}{\sqrt{2}}\end{bmatrix} 
\end{equation}
Similarly, we can do the same steps for RGBCWY CFA in \cite{Anzagira_Fossum_2015} to obtain the following transformation matrix.
\begin{equation}
\mT_{\mathrm{RGBCWY}}=\begin{bmatrix}\frac{13}{\sqrt{822}} & \frac{22}{\sqrt{822}} & \frac{13}{\sqrt{822}}\\\frac{1}{\sqrt{6}} & \frac{-2}{\sqrt{6}} & \frac{1}{\sqrt{6}}\\\frac{1}{\sqrt{2}} & 0 & \frac{-1}{\sqrt{2}}\end{bmatrix} 
\end{equation}
As for Bayer CFA, and the CFA in \cite{Cheng_Siddiqui_Luo_2015}, we use the following transformation matrix
\begin{equation}
\mT_{\mathrm{Bayer}}=\begin{bmatrix}\frac{1}{\sqrt{6}} & \frac{2}{\sqrt{6}} & \frac{1}{\sqrt{6}}\\\frac{1}{\sqrt{6}} & \frac{-2}{\sqrt{6}} & \frac{1}{\sqrt{6}}\\\frac{1}{\sqrt{2}} & 0 & \frac{-1}{\sqrt{2}}\end{bmatrix} 
\end{equation}
Finally, for the CFA in \cite{Hao_Li_Lin_2011}, we use the following transformation matrix
\begin{equation}
\mT=\begin{bmatrix}\frac{2}{\sqrt{22}} & \frac{3}{\sqrt{22}} & \frac{3}{\sqrt{22}}\\ 0 & \frac{-1}{\sqrt{2}} & \frac{1}{\sqrt{2}}\\\frac{-2}{\sqrt{6}} & \frac{1}{\sqrt{6}} & \frac{1}{\sqrt{6}}\end{bmatrix} 
\end{equation}

\clearpage

\section{Experiment 4: Color-Noise Trade-off} \label{subsec:tradeoff}
In this experiment, we compare the trade-off between noise amplification and color accuracy of our proposed CFAs and other CFAs in literature. To do so, we use the Macbeth color chart that comprises 24 color patches. The forward model consists of illumination using D65 light and mosaicking using a CFA and crosstalk using the crosstalk kernels:
\begin{equation}\label{eq:crosstalkKernels}
g_i=\begin{bmatrix}
0 & \alpha_i/4 & 0\\
\alpha_i/4 & 1-\alpha_i & \alpha_i/4\\
0 & \alpha_i/4 & 0
\end{bmatrix}, \; i\in\{r,g,b\},
\end{equation}
with $(\alpha_r,\alpha_g,\alpha_b)=(0.45,0.30,0.20)$ as suggested in \cite{Anzagira_Fossum_2015}. QIS parameters are $q=1$, $\alpha=2$ and $T=1000$. We use Algorithm II for demosaicking with frequency selection. The low pass filter is $m\times m$ Gaussian having standard deviation $\sigma=m/3$ and multiplied by a Hamming window to eliminate windowing effect. Since the ground truth color values of Macbeth color chart are known, we compute the color correction matrix $\mM$ by solving the following regularized linear least squares optimization problem with white balance constraint:
\begin{align}\label{eq:lincolorcorrNoise}
&\mM = \arg \min_{\mM} \epsilon_c(\mM) + \kappa \sum_{i=1}^{24} \, ||\textrm{Cov}(\mM\mQ^{(i)}_\textrm{False})||_2^2\nonumber\\
&\mbox{ subject to}\nonumber\\
&\mM\vu = \vu
\end{align}
where $\epsilon_c(\mM)=\mathrm{Tr}\left\{\left(\mM\mQ_{\textrm{False}}-\mQ_{\textrm{GT}}\right)^T\left(\mM\mQ_{\textrm{False}}-\mQ_{\textrm{GT}}\right)\right\}$ is the color error. $\mQ_{\textrm{False}}$ and  $\mQ_{\textrm{GT}}$ are $3\times K$ matrices containing the measured color values and the corresponding ground truth color values of $K$ pixels. ${\vu\bydef[0.95,1,1.0889]^T}$ is the white point for D65 illuminant.

To draw the noise-color trade-off curve, we vary the parameter $\mu$ in \eref{eq:lincolorcorrNoise} from $0$ to $10^{8}$ on the log-scale. Color error is quantified with the CIEDE2000 metric which is obtained by calculating the mean square color difference in the CIELAB color space \cite{Anzagira_Fossum_2015}. Visual noise is measured by the YSNR metric as defined in ISO 12232 \cite{Anzagira_Fossum_2015}. Since YSNR should be increased and color error should be decreased, the tradeoff curve is better when it is shifted to upper left.

\begin{figure}[ht]
\centering
\begin{tabular}{cc}
\hspace{-2.5ex}\includegraphics[width=0.3\linewidth]{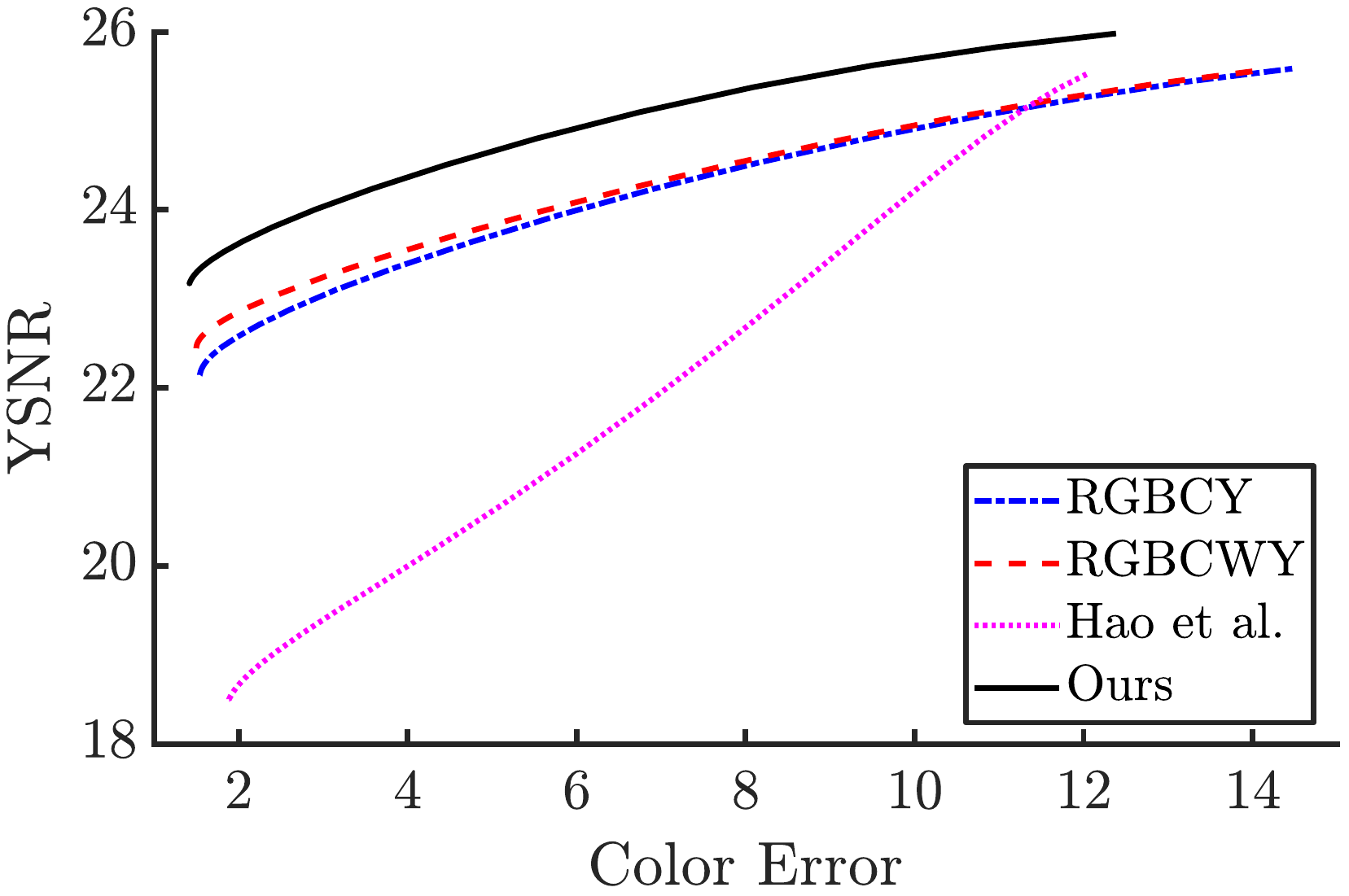}&
\hspace{-2.5ex}\includegraphics[width=0.3\linewidth]{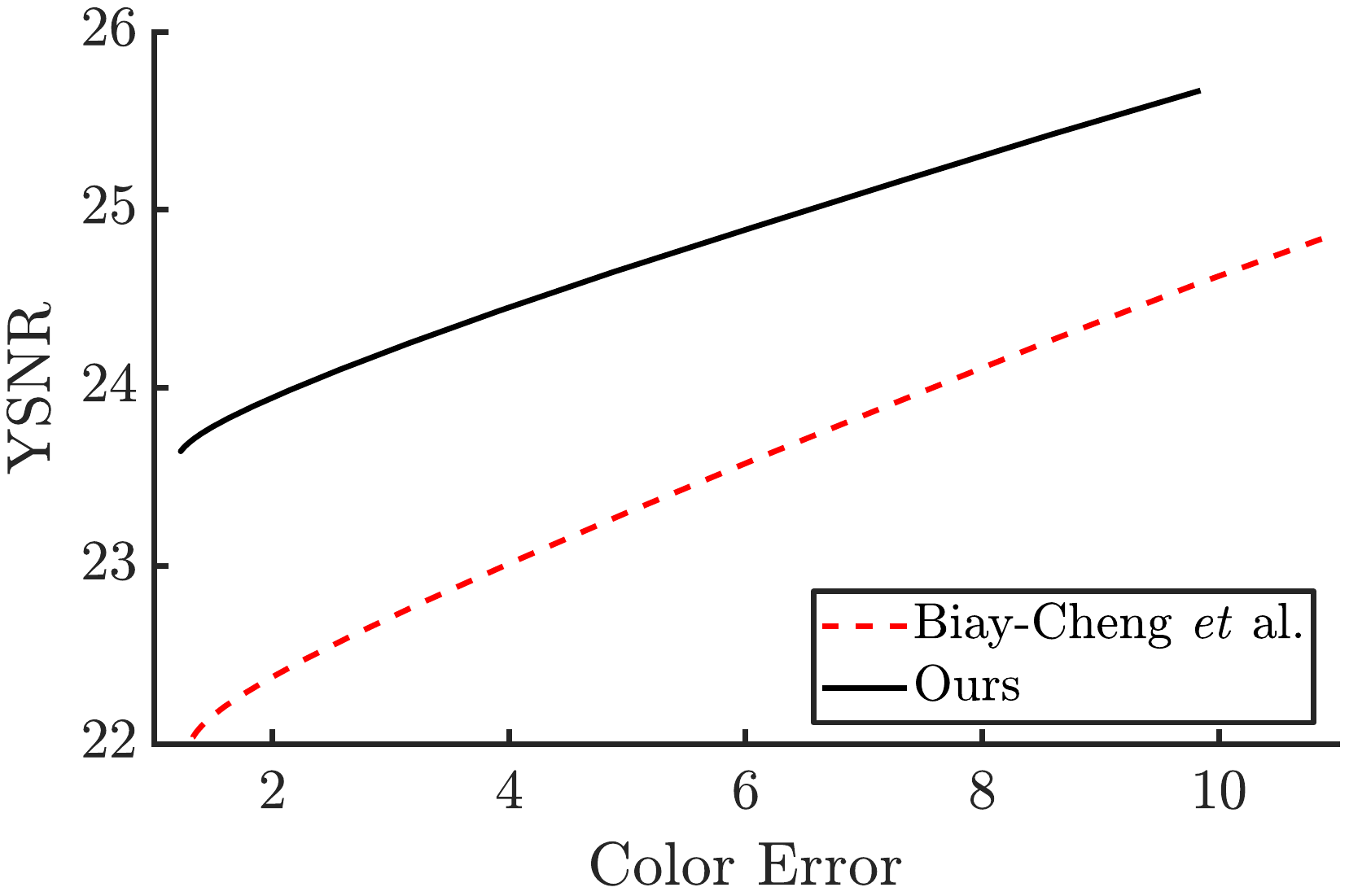}\\
\footnotesize{(a) $4\times 4$} &
\footnotesize{(b) $3\times 3$} \\
\end{tabular}
\begin{tabular}{ccc}
\hspace{-2.5ex}\includegraphics[width=0.3\linewidth]{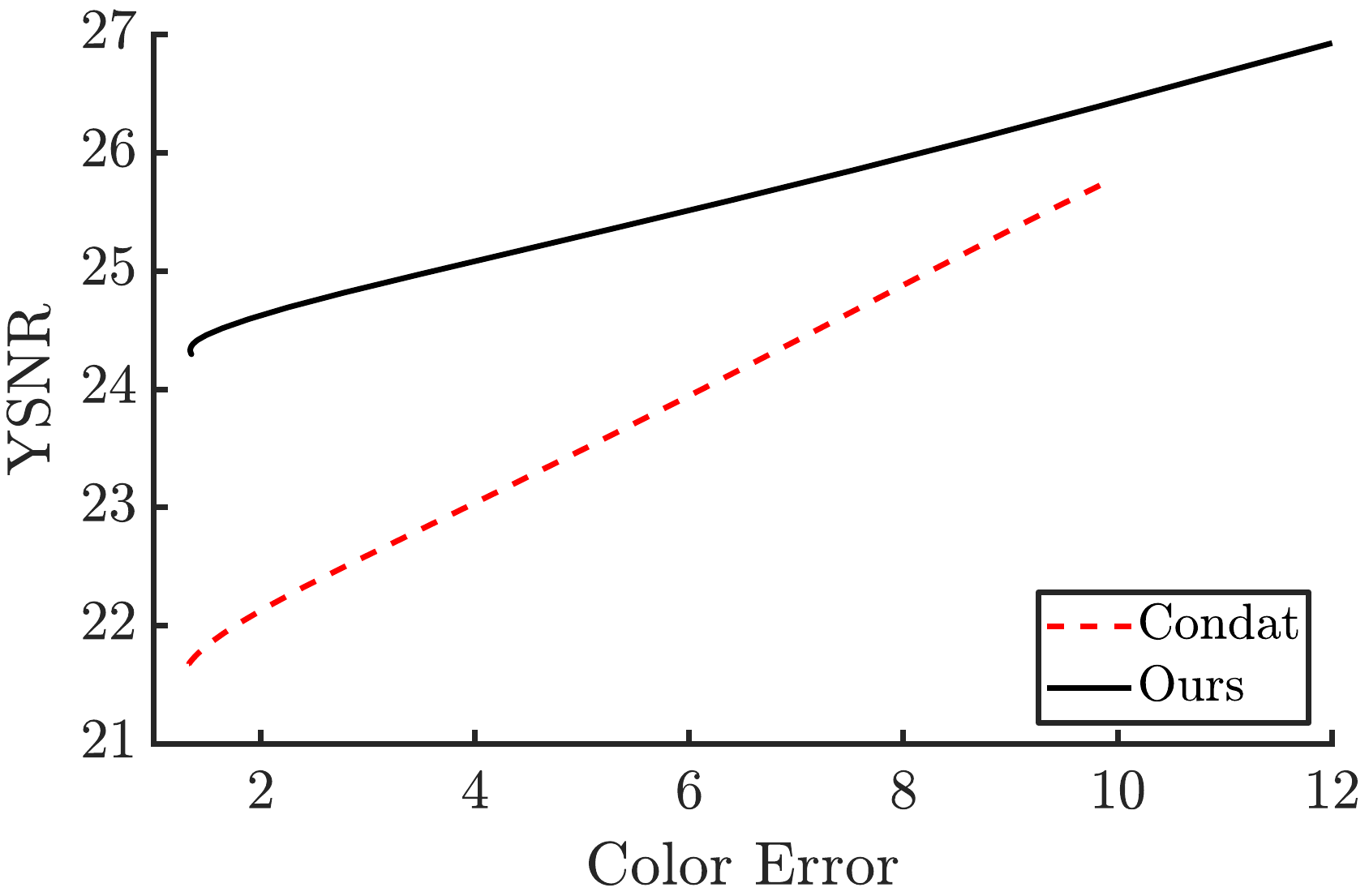}&
\hspace{-2ex}\includegraphics[width=0.3\linewidth]{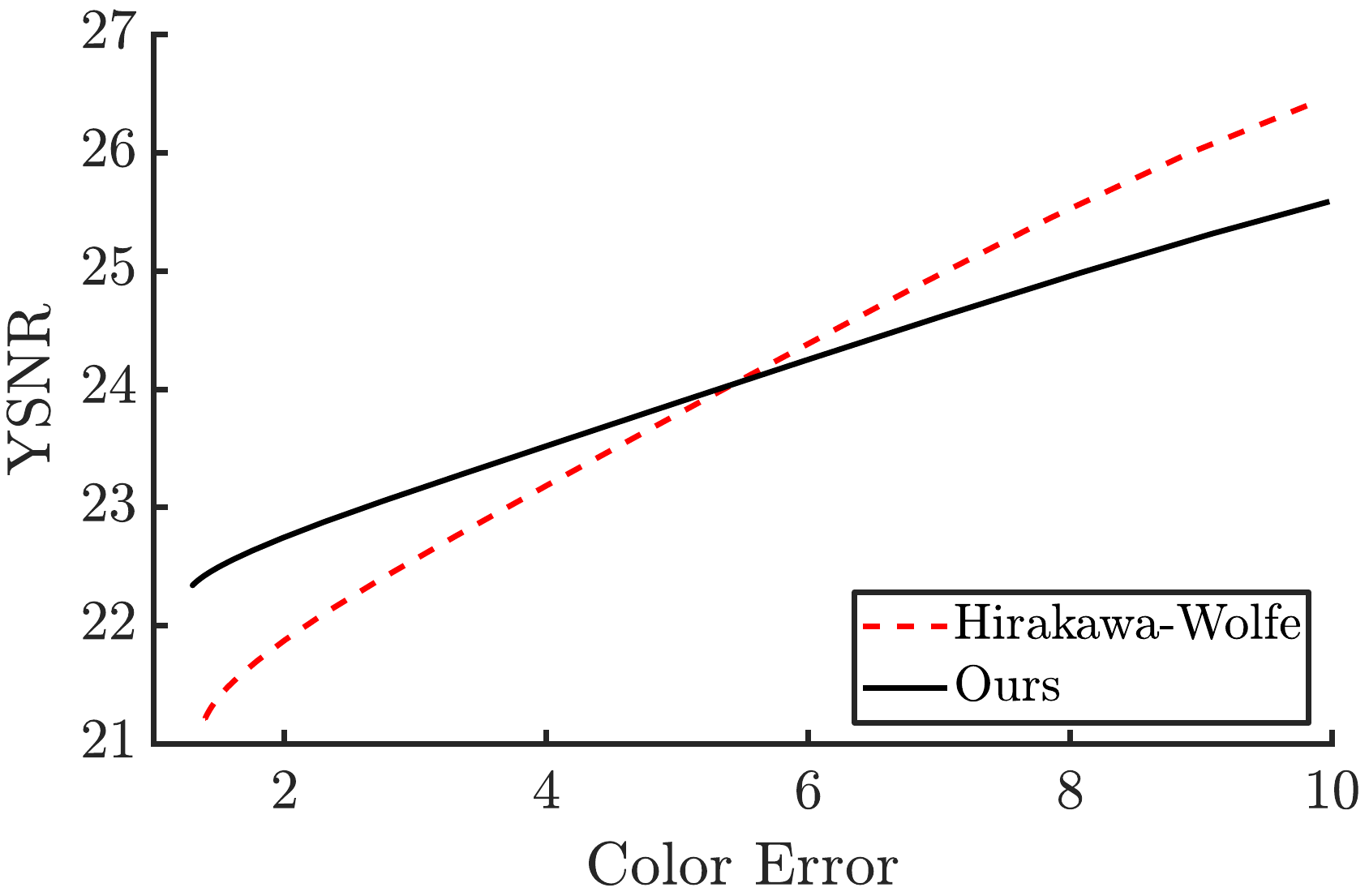}&
\hspace{-2ex}\includegraphics[width=0.3\linewidth]{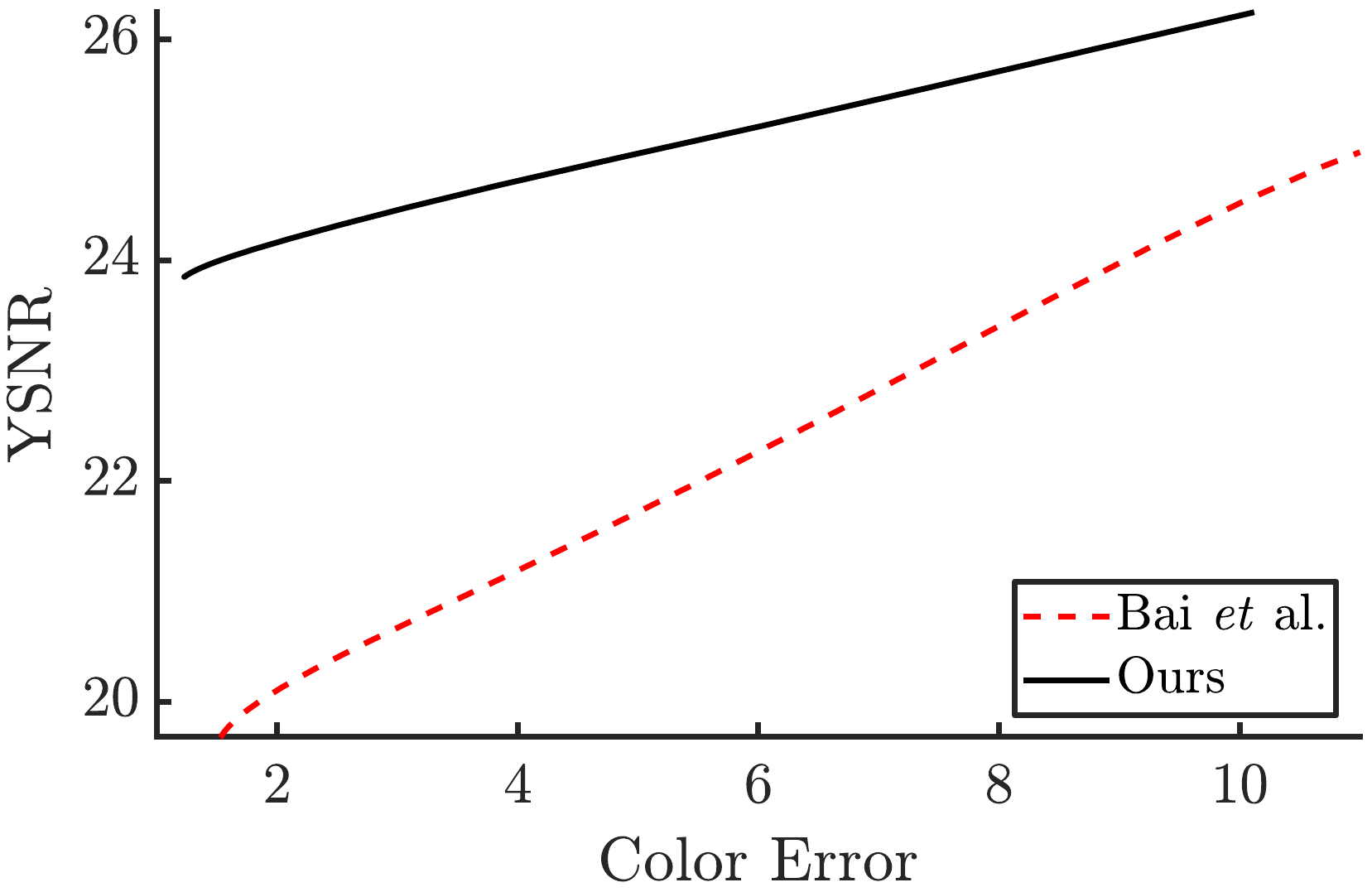}\\
\footnotesize{(c) $3\times 2$}&
\footnotesize{(d) $4\times 2$} &
\footnotesize{(e) $7\times 7$} \\
\end{tabular}
\caption{Color-Noise trade-off for different CFAs. Demosaicking is performed using Algorithm II. $\mu$ in \eref{eq:lincolorcorrNoise} is varied from $0$ to $10^{8}$. }
\label{fig:conv1}
\vspace{0ex}
\end{figure}

Figure~\ref{fig:conv1} shows the trade-off curves for the proposed CFAs and other CFAs.  Our $4\times4$, $3\times3$, $3\times2$ and $7\times7$ CFAs are better than corresponding CFAs with same atom size for all values of $\mu$. As for $4\times2$ CFAs, our CFA is better than \cite{Hirakawa_Wolfe_2008} if we restrict to small color error. However, if we allow larger color error, then \cite{Hirakawa_Wolfe_2008} is better. 

\section{Experiment 5: Natural Image Reconstruction} \label{subsec:imRec}
In this experiment, we repeat Experiment 3 in the main manuscript on other datasets. QIS parameters are chosen as $(q,\eta,T)=(1,2,1000)$. We simulate the diffraction of light occuring in QIS by blurring the ground truth image with a $3\times3$ box kernel. We choose Kodak \cite{Gunturk_Glotzbach_Altunbasak_2005} and McMaster \cite{Lei_Xiaolin_Antoni_2011} color datasets for performance evaluation. We use the CPSNR metric and SSIM metric  for measuring the reconstruction quality.  SSIM \cite{Wang_Bovik_Sheikh_2004} is a perceptual similarity metric that is more correlated to human perception than CPSNR. In this experiment, we fine-tune the low-pass filter size used to extract chrominance for every image to get the highest possible PSNR. We consider Gaussian filters of sizes in the set $\{9,11,15,17,19,21\}$ and multiplied by hamming windows of the same size. Afterwards, we average the CPSNR and SSIM overall images in the dataset.  

Table~\ref{tab:PSNR} shows the average CPSNR and SSIM values of all CFAs. We notice that our CFAs achieve higher quality in case of crosstalk compared to other CFAs of the same atom size except for $3\times3$ CFAs. For the case of $3\time3$ atom size, our CFA has slightly lower CPSNR and SSIM on McMaster dataset. Figures sets $3-15$, and $16-28$ shows recontructed images for McMaster and Kodak datasets, respectively. 
\begin{table*}[h]
\centering
\caption{Image reconstruction quality measured in CPSNR and SSIM for Kodak and McMaster datasets}
\begin{tabular}{c|c|cc|cc}
& & \multicolumn{2}{|c}{Kodak Dataset} & \multicolumn{2}{|c}{McMaster Dataset} \\
\hline
\cline{3-6}
 &  & w/o Ctk & w/ Ctk & w/o Ctk & w/ Ctk \\
\hline 
\hline
\multirow{ 4}{*}{\rotatebox[origin=c]{0}{$4\times 4$}}
& RGBCY \cite{Anzagira_Fossum_2015} & 31.96/0.9163 & 31.80/0.9132 & 29.64/0.9190 & 29.56/0.9171 \\
& RGBCWY \cite{Anzagira_Fossum_2015}  & 32.04/0.9163 & 31.92/0.9134 & 30.10/0.9254 & 29.99/0.9231\\
& Hao et al. \cite{Hao_Li_Lin_2011}   &  \textbf{33.42/0.9269} & 32.53/0.9134 &  \textbf{31.27/0.9361} &  \textbf{30.78}/0.9275\\
& Ours     & 32.78/0.9263 &  \textbf{32.63/0.9234} & 30.78/0.9343 & 30.73/ \textbf{0.9326} \\
\hline
\multirow{ 2}{*}{\rotatebox[origin=c]{0}{$3\times 3$}}
& Biay-Cheng et al. \cite{Cheng_Siddiqui_Luo_2015} & 33.00/0.9246 & 32.75/0.9206 & 31.04/0.9337 & 30.85/0.9307\\
& Ours 	&  \textbf{33.32/0.9345} &  \textbf{33.22/0.9301} & 31.27/0.9416 & 31.17/0.9386 \\
\hline
\multirow{ 2}{*}{\rotatebox[origin=c]{0}{$3\times 2$}}
& Condat \cite{Condat_2011}&  \textbf{33.79/0.9316} &  \textbf{33.17}/0.9231 &  \textbf{31.88/0.9394} &  \textbf{31.65/0.9349}\\
& Ours & 33.02/0.9259 & 32.88/ \textbf{0.9234} & 31.40/0.9352 & 31.23/0.9329\\
\hline
\multirow{ 2}{*}{\rotatebox[origin=c]{0}{$4\times 2$}}
& Hirakawa-Wolfe \cite{Hirakawa_Wolfe_2008} &  33.47/0.9276 &  32.78/0.9188 & 31.38/0.9347 & 30.99/0.9284 \\
& Ours 	&  \textbf{33.60/0.9292} &  \textbf{33.10/0.9231} &  \textbf{31.88/0.9387} &  \textbf{31.61/0.9348} \\
\hline
\multirow{ 2}{*}{\rotatebox[origin=c]{0}{$7\times 7$}}
& Bai et al. \cite{Bai_Li_Lin_2016} & 33.11/0.9244 & 32.58/0.9157 & 30.71/0.9324 & 30.55/0.9274 \\
& Ours &	 \textbf{34.51/0.9489} &  \textbf{34.20/0.9447} &  \textbf{32.10/0.9495} &  \textbf{32.11/0.9481} \\
\hline
\end{tabular}
\label{tab:PSNR}
\vspace{-2ex}
\end{table*}

\clearpage
\hypertarget{comparison:0}{}
\begin{figure*}[h!]
\centering
\frame{\includegraphics[width=0.7\textwidth]{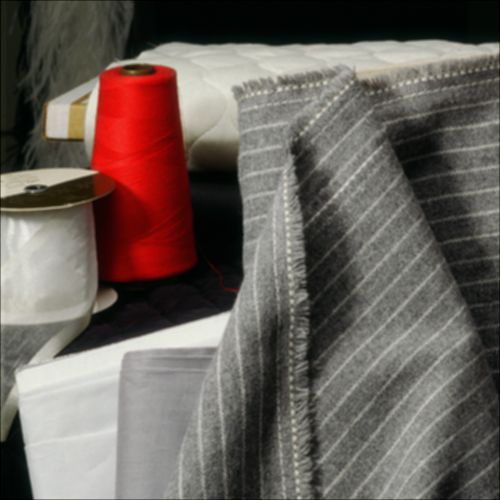}}
\caption{Reconstructed Images using different CFAs}\hypertarget{fig:0}{}
\end{figure*}
\begin{center}
\Large{Ground Truth}
\end{center}
\noindent\newline\vspace{3mm}
\hyperlink{fig:0}{\Large{Ground Truth}}\newline\vspace{2mm}
\Large{$4\times4$}: \hyperlink{fig:1}{\Large{RGBCY\cite{Anzagira_Fossum_2015} }}\hspace{2.0ex}\hyperlink{fig:2}{\Large{RGBCWY\cite{Anzagira_Fossum_2015} }}\hspace{2.0ex}\hyperlink{fig:3}{\Large{Hao et al. \cite{Hao_Li_Lin_2011} }}\hspace{2.0ex}\hyperlink{fig:4}{\Large{Ours}}\newline\vspace{2mm}
\Large{$3\times3$}: \hyperlink{fig:5}{\Large{Biay-Cheng et al. \cite{Cheng_Siddiqui_Luo_2015}}}\hspace{2.0ex}\hyperlink{fig:6}{\Large{Ours}}\hspace{2.0ex}\newline\vspace{2mm}
\Large{$3\times2$}: \hyperlink{fig:7}{\Large{Condat \cite{Condat_2011}}}\hspace{2.0ex}\hyperlink{fig:8}{\Large{Ours}}\hspace{2.0ex}\newline\vspace{2mm}
\Large{$4\times2$}: \hyperlink{fig:9}{\Large{Hirakawa-Wolfe \cite{Hirakawa_Wolfe_2008}}}\hspace{2.0ex}\hyperlink{fig:10}{\Large{Ours}}\hspace{2.0ex}\newline\vspace{2mm}
\Large{$7\times7$}: \hyperlink{fig:11}{\Large{Bai et al. \cite{Bai_Li_Lin_2016}}}\hspace{2.0ex}\hyperlink{fig:12}{\Large{Ours}}\hspace{2.0ex}\newline\vspace{2mm}
\\
\begin{center}
\textcolor{white}{$\leftarrow$ Previous Comparison}\qquad
\hyperlink{comparison:1}{Next Comparison $\rightarrow$}\end{center}
\clearpage
\begin{figure*}[h!]
\centering
\frame{\includegraphics[width=0.7\textwidth]{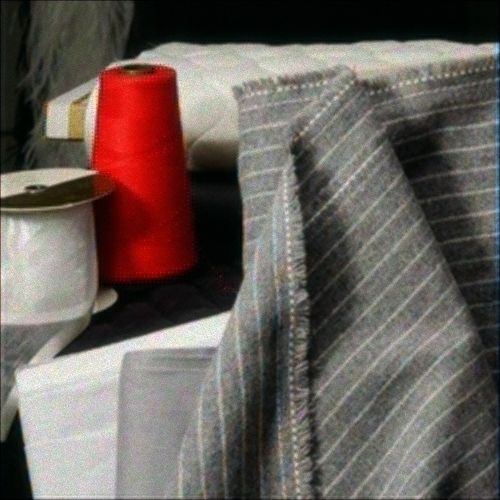}}
\caption{Reconstructed Images using different CFAs}\hypertarget{fig:1}{}
\end{figure*}
\begin{center}
\Large{RGBCY\cite{Anzagira_Fossum_2015}: $\mathrm{PSNR}_{\mathrm{max}}=32.43$dB at $19\times19$ filter size}
\end{center}
\noindent\newline\vspace{3mm}
\hyperlink{fig:0}{\Large{Ground Truth}}\newline\vspace{2mm}
\Large{$4\times4$}: \hyperlink{fig:1}{\Large{RGBCY\cite{Anzagira_Fossum_2015} }}\hspace{2.0ex}\hyperlink{fig:2}{\Large{RGBCWY\cite{Anzagira_Fossum_2015} }}\hspace{2.0ex}\hyperlink{fig:3}{\Large{Hao et al. \cite{Hao_Li_Lin_2011} }}\hspace{2.0ex}\hyperlink{fig:4}{\Large{Ours}}\newline\vspace{2mm}
\Large{$3\times3$}: \hyperlink{fig:5}{\Large{Biay-Cheng et al. \cite{Cheng_Siddiqui_Luo_2015}}}\hspace{2.0ex}\hyperlink{fig:6}{\Large{Ours}}\hspace{2.0ex}\newline\vspace{2mm}
\Large{$3\times2$}: \hyperlink{fig:7}{\Large{Condat \cite{Condat_2011}}}\hspace{2.0ex}\hyperlink{fig:8}{\Large{Ours}}\hspace{2.0ex}\newline\vspace{2mm}
\Large{$4\times2$}: \hyperlink{fig:9}{\Large{Hirakawa-Wolfe \cite{Hirakawa_Wolfe_2008}}}\hspace{2.0ex}\hyperlink{fig:10}{\Large{Ours}}\hspace{2.0ex}\newline\vspace{2mm}
\Large{$7\times7$}: \hyperlink{fig:11}{\Large{Bai et al. \cite{Bai_Li_Lin_2016}}}\hspace{2.0ex}\hyperlink{fig:12}{\Large{Ours}}\hspace{2.0ex}\newline\vspace{2mm}
\\
\begin{center}
\textcolor{white}{$\leftarrow$ Previous Comparison}\qquad
\hyperlink{comparison:1}{Next Comparison $\rightarrow$}\end{center}
\clearpage
\begin{figure*}[h!]
\centering
\frame{\includegraphics[width=0.7\textwidth]{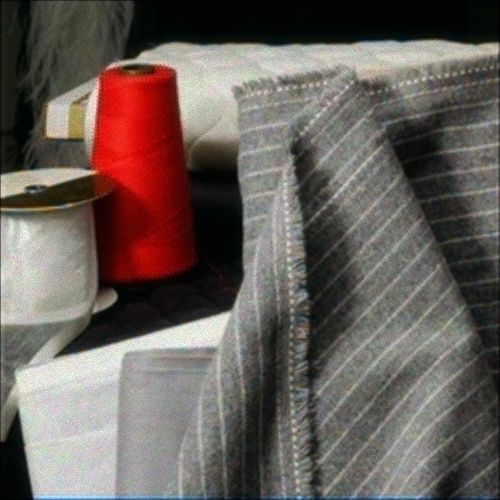}}
\caption{Reconstructed Images using different CFAs}\hypertarget{fig:2}{}
\end{figure*}
\begin{center}
\Large{RGBCWY\cite{Anzagira_Fossum_2015}: $\mathrm{PSNR}_{\mathrm{max}}=32.62$dB at $19\times19$ filter size}
\end{center}
\noindent\newline\vspace{3mm}
\hyperlink{fig:0}{\Large{Ground Truth}}\newline\vspace{2mm}
\Large{$4\times4$}: \hyperlink{fig:1}{\Large{RGBCY\cite{Anzagira_Fossum_2015} }}\hspace{2.0ex}\hyperlink{fig:2}{\Large{RGBCWY\cite{Anzagira_Fossum_2015} }}\hspace{2.0ex}\hyperlink{fig:3}{\Large{Hao et al. \cite{Hao_Li_Lin_2011} }}\hspace{2.0ex}\hyperlink{fig:4}{\Large{Ours}}\newline\vspace{2mm}
\Large{$3\times3$}: \hyperlink{fig:5}{\Large{Biay-Cheng et al. \cite{Cheng_Siddiqui_Luo_2015}}}\hspace{2.0ex}\hyperlink{fig:6}{\Large{Ours}}\hspace{2.0ex}\newline\vspace{2mm}
\Large{$3\times2$}: \hyperlink{fig:7}{\Large{Condat \cite{Condat_2011}}}\hspace{2.0ex}\hyperlink{fig:8}{\Large{Ours}}\hspace{2.0ex}\newline\vspace{2mm}
\Large{$4\times2$}: \hyperlink{fig:9}{\Large{Hirakawa-Wolfe \cite{Hirakawa_Wolfe_2008}}}\hspace{2.0ex}\hyperlink{fig:10}{\Large{Ours}}\hspace{2.0ex}\newline\vspace{2mm}
\Large{$7\times7$}: \hyperlink{fig:11}{\Large{Bai et al. \cite{Bai_Li_Lin_2016}}}\hspace{2.0ex}\hyperlink{fig:12}{\Large{Ours}}\hspace{2.0ex}\newline\vspace{2mm}
\\
\begin{center}
\textcolor{white}{$\leftarrow$ Previous Comparison}\qquad
\hyperlink{comparison:1}{Next Comparison $\rightarrow$}\end{center}
\clearpage
\begin{figure*}[h!]
\centering
\frame{\includegraphics[width=0.7\textwidth]{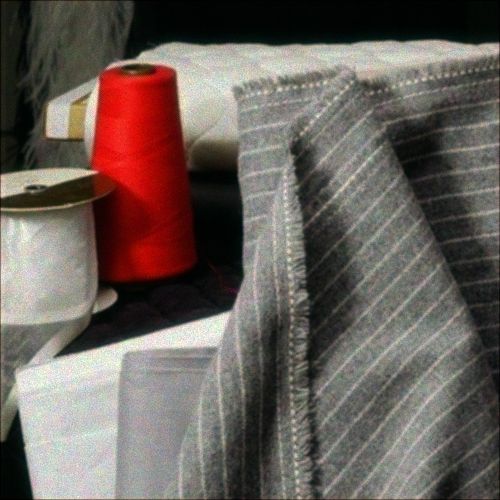}}
\caption{Reconstructed Images using different CFAs}\hypertarget{fig:3}{}
\end{figure*}
\begin{center}
\Large{Hao et al. \cite{Hao_Li_Lin_2011}: $\mathrm{PSNR}_{\mathrm{max}}=32.63$dB at $15\times15$ filter size}
\end{center}
\noindent\newline\vspace{3mm}
\hyperlink{fig:0}{\Large{Ground Truth}}\newline\vspace{2mm}
\Large{$4\times4$}: \hyperlink{fig:1}{\Large{RGBCY\cite{Anzagira_Fossum_2015} }}\hspace{2.0ex}\hyperlink{fig:2}{\Large{RGBCWY\cite{Anzagira_Fossum_2015} }}\hspace{2.0ex}\hyperlink{fig:3}{\Large{Hao et al. \cite{Hao_Li_Lin_2011} }}\hspace{2.0ex}\hyperlink{fig:4}{\Large{Ours}}\newline\vspace{2mm}
\Large{$3\times3$}: \hyperlink{fig:5}{\Large{Biay-Cheng et al. \cite{Cheng_Siddiqui_Luo_2015}}}\hspace{2.0ex}\hyperlink{fig:6}{\Large{Ours}}\hspace{2.0ex}\newline\vspace{2mm}
\Large{$3\times2$}: \hyperlink{fig:7}{\Large{Condat \cite{Condat_2011}}}\hspace{2.0ex}\hyperlink{fig:8}{\Large{Ours}}\hspace{2.0ex}\newline\vspace{2mm}
\Large{$4\times2$}: \hyperlink{fig:9}{\Large{Hirakawa-Wolfe \cite{Hirakawa_Wolfe_2008}}}\hspace{2.0ex}\hyperlink{fig:10}{\Large{Ours}}\hspace{2.0ex}\newline\vspace{2mm}
\Large{$7\times7$}: \hyperlink{fig:11}{\Large{Bai et al. \cite{Bai_Li_Lin_2016}}}\hspace{2.0ex}\hyperlink{fig:12}{\Large{Ours}}\hspace{2.0ex}\newline\vspace{2mm}
\\
\begin{center}
\textcolor{white}{$\leftarrow$ Previous Comparison}\qquad
\hyperlink{comparison:1}{Next Comparison $\rightarrow$}\end{center}
\clearpage
\begin{figure*}[h!]
\centering
\frame{\includegraphics[width=0.7\textwidth]{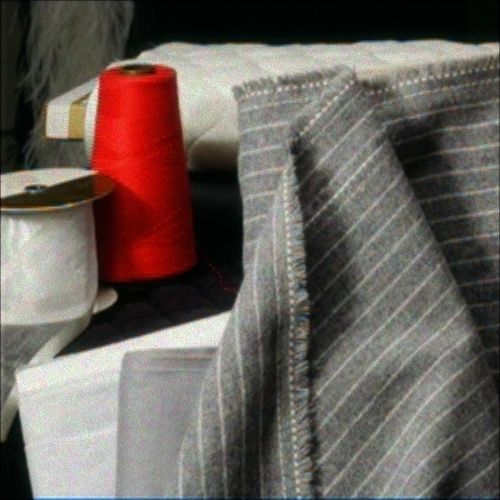}}
\caption{Reconstructed Images using different CFAs}\hypertarget{fig:4}{}
\end{figure*}
\begin{center}
\Large{Ours: $\mathrm{PSNR}_{\mathrm{max}}=33.14$dB at $15\times15$ filter size}
\end{center}
\noindent\newline\vspace{3mm}
\hyperlink{fig:0}{\Large{Ground Truth}}\newline\vspace{2mm}
\Large{$4\times4$}: \hyperlink{fig:1}{\Large{RGBCY\cite{Anzagira_Fossum_2015} }}\hspace{2.0ex}\hyperlink{fig:2}{\Large{RGBCWY\cite{Anzagira_Fossum_2015} }}\hspace{2.0ex}\hyperlink{fig:3}{\Large{Hao et al. \cite{Hao_Li_Lin_2011} }}\hspace{2.0ex}\hyperlink{fig:4}{\Large{Ours}}\newline\vspace{2mm}
\Large{$3\times3$}: \hyperlink{fig:5}{\Large{Biay-Cheng et al. \cite{Cheng_Siddiqui_Luo_2015}}}\hspace{2.0ex}\hyperlink{fig:6}{\Large{Ours}}\hspace{2.0ex}\newline\vspace{2mm}
\Large{$3\times2$}: \hyperlink{fig:7}{\Large{Condat \cite{Condat_2011}}}\hspace{2.0ex}\hyperlink{fig:8}{\Large{Ours}}\hspace{2.0ex}\newline\vspace{2mm}
\Large{$4\times2$}: \hyperlink{fig:9}{\Large{Hirakawa-Wolfe \cite{Hirakawa_Wolfe_2008}}}\hspace{2.0ex}\hyperlink{fig:10}{\Large{Ours}}\hspace{2.0ex}\newline\vspace{2mm}
\Large{$7\times7$}: \hyperlink{fig:11}{\Large{Bai et al. \cite{Bai_Li_Lin_2016}}}\hspace{2.0ex}\hyperlink{fig:12}{\Large{Ours}}\hspace{2.0ex}\newline\vspace{2mm}
\\
\begin{center}
\textcolor{white}{$\leftarrow$ Previous Comparison}\qquad
\hyperlink{comparison:1}{Next Comparison $\rightarrow$}\end{center}
\clearpage
\begin{figure*}[h!]
\centering
\frame{\includegraphics[width=0.7\textwidth]{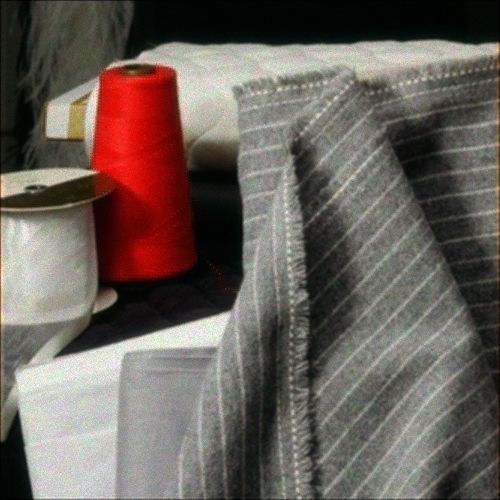}}
\caption{Reconstructed Images using different CFAs}\hypertarget{fig:5}{}
\end{figure*}
\begin{center}
\Large{Biay-Cheng et al. \cite{Cheng_Siddiqui_Luo_2015}: $\mathrm{PSNR}_{\mathrm{max}}=33.15$dB at $15\times15$ filter size}
\end{center}
\noindent\newline\vspace{3mm}
\hyperlink{fig:0}{\Large{Ground Truth}}\newline\vspace{2mm}
\Large{$4\times4$}: \hyperlink{fig:1}{\Large{RGBCY\cite{Anzagira_Fossum_2015} }}\hspace{2.0ex}\hyperlink{fig:2}{\Large{RGBCWY\cite{Anzagira_Fossum_2015} }}\hspace{2.0ex}\hyperlink{fig:3}{\Large{Hao et al. \cite{Hao_Li_Lin_2011} }}\hspace{2.0ex}\hyperlink{fig:4}{\Large{Ours}}\newline\vspace{2mm}
\Large{$3\times3$}: \hyperlink{fig:5}{\Large{Biay-Cheng et al. \cite{Cheng_Siddiqui_Luo_2015}}}\hspace{2.0ex}\hyperlink{fig:6}{\Large{Ours}}\hspace{2.0ex}\newline\vspace{2mm}
\Large{$3\times2$}: \hyperlink{fig:7}{\Large{Condat \cite{Condat_2011}}}\hspace{2.0ex}\hyperlink{fig:8}{\Large{Ours}}\hspace{2.0ex}\newline\vspace{2mm}
\Large{$4\times2$}: \hyperlink{fig:9}{\Large{Hirakawa-Wolfe \cite{Hirakawa_Wolfe_2008}}}\hspace{2.0ex}\hyperlink{fig:10}{\Large{Ours}}\hspace{2.0ex}\newline\vspace{2mm}
\Large{$7\times7$}: \hyperlink{fig:11}{\Large{Bai et al. \cite{Bai_Li_Lin_2016}}}\hspace{2.0ex}\hyperlink{fig:12}{\Large{Ours}}\hspace{2.0ex}\newline\vspace{2mm}
\\
\begin{center}
\textcolor{white}{$\leftarrow$ Previous Comparison}\qquad
\hyperlink{comparison:1}{Next Comparison $\rightarrow$}\end{center}
\clearpage
\begin{figure*}[h!]
\centering
\frame{\includegraphics[width=0.7\textwidth]{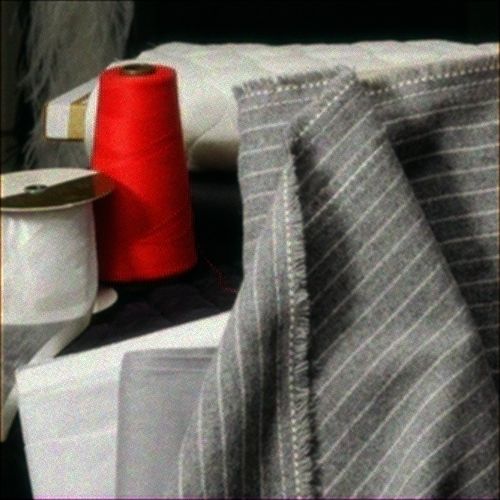}}
\caption{Reconstructed Images using different CFAs}\hypertarget{fig:6}{}
\end{figure*}
\begin{center}
\Large{Ours: $\mathrm{PSNR}_{\mathrm{max}}=33.71$dB at $15\times15$ filter size}
\end{center}
\noindent\newline\vspace{3mm}
\hyperlink{fig:0}{\Large{Ground Truth}}\newline\vspace{2mm}
\Large{$4\times4$}: \hyperlink{fig:1}{\Large{RGBCY\cite{Anzagira_Fossum_2015} }}\hspace{2.0ex}\hyperlink{fig:2}{\Large{RGBCWY\cite{Anzagira_Fossum_2015} }}\hspace{2.0ex}\hyperlink{fig:3}{\Large{Hao et al. \cite{Hao_Li_Lin_2011} }}\hspace{2.0ex}\hyperlink{fig:4}{\Large{Ours}}\newline\vspace{2mm}
\Large{$3\times3$}: \hyperlink{fig:5}{\Large{Biay-Cheng et al. \cite{Cheng_Siddiqui_Luo_2015}}}\hspace{2.0ex}\hyperlink{fig:6}{\Large{Ours}}\hspace{2.0ex}\newline\vspace{2mm}
\Large{$3\times2$}: \hyperlink{fig:7}{\Large{Condat \cite{Condat_2011}}}\hspace{2.0ex}\hyperlink{fig:8}{\Large{Ours}}\hspace{2.0ex}\newline\vspace{2mm}
\Large{$4\times2$}: \hyperlink{fig:9}{\Large{Hirakawa-Wolfe \cite{Hirakawa_Wolfe_2008}}}\hspace{2.0ex}\hyperlink{fig:10}{\Large{Ours}}\hspace{2.0ex}\newline\vspace{2mm}
\Large{$7\times7$}: \hyperlink{fig:11}{\Large{Bai et al. \cite{Bai_Li_Lin_2016}}}\hspace{2.0ex}\hyperlink{fig:12}{\Large{Ours}}\hspace{2.0ex}\newline\vspace{2mm}
\\
\begin{center}
\textcolor{white}{$\leftarrow$ Previous Comparison}\qquad
\hyperlink{comparison:1}{Next Comparison $\rightarrow$}\end{center}
\clearpage
\begin{figure*}[h!]
\centering
\frame{\includegraphics[width=0.7\textwidth]{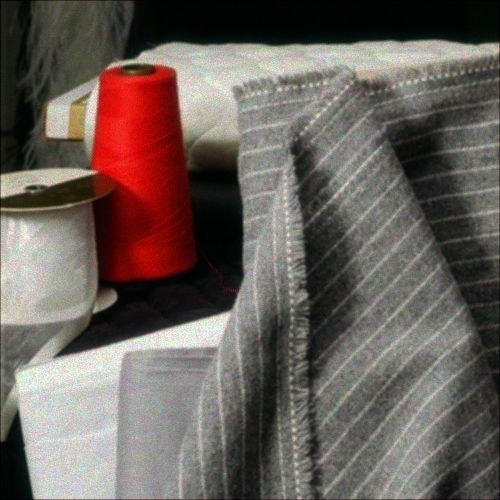}}
\caption{Reconstructed Images using different CFAs}\hypertarget{fig:7}{}
\end{figure*}
\begin{center}
\Large{Condat \cite{Condat_2011}: $\mathrm{PSNR}_{\mathrm{max}}=33.09$dB at $11\times11$ filter size}
\end{center}
\noindent\newline\vspace{3mm}
\hyperlink{fig:0}{\Large{Ground Truth}}\newline\vspace{2mm}
\Large{$4\times4$}: \hyperlink{fig:1}{\Large{RGBCY\cite{Anzagira_Fossum_2015} }}\hspace{2.0ex}\hyperlink{fig:2}{\Large{RGBCWY\cite{Anzagira_Fossum_2015} }}\hspace{2.0ex}\hyperlink{fig:3}{\Large{Hao et al. \cite{Hao_Li_Lin_2011} }}\hspace{2.0ex}\hyperlink{fig:4}{\Large{Ours}}\newline\vspace{2mm}
\Large{$3\times3$}: \hyperlink{fig:5}{\Large{Biay-Cheng et al. \cite{Cheng_Siddiqui_Luo_2015}}}\hspace{2.0ex}\hyperlink{fig:6}{\Large{Ours}}\hspace{2.0ex}\newline\vspace{2mm}
\Large{$3\times2$}: \hyperlink{fig:7}{\Large{Condat \cite{Condat_2011}}}\hspace{2.0ex}\hyperlink{fig:8}{\Large{Ours}}\hspace{2.0ex}\newline\vspace{2mm}
\Large{$4\times2$}: \hyperlink{fig:9}{\Large{Hirakawa-Wolfe \cite{Hirakawa_Wolfe_2008}}}\hspace{2.0ex}\hyperlink{fig:10}{\Large{Ours}}\hspace{2.0ex}\newline\vspace{2mm}
\Large{$7\times7$}: \hyperlink{fig:11}{\Large{Bai et al. \cite{Bai_Li_Lin_2016}}}\hspace{2.0ex}\hyperlink{fig:12}{\Large{Ours}}\hspace{2.0ex}\newline\vspace{2mm}
\\
\begin{center}
\textcolor{white}{$\leftarrow$ Previous Comparison}\qquad
\hyperlink{comparison:1}{Next Comparison $\rightarrow$}\end{center}
\clearpage
\begin{figure*}[h!]
\centering
\frame{\includegraphics[width=0.7\textwidth]{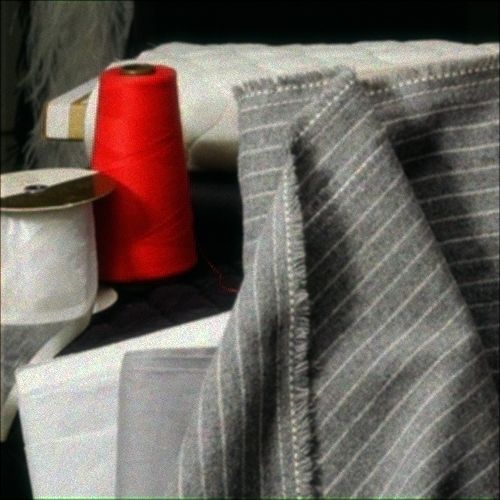}}
\caption{Reconstructed Images using different CFAs}\hypertarget{fig:8}{}
\end{figure*}
\begin{center}
\Large{Ours: $\mathrm{PSNR}_{\mathrm{max}}=33.10$dB at $15\times15$ filter size}
\end{center}
\noindent\newline\vspace{3mm}
\hyperlink{fig:0}{\Large{Ground Truth}}\newline\vspace{2mm}
\Large{$4\times4$}: \hyperlink{fig:1}{\Large{RGBCY\cite{Anzagira_Fossum_2015} }}\hspace{2.0ex}\hyperlink{fig:2}{\Large{RGBCWY\cite{Anzagira_Fossum_2015} }}\hspace{2.0ex}\hyperlink{fig:3}{\Large{Hao et al. \cite{Hao_Li_Lin_2011} }}\hspace{2.0ex}\hyperlink{fig:4}{\Large{Ours}}\newline\vspace{2mm}
\Large{$3\times3$}: \hyperlink{fig:5}{\Large{Biay-Cheng et al. \cite{Cheng_Siddiqui_Luo_2015}}}\hspace{2.0ex}\hyperlink{fig:6}{\Large{Ours}}\hspace{2.0ex}\newline\vspace{2mm}
\Large{$3\times2$}: \hyperlink{fig:7}{\Large{Condat \cite{Condat_2011}}}\hspace{2.0ex}\hyperlink{fig:8}{\Large{Ours}}\hspace{2.0ex}\newline\vspace{2mm}
\Large{$4\times2$}: \hyperlink{fig:9}{\Large{Hirakawa-Wolfe \cite{Hirakawa_Wolfe_2008}}}\hspace{2.0ex}\hyperlink{fig:10}{\Large{Ours}}\hspace{2.0ex}\newline\vspace{2mm}
\Large{$7\times7$}: \hyperlink{fig:11}{\Large{Bai et al. \cite{Bai_Li_Lin_2016}}}\hspace{2.0ex}\hyperlink{fig:12}{\Large{Ours}}\hspace{2.0ex}\newline\vspace{2mm}
\\
\begin{center}
\textcolor{white}{$\leftarrow$ Previous Comparison}\qquad
\hyperlink{comparison:1}{Next Comparison $\rightarrow$}\end{center}
\clearpage
\begin{figure*}[h!]
\centering
\frame{\includegraphics[width=0.7\textwidth]{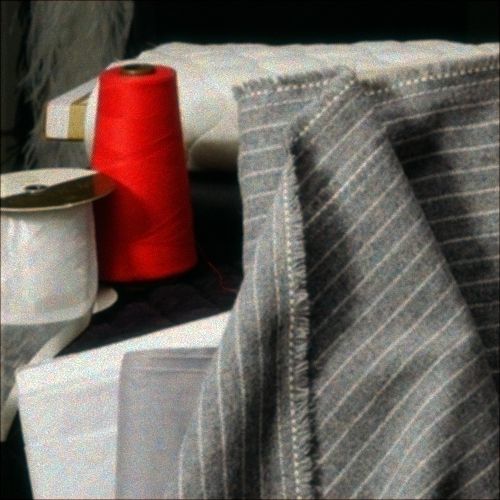}}
\caption{Reconstructed Images using different CFAs}\hypertarget{fig:9}{}
\end{figure*}
\begin{center}
\Large{Hirakawa-Wolfe \cite{Hirakawa_Wolfe_2008}: $\mathrm{PSNR}_{\mathrm{max}}=32.77$dB at $15\times15$ filter size}
\end{center}
\noindent\newline\vspace{3mm}
\hyperlink{fig:0}{\Large{Ground Truth}}\newline\vspace{2mm}
\Large{$4\times4$}: \hyperlink{fig:1}{\Large{RGBCY\cite{Anzagira_Fossum_2015} }}\hspace{2.0ex}\hyperlink{fig:2}{\Large{RGBCWY\cite{Anzagira_Fossum_2015} }}\hspace{2.0ex}\hyperlink{fig:3}{\Large{Hao et al. \cite{Hao_Li_Lin_2011} }}\hspace{2.0ex}\hyperlink{fig:4}{\Large{Ours}}\newline\vspace{2mm}
\Large{$3\times3$}: \hyperlink{fig:5}{\Large{Biay-Cheng et al. \cite{Cheng_Siddiqui_Luo_2015}}}\hspace{2.0ex}\hyperlink{fig:6}{\Large{Ours}}\hspace{2.0ex}\newline\vspace{2mm}
\Large{$3\times2$}: \hyperlink{fig:7}{\Large{Condat \cite{Condat_2011}}}\hspace{2.0ex}\hyperlink{fig:8}{\Large{Ours}}\hspace{2.0ex}\newline\vspace{2mm}
\Large{$4\times2$}: \hyperlink{fig:9}{\Large{Hirakawa-Wolfe \cite{Hirakawa_Wolfe_2008}}}\hspace{2.0ex}\hyperlink{fig:10}{\Large{Ours}}\hspace{2.0ex}\newline\vspace{2mm}
\Large{$7\times7$}: \hyperlink{fig:11}{\Large{Bai et al. \cite{Bai_Li_Lin_2016}}}\hspace{2.0ex}\hyperlink{fig:12}{\Large{Ours}}\hspace{2.0ex}\newline\vspace{2mm}
\\
\begin{center}
\textcolor{white}{$\leftarrow$ Previous Comparison}\qquad
\hyperlink{comparison:1}{Next Comparison $\rightarrow$}\end{center}
\clearpage
\begin{figure*}[h!]
\centering
\frame{\includegraphics[width=0.7\textwidth]{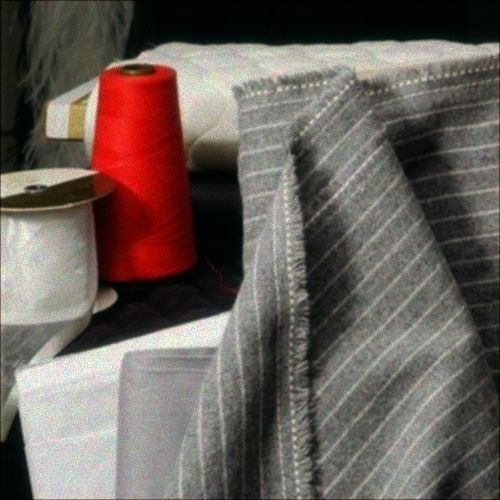}}
\caption{Reconstructed Images using different CFAs}\hypertarget{fig:10}{}
\end{figure*}
\begin{center}
\Large{Ours: $\mathrm{PSNR}_{\mathrm{max}}=33.05$dB at $15\times15$ filter size}
\end{center}
\noindent\newline\vspace{3mm}
\hyperlink{fig:0}{\Large{Ground Truth}}\newline\vspace{2mm}
\Large{$4\times4$}: \hyperlink{fig:1}{\Large{RGBCY\cite{Anzagira_Fossum_2015} }}\hspace{2.0ex}\hyperlink{fig:2}{\Large{RGBCWY\cite{Anzagira_Fossum_2015} }}\hspace{2.0ex}\hyperlink{fig:3}{\Large{Hao et al. \cite{Hao_Li_Lin_2011} }}\hspace{2.0ex}\hyperlink{fig:4}{\Large{Ours}}\newline\vspace{2mm}
\Large{$3\times3$}: \hyperlink{fig:5}{\Large{Biay-Cheng et al. \cite{Cheng_Siddiqui_Luo_2015}}}\hspace{2.0ex}\hyperlink{fig:6}{\Large{Ours}}\hspace{2.0ex}\newline\vspace{2mm}
\Large{$3\times2$}: \hyperlink{fig:7}{\Large{Condat \cite{Condat_2011}}}\hspace{2.0ex}\hyperlink{fig:8}{\Large{Ours}}\hspace{2.0ex}\newline\vspace{2mm}
\Large{$4\times2$}: \hyperlink{fig:9}{\Large{Hirakawa-Wolfe \cite{Hirakawa_Wolfe_2008}}}\hspace{2.0ex}\hyperlink{fig:10}{\Large{Ours}}\hspace{2.0ex}\newline\vspace{2mm}
\Large{$7\times7$}: \hyperlink{fig:11}{\Large{Bai et al. \cite{Bai_Li_Lin_2016}}}\hspace{2.0ex}\hyperlink{fig:12}{\Large{Ours}}\hspace{2.0ex}\newline\vspace{2mm}
\\
\begin{center}
\textcolor{white}{$\leftarrow$ Previous Comparison}\qquad
\hyperlink{comparison:1}{Next Comparison $\rightarrow$}\end{center}
\clearpage
\begin{figure*}[h!]
\centering
\frame{\includegraphics[width=0.7\textwidth]{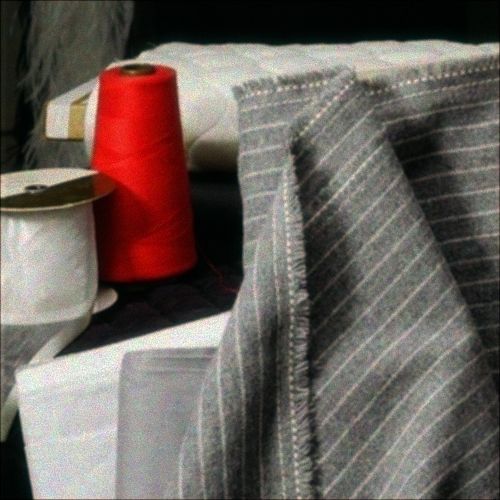}}
\caption{Reconstructed Images using different CFAs}\hypertarget{fig:11}{}
\end{figure*}
\begin{center}
\Large{Bai et al. \cite{Bai_Li_Lin_2016}: $\mathrm{PSNR}_{\mathrm{max}}=32.74$dB at $15\times15$ filter size}
\end{center}
\noindent\newline\vspace{3mm}
\hyperlink{fig:0}{\Large{Ground Truth}}\newline\vspace{2mm}
\Large{$4\times4$}: \hyperlink{fig:1}{\Large{RGBCY\cite{Anzagira_Fossum_2015} }}\hspace{2.0ex}\hyperlink{fig:2}{\Large{RGBCWY\cite{Anzagira_Fossum_2015} }}\hspace{2.0ex}\hyperlink{fig:3}{\Large{Hao et al. \cite{Hao_Li_Lin_2011} }}\hspace{2.0ex}\hyperlink{fig:4}{\Large{Ours}}\newline\vspace{2mm}
\Large{$3\times3$}: \hyperlink{fig:5}{\Large{Biay-Cheng et al. \cite{Cheng_Siddiqui_Luo_2015}}}\hspace{2.0ex}\hyperlink{fig:6}{\Large{Ours}}\hspace{2.0ex}\newline\vspace{2mm}
\Large{$3\times2$}: \hyperlink{fig:7}{\Large{Condat \cite{Condat_2011}}}\hspace{2.0ex}\hyperlink{fig:8}{\Large{Ours}}\hspace{2.0ex}\newline\vspace{2mm}
\Large{$4\times2$}: \hyperlink{fig:9}{\Large{Hirakawa-Wolfe \cite{Hirakawa_Wolfe_2008}}}\hspace{2.0ex}\hyperlink{fig:10}{\Large{Ours}}\hspace{2.0ex}\newline\vspace{2mm}
\Large{$7\times7$}: \hyperlink{fig:11}{\Large{Bai et al. \cite{Bai_Li_Lin_2016}}}\hspace{2.0ex}\hyperlink{fig:12}{\Large{Ours}}\hspace{2.0ex}\newline\vspace{2mm}
\\
\begin{center}
\textcolor{white}{$\leftarrow$ Previous Comparison}\qquad
\hyperlink{comparison:1}{Next Comparison $\rightarrow$}\end{center}
\clearpage
\begin{figure*}[h!]
\centering
\frame{\includegraphics[width=0.7\textwidth]{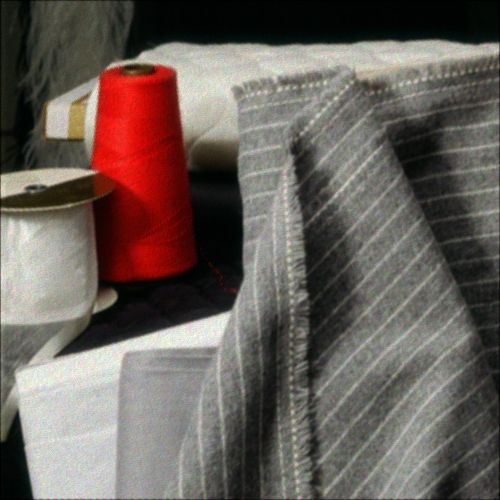}}
\caption{Reconstructed Images using different CFAs}\hypertarget{fig:12}{}
\end{figure*}
\begin{center}
\Large{Ours: $\mathrm{PSNR}_{\mathrm{max}}=34.71$dB at $9\times9$ filter size}
\end{center}
\noindent\newline\vspace{3mm}
\hyperlink{fig:0}{\Large{Ground Truth}}\newline\vspace{2mm}
\Large{$4\times4$}: \hyperlink{fig:1}{\Large{RGBCY\cite{Anzagira_Fossum_2015} }}\hspace{2.0ex}\hyperlink{fig:2}{\Large{RGBCWY\cite{Anzagira_Fossum_2015} }}\hspace{2.0ex}\hyperlink{fig:3}{\Large{Hao et al. \cite{Hao_Li_Lin_2011} }}\hspace{2.0ex}\hyperlink{fig:4}{\Large{Ours}}\newline\vspace{2mm}
\Large{$3\times3$}: \hyperlink{fig:5}{\Large{Biay-Cheng et al. \cite{Cheng_Siddiqui_Luo_2015}}}\hspace{2.0ex}\hyperlink{fig:6}{\Large{Ours}}\hspace{2.0ex}\newline\vspace{2mm}
\Large{$3\times2$}: \hyperlink{fig:7}{\Large{Condat \cite{Condat_2011}}}\hspace{2.0ex}\hyperlink{fig:8}{\Large{Ours}}\hspace{2.0ex}\newline\vspace{2mm}
\Large{$4\times2$}: \hyperlink{fig:9}{\Large{Hirakawa-Wolfe \cite{Hirakawa_Wolfe_2008}}}\hspace{2.0ex}\hyperlink{fig:10}{\Large{Ours}}\hspace{2.0ex}\newline\vspace{2mm}
\Large{$7\times7$}: \hyperlink{fig:11}{\Large{Bai et al. \cite{Bai_Li_Lin_2016}}}\hspace{2.0ex}\hyperlink{fig:12}{\Large{Ours}}\hspace{2.0ex}\newline\vspace{2mm}
\\
\begin{center}
\textcolor{white}{$\leftarrow$ Previous Comparison}\qquad
\hyperlink{comparison:1}{Next Comparison $\rightarrow$}\end{center}
\clearpage
\clearpage
\hypertarget{comparison:1}{}
\begin{figure*}[h!]
\centering
\frame{\includegraphics[width=0.9\textwidth]{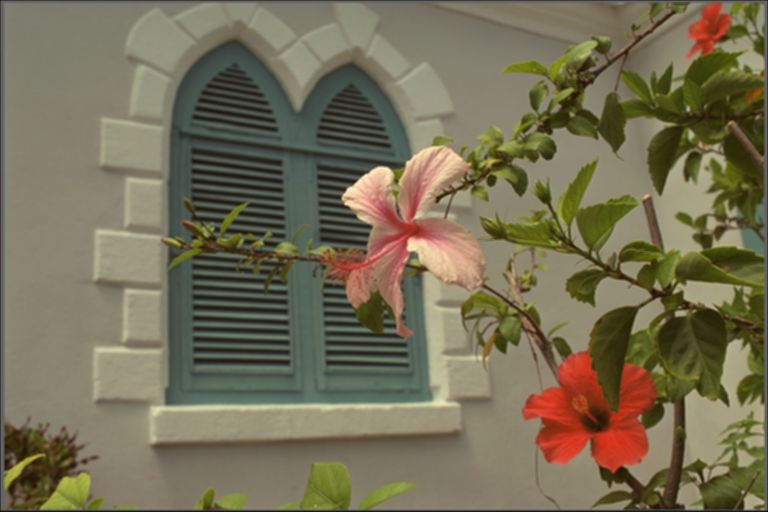}}
\caption{Reconstructed Images using different CFAs}\hypertarget{fig:00}{}
\end{figure*}
\begin{center}
\Large{Ground Truth}
\end{center}
\noindent\newline\vspace{3mm}
\hyperlink{fig:00}{\Large{Ground Truth}}\newline\vspace{2mm}
\Large{$4\times4$}: \hyperlink{fig:111}{\Large{RGBCY\cite{Anzagira_Fossum_2015} }}\hspace{2.0ex}\hyperlink{fig:22}{\Large{RGBCWY\cite{Anzagira_Fossum_2015} }}\hspace{2.0ex}\hyperlink{fig:33}{\Large{Hao et al. \cite{Hao_Li_Lin_2011} }}\hspace{2.0ex}\hyperlink{fig:44}{\Large{Ours}}\newline\vspace{2mm}
\Large{$3\times3$}: \hyperlink{fig:55}{\Large{Biay-Cheng et al. \cite{Cheng_Siddiqui_Luo_2015}}}\hspace{2.0ex}\hyperlink{fig:66}{\Large{Ours}}\hspace{2.0ex}\newline\vspace{2mm}
\Large{$3\times2$}: \hyperlink{fig:77}{\Large{Condat \cite{Condat_2011}}}\hspace{2.0ex}\hyperlink{fig:88}{\Large{Ours}}\hspace{2.0ex}\newline\vspace{2mm}
\Large{$4\times2$}: \hyperlink{fig:99}{\Large{Hirakawa-Wolfe \cite{Hirakawa_Wolfe_2008}}}\hspace{2.0ex}\hyperlink{fig:1010}{\Large{Ours}}\hspace{2.0ex}\newline\vspace{2mm}
\Large{$7\times7$}: \hyperlink{fig:1111}{\Large{Bai et al. \cite{Bai_Li_Lin_2016}}}\hspace{2.0ex}\hyperlink{fig:1212}{\Large{Ours}}\hspace{2.0ex}\newline\vspace{2mm}
\\
\begin{center}
\hyperlink{comparison:0}{$\leftarrow$ Previous Comparison}\qquad
\textcolor{white}{Next Comparison $\rightarrow$}\end{center}
\clearpage
\begin{figure*}[h!]
\centering
\frame{\includegraphics[width=0.9\textwidth]{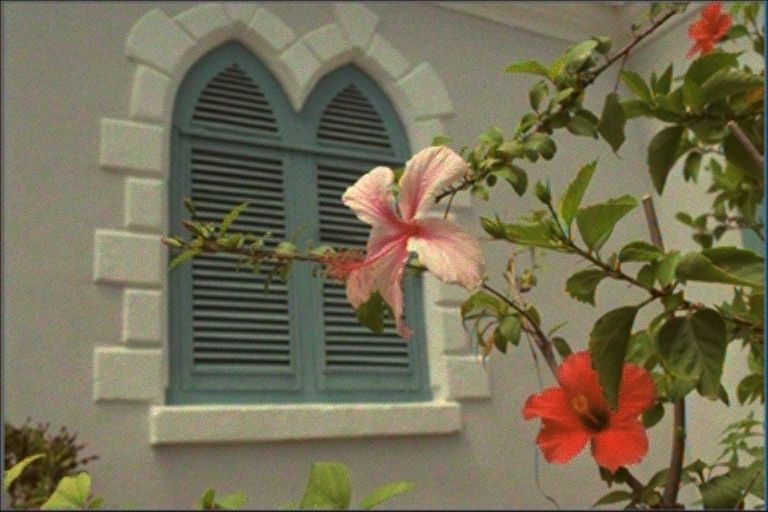}}
\caption{Reconstructed Images using different CFAs}\hypertarget{fig:111}{}
\end{figure*}
\begin{center}
\Large{RGBCY\cite{Anzagira_Fossum_2015}: $\mathrm{PSNR}_{\mathrm{max}}=32.60$dB at $15\times15$ filter size}
\end{center}
\noindent\newline\vspace{3mm}
\hyperlink{fig:00}{\Large{Ground Truth}}\newline\vspace{2mm}
\Large{$4\times4$}: \hyperlink{fig:111}{\Large{RGBCY\cite{Anzagira_Fossum_2015} }}\hspace{2.0ex}\hyperlink{fig:22}{\Large{RGBCWY\cite{Anzagira_Fossum_2015} }}\hspace{2.0ex}\hyperlink{fig:33}{\Large{Hao et al. \cite{Hao_Li_Lin_2011} }}\hspace{2.0ex}\hyperlink{fig:44}{\Large{Ours}}\newline\vspace{2mm}
\Large{$3\times3$}: \hyperlink{fig:55}{\Large{Biay-Cheng et al. \cite{Cheng_Siddiqui_Luo_2015}}}\hspace{2.0ex}\hyperlink{fig:66}{\Large{Ours}}\hspace{2.0ex}\newline\vspace{2mm}
\Large{$3\times2$}: \hyperlink{fig:77}{\Large{Condat \cite{Condat_2011}}}\hspace{2.0ex}\hyperlink{fig:88}{\Large{Ours}}\hspace{2.0ex}\newline\vspace{2mm}
\Large{$4\times2$}: \hyperlink{fig:99}{\Large{Hirakawa-Wolfe \cite{Hirakawa_Wolfe_2008}}}\hspace{2.0ex}\hyperlink{fig:1010}{\Large{Ours}}\hspace{2.0ex}\newline\vspace{2mm}
\Large{$7\times7$}: \hyperlink{fig:1111}{\Large{Bai et al. \cite{Bai_Li_Lin_2016}}}\hspace{2.0ex}\hyperlink{fig:1212}{\Large{Ours}}\hspace{2.0ex}\newline\vspace{2mm}
\\
\begin{center}
\hyperlink{comparison:0}{$\leftarrow$ Previous Comparison}\qquad
\textcolor{white}{Next Comparison $\rightarrow$}\end{center}
\clearpage
\begin{figure*}[h!]
\centering
\frame{\includegraphics[width=0.9\textwidth]{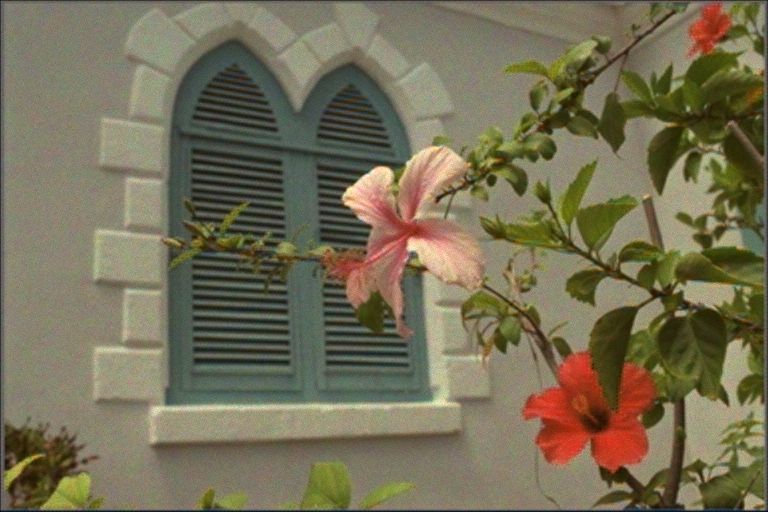}}
\caption{Reconstructed Images using different CFAs}\hypertarget{fig:22}{}
\end{figure*}
\begin{center}
\Large{RGBCWY\cite{Anzagira_Fossum_2015}: $\mathrm{PSNR}_{\mathrm{max}}=32.69$dB at $15\times15$ filter size}
\end{center}
\noindent\newline\vspace{3mm}
\hyperlink{fig:00}{\Large{Ground Truth}}\newline\vspace{2mm}
\Large{$4\times4$}: \hyperlink{fig:111}{\Large{RGBCY\cite{Anzagira_Fossum_2015} }}\hspace{2.0ex}\hyperlink{fig:22}{\Large{RGBCWY\cite{Anzagira_Fossum_2015} }}\hspace{2.0ex}\hyperlink{fig:33}{\Large{Hao et al. \cite{Hao_Li_Lin_2011} }}\hspace{2.0ex}\hyperlink{fig:44}{\Large{Ours}}\newline\vspace{2mm}
\Large{$3\times3$}: \hyperlink{fig:55}{\Large{Biay-Cheng et al. \cite{Cheng_Siddiqui_Luo_2015}}}\hspace{2.0ex}\hyperlink{fig:66}{\Large{Ours}}\hspace{2.0ex}\newline\vspace{2mm}
\Large{$3\times2$}: \hyperlink{fig:77}{\Large{Condat \cite{Condat_2011}}}\hspace{2.0ex}\hyperlink{fig:88}{\Large{Ours}}\hspace{2.0ex}\newline\vspace{2mm}
\Large{$4\times2$}: \hyperlink{fig:99}{\Large{Hirakawa-Wolfe \cite{Hirakawa_Wolfe_2008}}}\hspace{2.0ex}\hyperlink{fig:1010}{\Large{Ours}}\hspace{2.0ex}\newline\vspace{2mm}
\Large{$7\times7$}: \hyperlink{fig:1111}{\Large{Bai et al. \cite{Bai_Li_Lin_2016}}}\hspace{2.0ex}\hyperlink{fig:1212}{\Large{Ours}}\hspace{2.0ex}\newline\vspace{2mm}
\\
\begin{center}
\hyperlink{comparison:0}{$\leftarrow$ Previous Comparison}\qquad
\textcolor{white}{Next Comparison $\rightarrow$}\end{center}
\clearpage
\begin{figure*}[h!]
\centering
\frame{\includegraphics[width=0.9\textwidth]{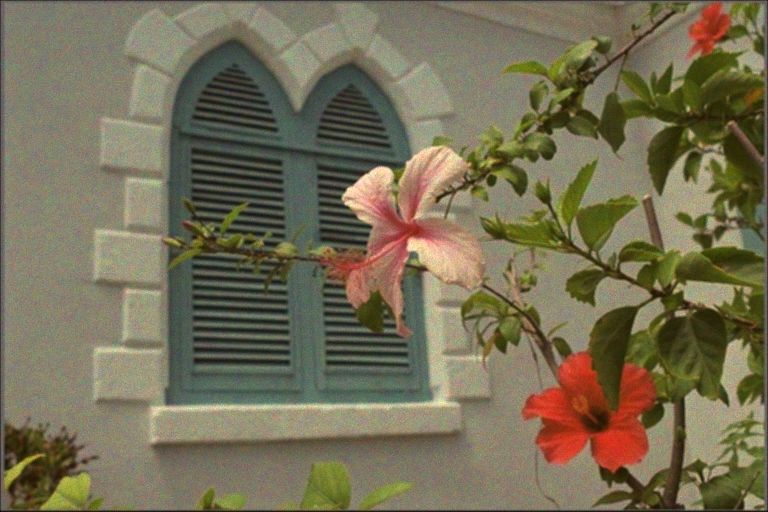}}
\caption{Reconstructed Images using different CFAs}\hypertarget{fig:33}{}
\end{figure*}
\begin{center}
\Large{Hao et al. \cite{Hao_Li_Lin_2011}: $\mathrm{PSNR}_{\mathrm{max}}=32.84$dB at $11\times11$ filter size}
\end{center}
\noindent\newline\vspace{3mm}
\hyperlink{fig:00}{\Large{Ground Truth}}\newline\vspace{2mm}
\Large{$4\times4$}: \hyperlink{fig:111}{\Large{RGBCY\cite{Anzagira_Fossum_2015} }}\hspace{2.0ex}\hyperlink{fig:22}{\Large{RGBCWY\cite{Anzagira_Fossum_2015} }}\hspace{2.0ex}\hyperlink{fig:33}{\Large{Hao et al. \cite{Hao_Li_Lin_2011} }}\hspace{2.0ex}\hyperlink{fig:44}{\Large{Ours}}\newline\vspace{2mm}
\Large{$3\times3$}: \hyperlink{fig:55}{\Large{Biay-Cheng et al. \cite{Cheng_Siddiqui_Luo_2015}}}\hspace{2.0ex}\hyperlink{fig:66}{\Large{Ours}}\hspace{2.0ex}\newline\vspace{2mm}
\Large{$3\times2$}: \hyperlink{fig:77}{\Large{Condat \cite{Condat_2011}}}\hspace{2.0ex}\hyperlink{fig:88}{\Large{Ours}}\hspace{2.0ex}\newline\vspace{2mm}
\Large{$4\times2$}: \hyperlink{fig:99}{\Large{Hirakawa-Wolfe \cite{Hirakawa_Wolfe_2008}}}\hspace{2.0ex}\hyperlink{fig:1010}{\Large{Ours}}\hspace{2.0ex}\newline\vspace{2mm}
\Large{$7\times7$}: \hyperlink{fig:1111}{\Large{Bai et al. \cite{Bai_Li_Lin_2016}}}\hspace{2.0ex}\hyperlink{fig:1212}{\Large{Ours}}\hspace{2.0ex}\newline\vspace{2mm}
\\
\begin{center}
\hyperlink{comparison:0}{$\leftarrow$ Previous Comparison}\qquad
\textcolor{white}{Next Comparison $\rightarrow$}\end{center}
\clearpage
\begin{figure*}[h!]
\centering
\frame{\includegraphics[width=0.9\textwidth]{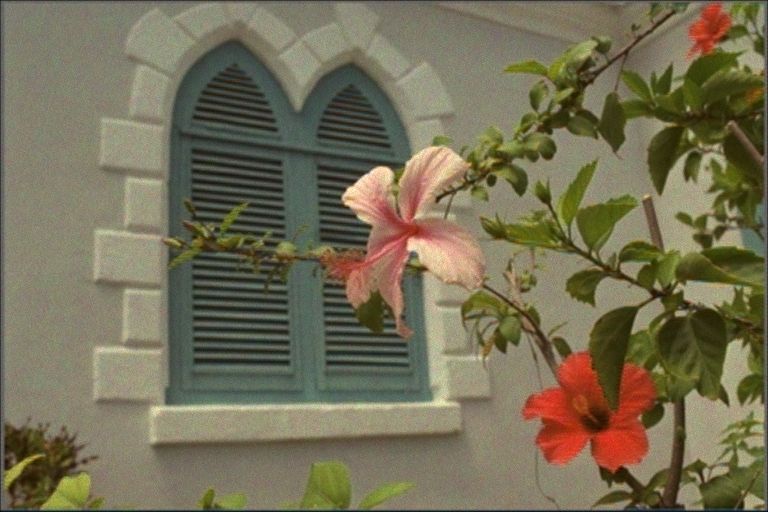}}
\caption{Reconstructed Images using different CFAs}\hypertarget{fig:44}{}
\end{figure*}
\begin{center}
\Large{Ours: $\mathrm{PSNR}_{\mathrm{max}}=33.29$dB at $15\times15$ filter size}
\end{center}
\noindent\newline\vspace{3mm}
\hyperlink{fig:00}{\Large{Ground Truth}}\newline\vspace{2mm}
\Large{$4\times4$}: \hyperlink{fig:111}{\Large{RGBCY\cite{Anzagira_Fossum_2015} }}\hspace{2.0ex}\hyperlink{fig:22}{\Large{RGBCWY\cite{Anzagira_Fossum_2015} }}\hspace{2.0ex}\hyperlink{fig:33}{\Large{Hao et al. \cite{Hao_Li_Lin_2011} }}\hspace{2.0ex}\hyperlink{fig:44}{\Large{Ours}}\newline\vspace{2mm}
\Large{$3\times3$}: \hyperlink{fig:55}{\Large{Biay-Cheng et al. \cite{Cheng_Siddiqui_Luo_2015}}}\hspace{2.0ex}\hyperlink{fig:66}{\Large{Ours}}\hspace{2.0ex}\newline\vspace{2mm}
\Large{$3\times2$}: \hyperlink{fig:77}{\Large{Condat \cite{Condat_2011}}}\hspace{2.0ex}\hyperlink{fig:88}{\Large{Ours}}\hspace{2.0ex}\newline\vspace{2mm}
\Large{$4\times2$}: \hyperlink{fig:99}{\Large{Hirakawa-Wolfe \cite{Hirakawa_Wolfe_2008}}}\hspace{2.0ex}\hyperlink{fig:1010}{\Large{Ours}}\hspace{2.0ex}\newline\vspace{2mm}
\Large{$7\times7$}: \hyperlink{fig:1111}{\Large{Bai et al. \cite{Bai_Li_Lin_2016}}}\hspace{2.0ex}\hyperlink{fig:1212}{\Large{Ours}}\hspace{2.0ex}\newline\vspace{2mm}
\\
\begin{center}
\hyperlink{comparison:0}{$\leftarrow$ Previous Comparison}\qquad
\textcolor{white}{Next Comparison $\rightarrow$}\end{center}
\clearpage
\begin{figure*}[h!]
\centering
\frame{\includegraphics[width=0.9\textwidth]{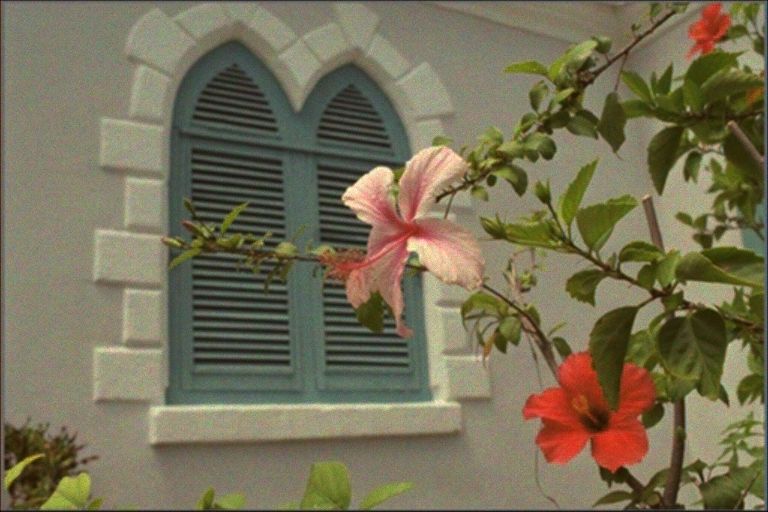}}
\caption{Reconstructed Images using different CFAs}\hypertarget{fig:55}{}
\end{figure*}
\begin{center}
\Large{Biay-Cheng et al. \cite{Cheng_Siddiqui_Luo_2015}: $\mathrm{PSNR}_{\mathrm{max}}=33.09$dB at $11\times11$ filter size}
\end{center}
\noindent\newline\vspace{3mm}
\hyperlink{fig:00}{\Large{Ground Truth}}\newline\vspace{2mm}
\Large{$4\times4$}: \hyperlink{fig:111}{\Large{RGBCY\cite{Anzagira_Fossum_2015} }}\hspace{2.0ex}\hyperlink{fig:22}{\Large{RGBCWY\cite{Anzagira_Fossum_2015} }}\hspace{2.0ex}\hyperlink{fig:33}{\Large{Hao et al. \cite{Hao_Li_Lin_2011} }}\hspace{2.0ex}\hyperlink{fig:44}{\Large{Ours}}\newline\vspace{2mm}
\Large{$3\times3$}: \hyperlink{fig:55}{\Large{Biay-Cheng et al. \cite{Cheng_Siddiqui_Luo_2015}}}\hspace{2.0ex}\hyperlink{fig:66}{\Large{Ours}}\hspace{2.0ex}\newline\vspace{2mm}
\Large{$3\times2$}: \hyperlink{fig:77}{\Large{Condat \cite{Condat_2011}}}\hspace{2.0ex}\hyperlink{fig:88}{\Large{Ours}}\hspace{2.0ex}\newline\vspace{2mm}
\Large{$4\times2$}: \hyperlink{fig:99}{\Large{Hirakawa-Wolfe \cite{Hirakawa_Wolfe_2008}}}\hspace{2.0ex}\hyperlink{fig:1010}{\Large{Ours}}\hspace{2.0ex}\newline\vspace{2mm}
\Large{$7\times7$}: \hyperlink{fig:1111}{\Large{Bai et al. \cite{Bai_Li_Lin_2016}}}\hspace{2.0ex}\hyperlink{fig:1212}{\Large{Ours}}\hspace{2.0ex}\newline\vspace{2mm}
\\
\begin{center}
\hyperlink{comparison:0}{$\leftarrow$ Previous Comparison}\qquad
\textcolor{white}{Next Comparison $\rightarrow$}\end{center}
\clearpage
\begin{figure*}[h!]
\centering
\frame{\includegraphics[width=0.9\textwidth]{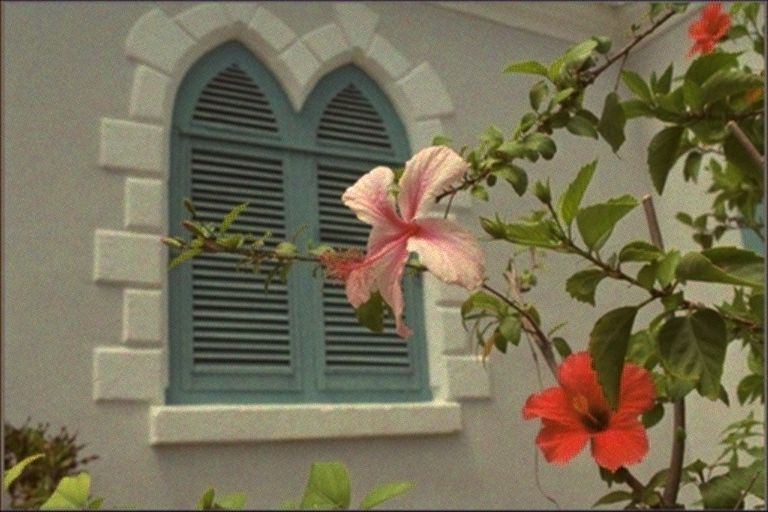}}
\caption{Reconstructed Images using different CFAs}\hypertarget{fig:66}{}
\end{figure*}
\begin{center}
\Large{Ours: $\mathrm{PSNR}_{\mathrm{max}}=33.63$dB at $15\times15$ filter size}
\end{center}
\noindent\newline\vspace{3mm}
\hyperlink{fig:00}{\Large{Ground Truth}}\newline\vspace{2mm}
\Large{$4\times4$}: \hyperlink{fig:111}{\Large{RGBCY\cite{Anzagira_Fossum_2015} }}\hspace{2.0ex}\hyperlink{fig:22}{\Large{RGBCWY\cite{Anzagira_Fossum_2015} }}\hspace{2.0ex}\hyperlink{fig:33}{\Large{Hao et al. \cite{Hao_Li_Lin_2011} }}\hspace{2.0ex}\hyperlink{fig:44}{\Large{Ours}}\newline\vspace{2mm}
\Large{$3\times3$}: \hyperlink{fig:55}{\Large{Biay-Cheng et al. \cite{Cheng_Siddiqui_Luo_2015}}}\hspace{2.0ex}\hyperlink{fig:66}{\Large{Ours}}\hspace{2.0ex}\newline\vspace{2mm}
\Large{$3\times2$}: \hyperlink{fig:77}{\Large{Condat \cite{Condat_2011}}}\hspace{2.0ex}\hyperlink{fig:88}{\Large{Ours}}\hspace{2.0ex}\newline\vspace{2mm}
\Large{$4\times2$}: \hyperlink{fig:99}{\Large{Hirakawa-Wolfe \cite{Hirakawa_Wolfe_2008}}}\hspace{2.0ex}\hyperlink{fig:1010}{\Large{Ours}}\hspace{2.0ex}\newline\vspace{2mm}
\Large{$7\times7$}: \hyperlink{fig:1111}{\Large{Bai et al. \cite{Bai_Li_Lin_2016}}}\hspace{2.0ex}\hyperlink{fig:1212}{\Large{Ours}}\hspace{2.0ex}\newline\vspace{2mm}
\\
\begin{center}
\hyperlink{comparison:0}{$\leftarrow$ Previous Comparison}\qquad
\textcolor{white}{Next Comparison $\rightarrow$}\end{center}
\clearpage
\begin{figure*}[h!]
\centering
\frame{\includegraphics[width=0.9\textwidth]{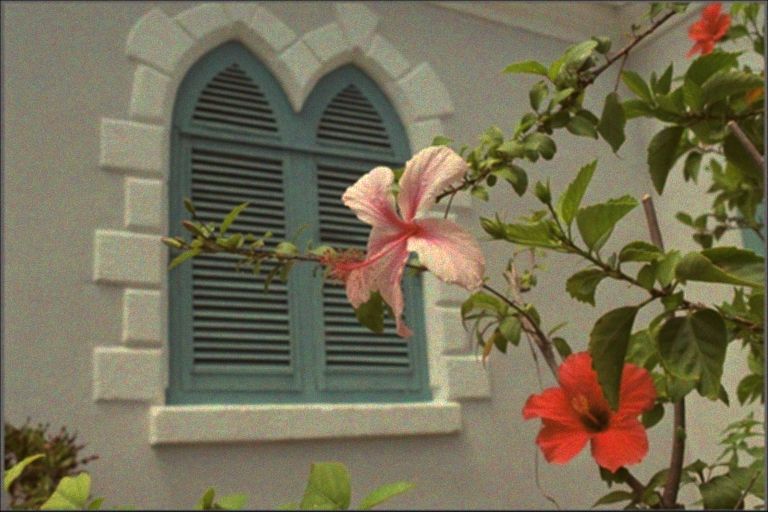}}
\caption{Reconstructed Images using different CFAs}\hypertarget{fig:77}{}
\end{figure*}
\begin{center}
\Large{Condat \cite{Condat_2011}: $\mathrm{PSNR}_{\mathrm{max}}=33.73$dB at $9\times9$ filter size}
\end{center}
\noindent\newline\vspace{3mm}
\hyperlink{fig:00}{\Large{Ground Truth}}\newline\vspace{2mm}
\Large{$4\times4$}: \hyperlink{fig:111}{\Large{RGBCY\cite{Anzagira_Fossum_2015} }}\hspace{2.0ex}\hyperlink{fig:22}{\Large{RGBCWY\cite{Anzagira_Fossum_2015} }}\hspace{2.0ex}\hyperlink{fig:33}{\Large{Hao et al. \cite{Hao_Li_Lin_2011} }}\hspace{2.0ex}\hyperlink{fig:44}{\Large{Ours}}\newline\vspace{2mm}
\Large{$3\times3$}: \hyperlink{fig:55}{\Large{Biay-Cheng et al. \cite{Cheng_Siddiqui_Luo_2015}}}\hspace{2.0ex}\hyperlink{fig:66}{\Large{Ours}}\hspace{2.0ex}\newline\vspace{2mm}
\Large{$3\times2$}: \hyperlink{fig:77}{\Large{Condat \cite{Condat_2011}}}\hspace{2.0ex}\hyperlink{fig:88}{\Large{Ours}}\hspace{2.0ex}\newline\vspace{2mm}
\Large{$4\times2$}: \hyperlink{fig:99}{\Large{Hirakawa-Wolfe \cite{Hirakawa_Wolfe_2008}}}\hspace{2.0ex}\hyperlink{fig:1010}{\Large{Ours}}\hspace{2.0ex}\newline\vspace{2mm}
\Large{$7\times7$}: \hyperlink{fig:1111}{\Large{Bai et al. \cite{Bai_Li_Lin_2016}}}\hspace{2.0ex}\hyperlink{fig:1212}{\Large{Ours}}\hspace{2.0ex}\newline\vspace{2mm}
\\
\begin{center}
\hyperlink{comparison:0}{$\leftarrow$ Previous Comparison}\qquad
\textcolor{white}{Next Comparison $\rightarrow$}\end{center}
\clearpage
\begin{figure*}[h!]
\centering
\frame{\includegraphics[width=0.9\textwidth]{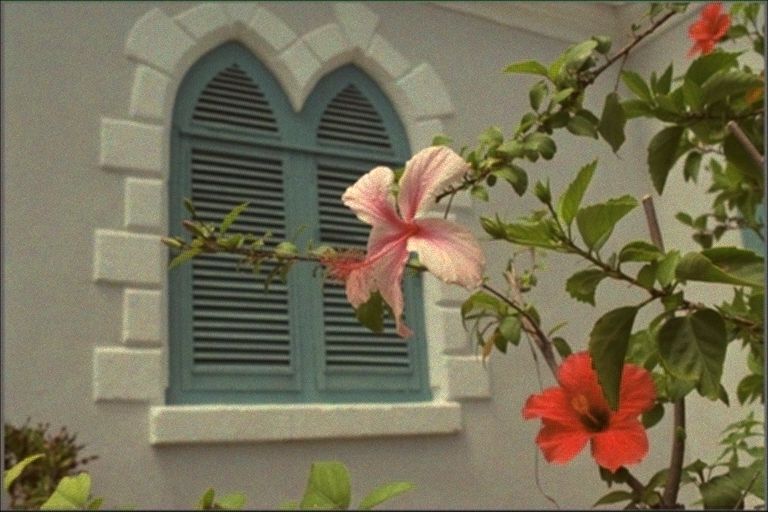}}
\caption{Reconstructed Images using different CFAs}\hypertarget{fig:88}{}
\end{figure*}
\begin{center}
\Large{Ours: $\mathrm{PSNR}_{\mathrm{max}}=33.38$dB at $15\times15$ filter size}
\end{center}
\noindent\newline\vspace{3mm}
\hyperlink{fig:00}{\Large{Ground Truth}}\newline\vspace{2mm}
\Large{$4\times4$}: \hyperlink{fig:111}{\Large{RGBCY\cite{Anzagira_Fossum_2015} }}\hspace{2.0ex}\hyperlink{fig:22}{\Large{RGBCWY\cite{Anzagira_Fossum_2015} }}\hspace{2.0ex}\hyperlink{fig:33}{\Large{Hao et al. \cite{Hao_Li_Lin_2011} }}\hspace{2.0ex}\hyperlink{fig:44}{\Large{Ours}}\newline\vspace{2mm}
\Large{$3\times3$}: \hyperlink{fig:55}{\Large{Biay-Cheng et al. \cite{Cheng_Siddiqui_Luo_2015}}}\hspace{2.0ex}\hyperlink{fig:66}{\Large{Ours}}\hspace{2.0ex}\newline\vspace{2mm}
\Large{$3\times2$}: \hyperlink{fig:77}{\Large{Condat \cite{Condat_2011}}}\hspace{2.0ex}\hyperlink{fig:88}{\Large{Ours}}\hspace{2.0ex}\newline\vspace{2mm}
\Large{$4\times2$}: \hyperlink{fig:99}{\Large{Hirakawa-Wolfe \cite{Hirakawa_Wolfe_2008}}}\hspace{2.0ex}\hyperlink{fig:1010}{\Large{Ours}}\hspace{2.0ex}\newline\vspace{2mm}
\Large{$7\times7$}: \hyperlink{fig:1111}{\Large{Bai et al. \cite{Bai_Li_Lin_2016}}}\hspace{2.0ex}\hyperlink{fig:1212}{\Large{Ours}}\hspace{2.0ex}\newline\vspace{2mm}
\\
\begin{center}
\hyperlink{comparison:0}{$\leftarrow$ Previous Comparison}\qquad
\textcolor{white}{Next Comparison $\rightarrow$}\end{center}
\clearpage
\begin{figure*}[h!]
\centering
\frame{\includegraphics[width=0.9\textwidth]{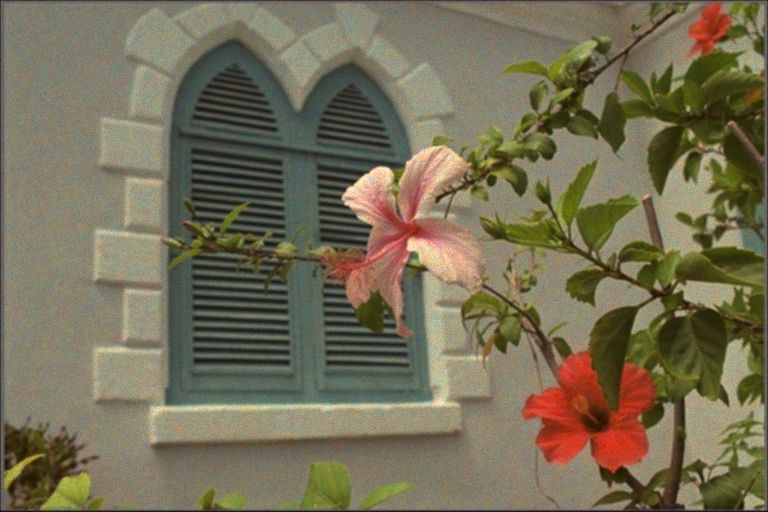}}
\caption{Reconstructed Images using different CFAs}\hypertarget{fig:99}{}
\end{figure*}
\begin{center}
\Large{Hirakawa-Wolfe \cite{Hirakawa_Wolfe_2008}: $\mathrm{PSNR}_{\mathrm{max}}=33.22$dB at $11\times11$ filter size}
\end{center}
\noindent\newline\vspace{3mm}
\hyperlink{fig:00}{\Large{Ground Truth}}\newline\vspace{2mm}
\Large{$4\times4$}: \hyperlink{fig:111}{\Large{RGBCY\cite{Anzagira_Fossum_2015} }}\hspace{2.0ex}\hyperlink{fig:22}{\Large{RGBCWY\cite{Anzagira_Fossum_2015} }}\hspace{2.0ex}\hyperlink{fig:33}{\Large{Hao et al. \cite{Hao_Li_Lin_2011} }}\hspace{2.0ex}\hyperlink{fig:44}{\Large{Ours}}\newline\vspace{2mm}
\Large{$3\times3$}: \hyperlink{fig:55}{\Large{Biay-Cheng et al. \cite{Cheng_Siddiqui_Luo_2015}}}\hspace{2.0ex}\hyperlink{fig:66}{\Large{Ours}}\hspace{2.0ex}\newline\vspace{2mm}
\Large{$3\times2$}: \hyperlink{fig:77}{\Large{Condat \cite{Condat_2011}}}\hspace{2.0ex}\hyperlink{fig:88}{\Large{Ours}}\hspace{2.0ex}\newline\vspace{2mm}
\Large{$4\times2$}: \hyperlink{fig:99}{\Large{Hirakawa-Wolfe \cite{Hirakawa_Wolfe_2008}}}\hspace{2.0ex}\hyperlink{fig:1010}{\Large{Ours}}\hspace{2.0ex}\newline\vspace{2mm}
\Large{$7\times7$}: \hyperlink{fig:1111}{\Large{Bai et al. \cite{Bai_Li_Lin_2016}}}\hspace{2.0ex}\hyperlink{fig:1212}{\Large{Ours}}\hspace{2.0ex}\newline\vspace{2mm}
\\
\begin{center}
\hyperlink{comparison:0}{$\leftarrow$ Previous Comparison}\qquad
\textcolor{white}{Next Comparison $\rightarrow$}\end{center}
\clearpage
\begin{figure*}[h!]
\centering
\frame{\includegraphics[width=0.9\textwidth]{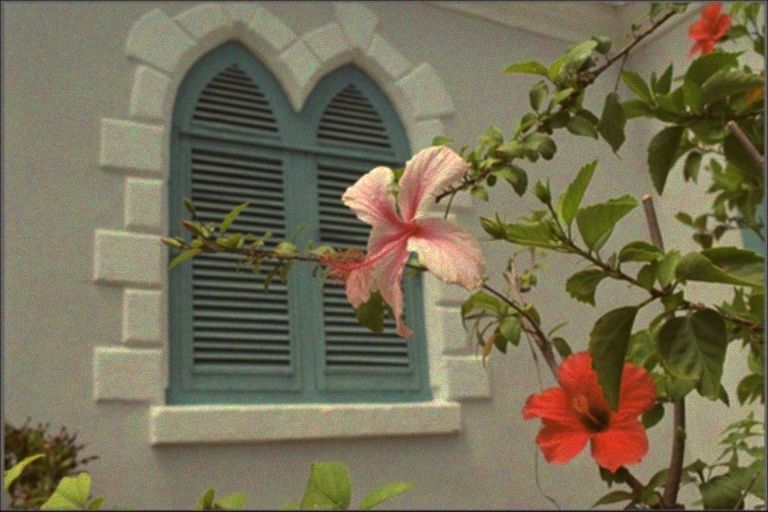}}
\caption{Reconstructed Images using different CFAs}\hypertarget{fig:1010}{}
\end{figure*}
\begin{center}
\Large{Ours: $\mathrm{PSNR}_{\mathrm{max}}=33.61$dB at $11\times11$ filter size}
\end{center}
\noindent\newline\vspace{3mm}
\hyperlink{fig:00}{\Large{Ground Truth}}\newline\vspace{2mm}
\Large{$4\times4$}: \hyperlink{fig:111}{\Large{RGBCY\cite{Anzagira_Fossum_2015} }}\hspace{2.0ex}\hyperlink{fig:22}{\Large{RGBCWY\cite{Anzagira_Fossum_2015} }}\hspace{2.0ex}\hyperlink{fig:33}{\Large{Hao et al. \cite{Hao_Li_Lin_2011} }}\hspace{2.0ex}\hyperlink{fig:44}{\Large{Ours}}\newline\vspace{2mm}
\Large{$3\times3$}: \hyperlink{fig:55}{\Large{Biay-Cheng et al. \cite{Cheng_Siddiqui_Luo_2015}}}\hspace{2.0ex}\hyperlink{fig:66}{\Large{Ours}}\hspace{2.0ex}\newline\vspace{2mm}
\Large{$3\times2$}: \hyperlink{fig:77}{\Large{Condat \cite{Condat_2011}}}\hspace{2.0ex}\hyperlink{fig:88}{\Large{Ours}}\hspace{2.0ex}\newline\vspace{2mm}
\Large{$4\times2$}: \hyperlink{fig:99}{\Large{Hirakawa-Wolfe \cite{Hirakawa_Wolfe_2008}}}\hspace{2.0ex}\hyperlink{fig:1010}{\Large{Ours}}\hspace{2.0ex}\newline\vspace{2mm}
\Large{$7\times7$}: \hyperlink{fig:1111}{\Large{Bai et al. \cite{Bai_Li_Lin_2016}}}\hspace{2.0ex}\hyperlink{fig:1212}{\Large{Ours}}\hspace{2.0ex}\newline\vspace{2mm}
\\
\begin{center}
\hyperlink{comparison:0}{$\leftarrow$ Previous Comparison}\qquad
\textcolor{white}{Next Comparison $\rightarrow$}\end{center}
\clearpage
\begin{figure*}[h!]
\centering
\frame{\includegraphics[width=0.9\textwidth]{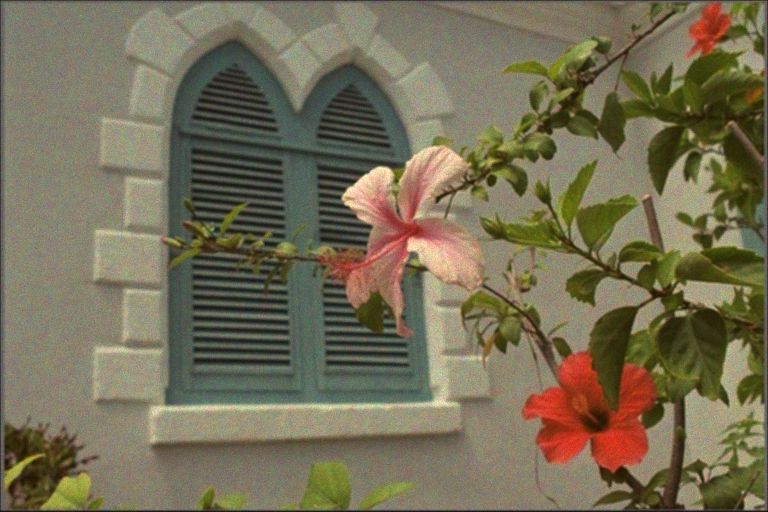}}
\caption{Reconstructed Images using different CFAs}\hypertarget{fig:1111}{}
\end{figure*}
\begin{center}
\Large{Bai et al. \cite{Bai_Li_Lin_2016}: $\mathrm{PSNR}_{\mathrm{max}}=32.77$dB at $11\times11$ filter size}
\end{center}
\noindent\newline\vspace{3mm}
\hyperlink{fig:00}{\Large{Ground Truth}}\newline\vspace{2mm}
\Large{$4\times4$}: \hyperlink{fig:111}{\Large{RGBCY\cite{Anzagira_Fossum_2015} }}\hspace{2.0ex}\hyperlink{fig:22}{\Large{RGBCWY\cite{Anzagira_Fossum_2015} }}\hspace{2.0ex}\hyperlink{fig:33}{\Large{Hao et al. \cite{Hao_Li_Lin_2011} }}\hspace{2.0ex}\hyperlink{fig:44}{\Large{Ours}}\newline\vspace{2mm}
\Large{$3\times3$}: \hyperlink{fig:55}{\Large{Biay-Cheng et al. \cite{Cheng_Siddiqui_Luo_2015}}}\hspace{2.0ex}\hyperlink{fig:66}{\Large{Ours}}\hspace{2.0ex}\newline\vspace{2mm}
\Large{$3\times2$}: \hyperlink{fig:77}{\Large{Condat \cite{Condat_2011}}}\hspace{2.0ex}\hyperlink{fig:88}{\Large{Ours}}\hspace{2.0ex}\newline\vspace{2mm}
\Large{$4\times2$}: \hyperlink{fig:99}{\Large{Hirakawa-Wolfe \cite{Hirakawa_Wolfe_2008}}}\hspace{2.0ex}\hyperlink{fig:1010}{\Large{Ours}}\hspace{2.0ex}\newline\vspace{2mm}
\Large{$7\times7$}: \hyperlink{fig:1111}{\Large{Bai et al. \cite{Bai_Li_Lin_2016}}}\hspace{2.0ex}\hyperlink{fig:1212}{\Large{Ours}}\hspace{2.0ex}\newline\vspace{2mm}
\\
\begin{center}
\hyperlink{comparison:0}{$\leftarrow$ Previous Comparison}\qquad
\textcolor{white}{Next Comparison $\rightarrow$}\end{center}
\clearpage
\begin{figure*}[h!]
\centering
\frame{\includegraphics[width=0.9\textwidth]{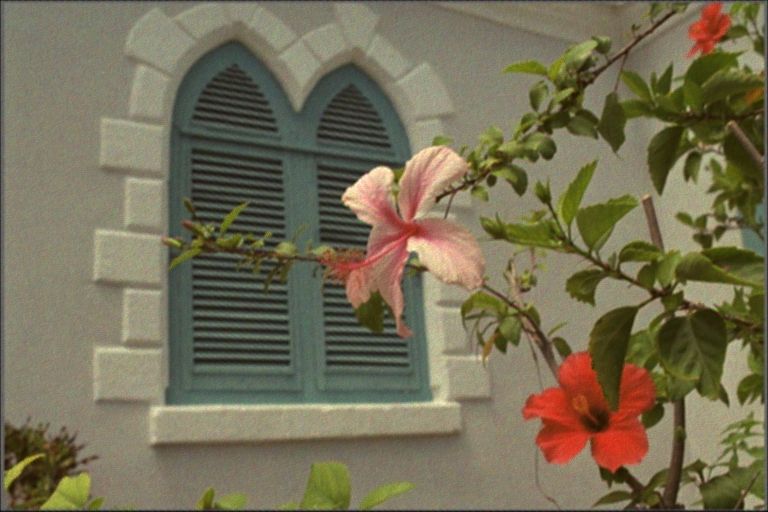}}
\caption{Reconstructed Images using different CFAs}\hypertarget{fig:1212}{}
\end{figure*}
\begin{center}
\Large{Ours: $\mathrm{PSNR}_{\mathrm{max}}=34.86$dB at $9\times9$ filter size}
\end{center}
\noindent\newline\vspace{3mm}
\hyperlink{fig:00}{\Large{Ground Truth}}\newline\vspace{2mm}
\Large{$4\times4$}: \hyperlink{fig:111}{\Large{RGBCY\cite{Anzagira_Fossum_2015} }}\hspace{2.0ex}\hyperlink{fig:22}{\Large{RGBCWY\cite{Anzagira_Fossum_2015} }}\hspace{2.0ex}\hyperlink{fig:33}{\Large{Hao et al. \cite{Hao_Li_Lin_2011} }}\hspace{2.0ex}\hyperlink{fig:44}{\Large{Ours}}\newline\vspace{2mm}
\Large{$3\times3$}: \hyperlink{fig:55}{\Large{Biay-Cheng et al. \cite{Cheng_Siddiqui_Luo_2015}}}\hspace{2.0ex}\hyperlink{fig:66}{\Large{Ours}}\hspace{2.0ex}\newline\vspace{2mm}
\Large{$3\times2$}: \hyperlink{fig:77}{\Large{Condat \cite{Condat_2011}}}\hspace{2.0ex}\hyperlink{fig:88}{\Large{Ours}}\hspace{2.0ex}\newline\vspace{2mm}
\Large{$4\times2$}: \hyperlink{fig:99}{\Large{Hirakawa-Wolfe \cite{Hirakawa_Wolfe_2008}}}\hspace{2.0ex}\hyperlink{fig:1010}{\Large{Ours}}\hspace{2.0ex}\newline\vspace{2mm}
\Large{$7\times7$}: \hyperlink{fig:1111}{\Large{Bai et al. \cite{Bai_Li_Lin_2016}}}\hspace{2.0ex}\hyperlink{fig:1212}{\Large{Ours}}\hspace{2.0ex}\newline\vspace{2mm}
\\
\begin{center}
\hyperlink{comparison:0}{$\leftarrow$ Previous Comparison}\qquad
\textcolor{white}{Next Comparison $\rightarrow$}\end{center}
\clearpage
\normalsize
\bibliographystyle{IEEEbib}
\bibliography{refs}